\pdfoutput=1

\documentclass[11pt]{article}

\usepackage[preprint]{acl}

\usepackage{times}
\usepackage{latexsym}
\usepackage[T1]{fontenc}
\usepackage[utf8]{inputenc}
\usepackage{microtype}
\usepackage{inconsolata}
\usepackage{graphicx}
\usepackage{multicol}
\usepackage{algpseudocode} 
\usepackage{tcolorbox}
\tcbuselibrary{skins, breakable, theorems}
\usepackage{amsmath}
\usepackage{amssymb}
\usepackage{amsthm}
\usepackage{booktabs}
\usepackage{framed} 
\usepackage{multirow}
\usepackage{enumitem}
\usepackage{caption}
\usepackage{float} 
\usepackage{etoolbox}
\usepackage{xspace}
\usepackage{accents}
\usepackage{subfig}
\usepackage{colortbl}
\usepackage{xcolor}
\usepackage{arydshln}
\usepackage{hyperref}
\usepackage[ruled,vlined]{algorithm2e} 
\usepackage{array}

\newtheorem{prob}{Problem}

\newcommand{\eg}{\emph{e.g.},\xspace}

\newcommand{\ie}{\emph{i.e.},\xspace}

\newcommand{\etc}{\emph{etc.}\xspace}

\newcommand\figref[1]{Figure~\ref{#1}}
\newcommand\algoref[1]{Algorithm~\ref{#1}}
\newcommand\tabref[1]{Table~\ref{#1}}
\newcommand\secref[1]{Section~\ref{#1}}
\newcommand\appref[1]{Appendix~\ref{#1}}

\newcommand{\eat}[1]{}

\title{TP-RAG: Benchmarking Retrieval-Augmented Large Language Model Agents for Spatiotemporal-Aware Travel Planning}

\author{
 \textbf{Hang Ni\textsuperscript{1,2}\thanks{Work done during the internship at Baidu Inc..}},
 \textbf{Fan Liu\textsuperscript{1}},
 \textbf{Xinyu Ma\textsuperscript{2}},
 \textbf{Lixin Su\textsuperscript{2}},
 \textbf{Shuaiqiang Wang\textsuperscript{2}}
\\
 \textbf{Dawei Yin\textsuperscript{2}\thanks{Corresponding authors.}}, 
 \textbf{Hui Xiong\textsuperscript{1}}
 \textbf{Hao Liu\textsuperscript{1}\footnotemark[2]},
\\
 \textsuperscript{1}The Hong Kong University of Science and Technology (Guangzhou),
 \textsuperscript{2}Baidu Inc.
\\
\texttt{\{hni017,fliu236\}@connect.hkust-gz.edu.cn}\\ 
\texttt{\{xinyuma2016,sulixinict,shqiang.wang\}@gmail.com}\\
\texttt{yindawei@acm.org, \{xionghui,liuh\}@ust.hk} \\
}

\begin{document}
\maketitle

\begin{abstract}
Large language models (LLMs) have shown promise in automating travel planning, yet they often fall short in addressing nuanced spatiotemporal rationality. 
While existing benchmarks focus on basic plan validity, they neglect critical aspects such as route efficiency, POI appeal, and real-time adaptability. 
This paper introduces \textit{TP-RAG}, the first benchmark tailored for retrieval-augmented, spatiotemporal-aware travel planning. 
Our dataset includes 2,348 real-world travel queries, 85,575 fine-grain annotated POIs, and 18,784 high-quality travel trajectory references sourced from online tourist documents, enabling dynamic and context-aware planning. 
Through extensive experiments, we reveal that integrating reference trajectories significantly improves spatial efficiency and POI rationality of the travel plan, while challenges persist in universality and robustness due to conflicting references and noisy data.
To address these issues, we propose \textit{EvoRAG}, an evolutionary framework that potently synergizes diverse retrieved trajectories with LLMs’ intrinsic reasoning. 
\textit{EvoRAG} achieves state-of-the-art performance, improving spatiotemporal compliance and reducing commonsense violation compared to ground-up and retrieval-augmented baselines. 
Our work underscores the potential of hybridizing Web knowledge with LLM-driven optimization, paving the way for more reliable and adaptive travel planning agents.
\end{abstract}
\section{Introduction}

Emerging studies have explored the potential of Large Language Models (LLMs) to serve as travel agents capable of interpreting natural language inquiries, and autonomously generating travel plans that comprise daily tourist activities detailed with various Points of Interest (POIs)~\cite{wong2023autonomous}.
Despite their promise, existing works~\cite{xie2024travelplanner,singh2024personal,hao2024large} focus primarily on whether the generated plans meet basic commonsense requirements (\eg complete information, actual POIs), while overlooking the nuanced spatiotemporal rationality that is critical for practicality, such as spatiotemporal coherence (\eg transit efficiency and schedule comfort), POI attractiveness (\ie scenic spots with local cultural characteristics and high popularity), and temporal adaptability (\ie capturing time-evolving POI information, such as seasonal closures and altered opening hours).
These oversights may result in flawed plans characterized by inefficient routes, exhausting journeys, unappealing POIs, or limited flexibility. Therefore, this study investigates the capabilities of LLM agents in travel planning while emphasizing spatiotemporal awareness.

\begin{figure}[t]
\centering
\includegraphics[width=\linewidth]{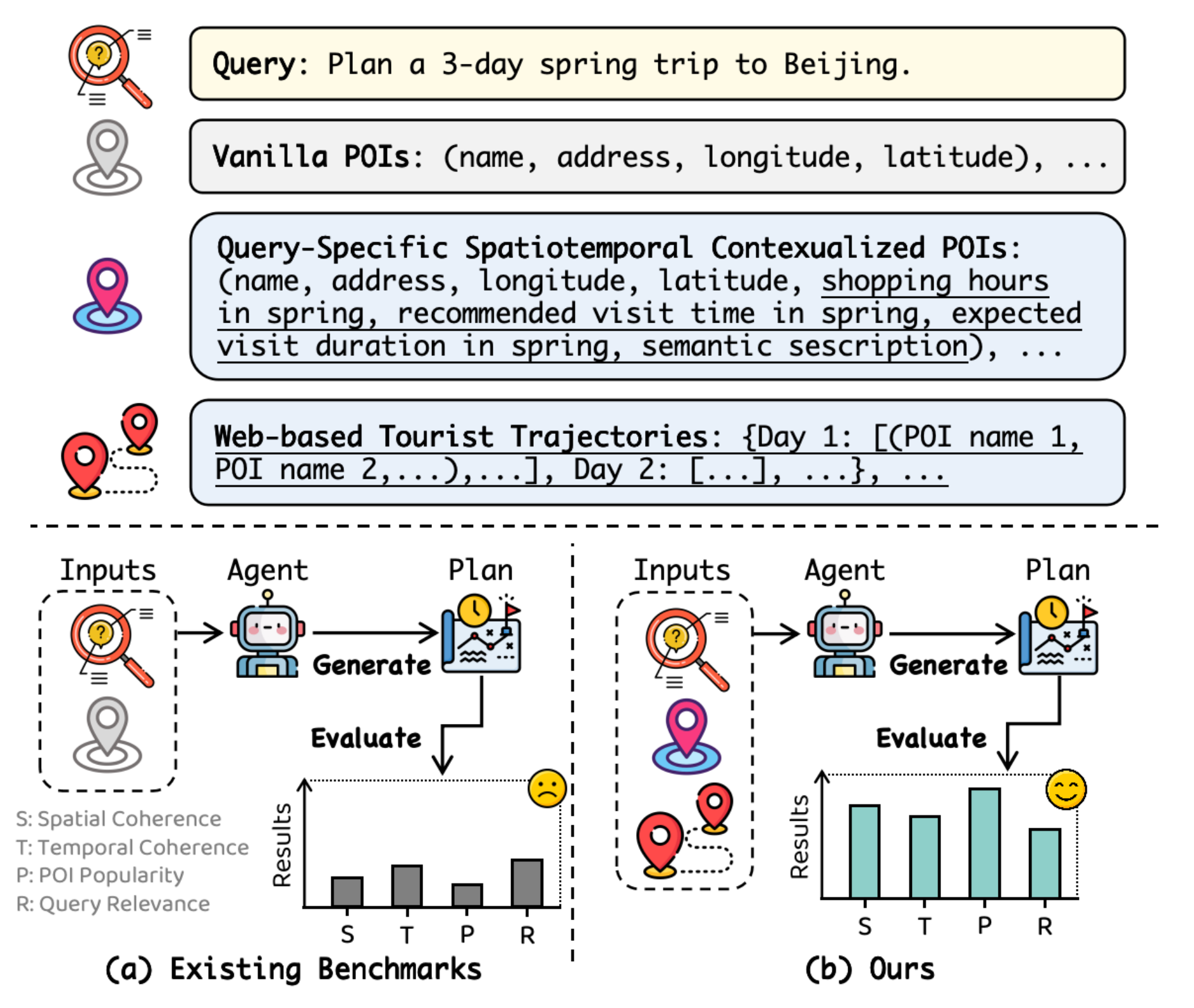}
\vspace{-24pt}
\caption{TP-RAG distinguishes itself from existing works by: (1) query-specific spatiotemporal contextualization and (2) trajectory-level knowledge utilization.}
\vspace{-22pt}
\label{figure:motivation}
\end{figure}

Contemporary research primarily focuses on benchmarking the abilities of LLM agents in travel planning~\cite{xie2024travelplanner,singh2024personal} or exploring sophisticated strategies to enhance planning effectiveness~\cite{tang2024itinera,xie2024human}.
However, these studies rely exclusively on the POI-level knowledge, \ie the metadata of the candidate POIs, and agents’ internal reasoning capabilities to construct travel plans from scratch.
This ground-up approach faces inherent limitations stemming from insufficient spatiotemporal reasoning~\cite{manvi2024geollm,li2024reframing,chu2024timebench} and the generation of hallucinated or outdated content~\cite{huang2025survey,gao2023retrieval}, which restrict the spatiotemporal coherence of the produced travel plans.
Notably, beyond POI-level knowledge, the Web offers a wealth of up-to-date travel documents that encapsulate real-life experience and collective wisdom into practical tourist trajectories, \ie the sequences of POIs interconnected by spatiotemporal logic, which are neglected by current studies.
To address this gap, advances in Retrieval-augmented Generation (RAG) approaches~\cite{fan2024survey} can enable LLM agents to integrate such trajectory-level knowledge represented in Web documents, which provide substantial spatiotemporal-aware insights for travel planning.

In this paper, we introduce a new travel planning benchmark, \textit{TP-RAG}, to investigate whether retrieval-augmented LLM agents can effectively leverage trajectory-level knowledge to produce spatiotemporally coherent travel plans.
Our benchmark comprises a dataset grounded in the real-world search engine, featuring high-quality data sources.
It is designed to develop and evaluate LLM agents in generating spatiotemporal coherent travel plans, adhering to user queries and utilizing relevant POI and trajectory information.
Since privacy concerns, in our dataset, we enclose the verbalized trajectories extracted from the newest Web documents instead of using full document content.
Totally, our dataset includes 2,348 travel queries, 85,575 geotagged POIs and 18,784 tourist trajectory references.
Unlike prior datasets, our dataset incorporates query-customized latest spatiotemporal attributes into POI information, enabling time-adaptive and spatiotemporal-aware planning.
In addition, the inclusion of tourist trajectories encourages retrieval-augmented LLM agents to employ the vast repository of Web knowledge for plan enhancements.
The comparison between \textit{TP-RAG} and existing benchmarks is illustrated in \figref{figure:motivation}.

Based on our benchmark, we evaluate various LLM-based travel planning methods.
The results reveal notable limitations of advanced LLM agents which are constrained by internal knowledge, while highlighting promising prospects for the utilization of Web-based tourist trajectories in spatiotemporal travel planning.
Our in-depth analysis further uncovers concerns regarding the universality and robustness of retrieval-augmented travel planning approaches.
To address these issues, we propose \textit{EvoRAG}, a LLM-based evolutionary framework that iteratively optimizes travel plans through population-based selection, crossover and mutation of varied trajectory knowledge.
It effectively blends the merits of divergent retrieved knowledge and agents’ intrinsic planning capacity, while alleviating the impact of noisy information, the superiority of which is demonstrated by our experiments.

Our main contributions are three-fold:
(1) \textit{TP-RAG}, the first travel planning benchmark for retrieval-augmented and spatiotemporal-aware travel planning, using around 1 billion GPT-4o tokens for dataset construction.
(2) Extensive experiments (\ie over 5,000 A800-80G GPU hours) with various travel planning methods in different evaluation dimensions, showcasing both the opportunities and challenges of incorporating trajectory-level knowledge.
(3) A simple yet effective method, \textit{EvoRAG}, that further counters the limitations of retrieval-augmented travel planning.

\section{Related Work}

\subsection{Benchmarks of LLM-based Travel Planning}

To assess the capabilities of LLM agents in complex and realistic planning tasks, recent benchmarks have proliferated in travel planning, which stands out as a significant domain.
One line of research, into which our study falls, delves into the LLM-centric travel planning~\cite{xie2024travelplanner,singh2024personal,zhang2024ask,chaudhuri2025tripcraft}.
For example, TravelPlanner~\cite{xie2024travelplanner} investigates long-horizon travel planning in multi-constraint scenarios.
Beyond relying solely on LLMs, another line of benchmarks examine hybrid approaches that leverage LLMs for natural language interpretation paired with symbolic solvers to ensure solution validity through formal verification~\cite{hao2024large,de2024trip,shao2024chinatravel}. 
Despite progress, these benchmarks fail to incorporate sufficient fine-grained spatiotemporal contexts into planning, and are confined to agents' internal reasoning processes, which hinder the real-world deployment.

\subsection{LLM Agent for Travel Planning}

Travel planning remains an intricate problem involving intent comprehension, information seeking, and long-horizon planning.
To automate this task, current research on LLM travel agents bifurcates into two paradigms: LLM-driven and hybrid. 
LLM-driven approaches seek to enhance LLM agents' intrinsic planning capacities via advanced techniques, such as multi-agent collaboration which achieves coordination among LLM specialists~\cite{xie2024human,zhang2025planning} and LLM-based optimization that iteratively refines the quality of travel schedules~\cite{yuan2024evoagent,lee2025evolving}.
In contrast, hybrid approaches tackle LLM agents' limitations through the integration of computational planning modules, such as route optimizers~\cite{tang2024itinera}, heuristic POI selection algorithms~\cite{chen2024travelagent}, and symbolic solvers~\cite{ju2024globe}.

\subsection{Retrieval-Augmented Generation}

Retrieval-Augmented Generation (RAG) systems have emerged as pivotal solutions for enhancing LLMs with external knowledge~\cite{fan2024survey}. 
Recent research efforts have been devoted to building reliable benchmarks to evaluate RAG performance, emphasizing two assessment aspects: retrieval efficacy (\eg relevance, utility)~\cite{lyu2025crud,saad2024ares} and generation quality (\eg accuracy, coherence)~\cite{chen2024benchmarking,qi2024long2rag}, with our study concentrating on the latter.
Beyond open-domain scenarios, while some studies have demonstrated success in specific domains such as medical, legal and financial fields~\cite{xiong2024benchmarking,pipitone2024legalbench,wang2024omnieval}, travel applications remain nascent, restricted in POI-level tasks such as question-answering (QA)~\cite{song2024travelrag,yu2025spatial} and city or POI recommendation~\cite{banerjee2024enhancing,qi2024rag}.
And there is a lack of benchmarks for travel planning tasks that require real-time knowledge integration and multi-objective resolution. 

\section{TP-RAG}

\begin{figure*}[ht]
\centering
\includegraphics[width=\textwidth]{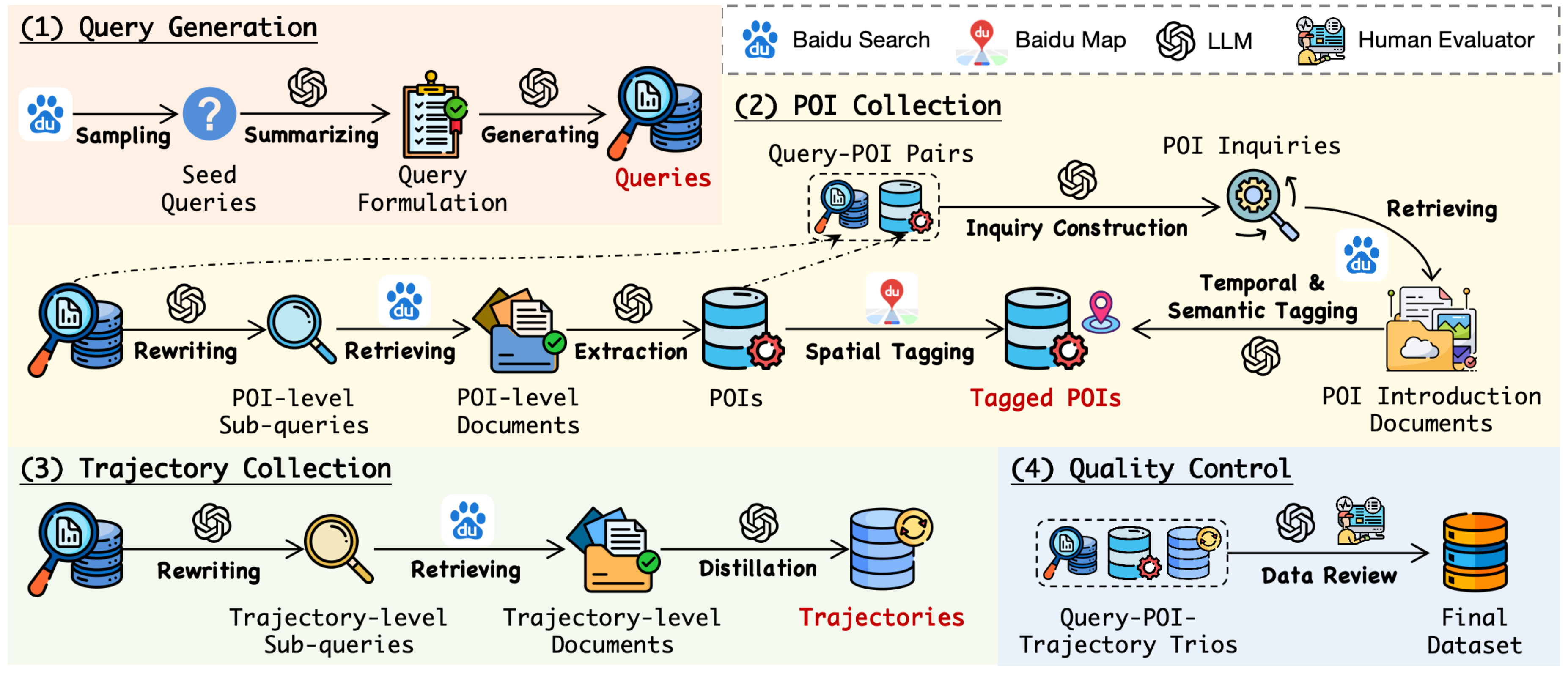}
\vspace{-20pt}
\caption{Dataset construction pipeline.}
\vspace{-18pt}
\label{figure:main}
\end{figure*}


\subsection{Background}
\label{section:problem}

We focus on generating single-city, multi-day travel plan consisting of attraction POIs, with some critical definitions delineated below:

\noindent\textbf{Query}.
A travel query $q$ is articulated by the user in natural language, and comprises significant elements such as the city name, travel duration, and personalized travel constraints.

\noindent\textbf{Point of Interest (POI)}.
A Point of Interest (POI) $p=(p_n, p_s, p_t, p_d)$ refers to a specific location that holds significance or interest for travelers.
Its attributes include the POI name $p_n$, spatial details $p_s$ (\eg address, geocoordinates), temporal references $p_t$ (\eg opening hours, recommended visit time, expected visit duration), and POI's semantic description $p_d$.
The candidate POIs dependent on the query $q$ are defined as $P^q=\{p^q_i\}_{i=1}^{|P^q|}$.

\noindent\textbf{Trajectory}.
A tourist trajectory is extracted from the retrieved travel-related Web document, and is denoted as $t=(p_{n,1}, p_{n,2},...,p_{n,|t|})$ consisting only POI names.
The trajectories relevant to the query $q$ are defined as $T^q=\{t^q_i\}_{i=1}^{|T^q|}$.

\begin{prob}
\textbf{Travel Planning}. 
Given a user query $q$, query-dependent POI candidates $P^q=\{p^q_i\}_{i=1}^{|P^q|}$, and query-relevant trajectories $T^q=\{t^q_i\}_{i=1}^{|T^q|}$, 
the travel plan $I=[(p^q_{n,1},t_{s,1},t_{e,1}),...,(p^q_{n,|I|},t_{s,|I|},t_{e,|I|})]$ is derived by LLM agents via $I=A(q,P^q,T^q)$, where $t_s$ and $t_e$ denote the scheduled start and end times of the POI visit, respectively.
\end{prob}

\subsection{Dataset Construction}
\label{section:data}

As illustrated in \figref{figure:main}, the construction pipeline of our dataset comprises the following steps:
\textit{(1) query generation}; \textit{(2) POI collection}; \textit{(3) trajectory collection}; culminating in a \textit{(4) quality control} procedure, which are detailed below.

\noindent\textbf{Query Generation.} 
Unlike previous works that rely on simulated queries, we sample the latest realistic queries from the data sessions in the Baidu search engine\footnote{\url{https://www.baidu.com}}, one of China’s largest search engines.
These queries are characterized by their conciseness, effectively expressing users' tourist needs through brief statements, such as \textit{'plan a 3-day trip to Beijing'}.

Based on the collected seed queries, we curate the LLM to establish a standard for query formulation, which is identified by several fundamental elements: \textit{(1) city name, (2) travel duration, and (3) personal constraints} that reflect user preferences.
For the destination city, we select the most popular cities in China, given their wealth of tourist resources and public travel narratives.
Regarding travel duration, we divide our queries into short-, medium- and long-term categories based on the number of travel days.
Depending on the presence of personalized constraints, our queries can also be categorized into generic and personal types.
To ensure the diversity of personal queries, we further classify them by constraint types, encompassing seasonal preferences, holiday-specific requirements, POI category restrictions, traveler demographics, and trip compactness parameters.
Refer to \appref{appendix:query} for more details of the query data.
This formulation systematically tests LLM agent’s capabilities in handling different user focuses and planning horizons. 
In accordance with the standard, we engage the LLM to generate a practical and extensive query dataset.

\noindent\textbf{POI Collection.}
To gather candidate attraction POIs and their associated spatiotemporal attributes for each query, we resort to LLMs augmented by search engines.
Initially, we exploit a query rewriting strategy~\cite{ma2023query}, whereby the LLM reformulates the original queries into POI-level sub-queries tailored for attraction recommendations.
The Baidu search engine is then utilized to retrieve pertinent documents for each sub-query.
Given that the documents generally contain a lot of irrelevant snippets, we apply the LLM to extract POI information exactly from each document.

Furthermore, to annotate fine-grained spatiotemporal information and contextual semantics for the extracted POIs, we combine two processes: 
\textit{(1) Spatial Tagging:}
We leverage the Baidu Map platform\footnote{\url{https://lbsyun.baidu.com/}} to enrich POIs with comprehensive spatial information including addresses and geocoordinates (\ie latitude and longitude). 
This process also facilitates the standardization of POIs into unified names, thereby preventing noisy and hallucinated POIs.
\textit{(2) Temporal and Semantic Tagging:}
For each query-POI pair (\eg \textit{Plan a spring trip in Beijing - the Great Wall}), we create query-tailored POI inquiries (\eg \textit{What's the recommended travel time period of the Great Wall in spring?}) to search for related documents about this POI, representing the real-time needs (\eg Spring trip).
Subsequently, the retrieval-augmented LLM is employed to summarize nuanced temporal and semantic insights, entailing opening hours, recommended visit times, expected visit durations, and semantic POI descriptions of the POIs, which foster agents' awareness of spatiotemporal coherence and POI distinctiveness in travel planning.

\noindent\textbf{Trajectory Collection.}
To collect the trajectory-level knowledge, we first retrieve up-to-date documents that pertain to real-life travel experiences relevant to the given query.
In order to address user privacy concerns, full documents are not disclosed in our dataset. 
Instead, we implement a LLM-based desensitization process that distills tourist trajectories from the original lengthy documents.
To maintain the integrity of the data, we instruct the LLM to refrain from answering if the document lacks plausible trajectory-level information.
Additionally, to emulate the retrieval noise commonly encountered in real-world scenarios, we retain the disturbances present within these trajectories.
As outlined in the quality control stage below, we also prepare a denoised version of trajectory data for further in-depth analysis in \secref{section:in-depth}.

\noindent\textbf{Quality Control.}
To ensure the quality of the generated dataset, we employ both LLM and human evaluators to review all \textit{query-POI-trajectory} data instances and eliminate the noise in both POIs and trajectories.
Instances that lack sufficient trajectory references are discarded.
This procedure guarantees the feasibility of our retrieval-augmented spatiotemporal travel planning task.

In conclusion, our dataset consists of 2,348 travel queries, including 115 generic queries and 2,233 personal queries. 
This dataset is associated with 85,575 attraction POIs (averaging 36.45 POIs per request) derived from 5,018 unique attractions, and 18,784 retrieved trajectories (averaging 8 trajectories per request). 
Throughout the data construction process, we utilize GPT-4o~\cite{openai2024gpt4o}, and the prompts are presented in \appref{appendix:prompt_data_generation}.

\subsection{Evaluation}
\label{section:evaluation}

Beyond commonsense constraints, our evaluation system illuminates the nuanced aspects concerning spatiotemporal rationality and the semantic prominence of POIs. 
Unlike TripCraft~\cite{chaudhuri2025tripcraft}, we conceptualize spatiotemporal travel planning as a complex problem without unique optimal solutions, challenging the reliability of annotating ideal plans.
Following existing works~\cite{tang2024itinera}, our evaluation system integrates rule-based metrics alongside LLM-as-a-Judge techniques, emphasizing five critical evaluation dimensions: commonsense, spatial, temporal, POI semantic and query relevance.
This approach effectively circumvents costly annotations and mitigates evaluation biases, with metric descriptions provided below.

\noindent\textbf{Commonsense.}
Commonsense metrics measure whether the generated plan adheres to basic validity standards, including:
\textit{(1) Failure Rate (FR)}: the percentage of legitimate POIs without hallucination;
\textit{(2)Repetition Rate (RR)}: the frequency of POI repetition within the plan.

\noindent\textbf{Spatial.}
The spatial metric evaluates the route efficiency of the plan.
Specifically, we use \textit{Distance Margin Ratio (DMR)} to quantify the distance gap from the theoretically optimal route.

\noindent\textbf{Temporal.}
Temporal metrics assess the rationality of the scheduled visit periods of POIs, which embrace 
\textit{(1) Start Time Rationality (STR)}: whether the arranged arrival time for POI visit is appropriate;
\textit{(2) Duration Underflow Ratio (DUR)}: the extent to which the planned visit duration meets expectations;
\textit{(3) Time Buffer Ratio (TBR)}: the proportion of buffer time available throughout the plan, indicating the degree of tourist comfort.

\noindent\textbf{POI Semantic.}
The semantic metric examines popularity and distinctiveness of the selected POIs.
In particular, we design a \textit{POI Popularity (PP)} metric to measure the recall rate within the retrieved attraction leaderboard.

\noindent\textbf{Query Relevance.}
The relevance metrics focus on whether the user demands specified in the queries (\eg time-sensitive desires) are fulfilled, concerning two aspects:
\textit{(1) POI Relevance (PR)}: the alignment between planned POIs and the user query;
\textit{(2) Time Schedule Relevance (TSR)}: the pertinence of the arranged POI visit period and personal needs.

\begin{table*}[ht]
    \centering
    \vspace{-10pt}
    \caption{Main results (\%) of different methods on our dataset. The best strategies are marked in bold, while the second-best ones are underlined.}
    \vspace{-8pt}
    \resizebox{\linewidth}{!}{
    \begin{tabular}{l|cc|ccccccc|cccc|c}
    \toprule
    \textbf{Method} & \textbf{FR$\downarrow$} & \textbf{RR$\downarrow$} & \textbf{DMR$\downarrow$} & \textbf{DUR$\downarrow$} & \textbf{TBR$\uparrow$} & \textbf{STR$\uparrow$} & \textbf{PP$\uparrow$} & \textbf{PR$\uparrow$} & \textbf{TSR$\uparrow$} & $R_S\downarrow$ & $R_T\downarrow$ & $R_P\downarrow$ & $R_R\downarrow$ & $R_C\downarrow$ \\ 
    \hline
    \multicolumn{15}{c}{GPT-4o} \\ 
    \hline
    Direct & \textbf{0.32} & \textbf{0.00} & 67.92 & 3.03 & 22.01 & 77.22 & 50.82 & 80.51 & \underline{92.52} & 6.00 & 7.67 & 7.00 & \textbf{3.00} & 5.92 \\ 
    CoT & \underline{0.39} & \underline{0.01} & 69.4 & 2.78 & 22.13 & 76.96 & 50.09 & 79.92 & 91.99 & 8.00 & 8.00 & 10.00 & 6.50 & 8.12 \\ 
    Reflextion & 0.67 & 0.34 & 73.35 & 3.84 & 21.46 & 77.3 & 50.52 & 80.34 & \textbf{92.71} & 10.00 & 8.67 & 8.00 & \underline{3.50} & 7.54 \\ 
    MAC & 1.40 & 0.72 & 66.11 & 3.67 & \textbf{24.07} & 75.41 & 46.38 & \textbf{81.76} & 89.04 & 3.00 & 7.00 & 11.00 & 6.00 & 6.75 \\ 
    MAD & 0.68 & 0.07 & 74.95 & 3.68 & 20.67 & 77.14 & 50.52 & 79.53 & 92.4 & 11.00 & 9.67 & 8.00 & 6.50 & 8.79 \\ 
    \hdashline[0.5pt/1pt]
    RAG($M$=8) & 2.05 & \underline{0.01} & 68.37 & 2.57 & \underline{23.82} & 77.19 & \underline{58.00} & \underline{80.55} & 91.58 & 7.00 & 4.67 & \underline{2.00} & 5.00 & 4.67 \\ 
    RAG($M$=4) & 2.08 & 0.02 & 67.75 & 2.55 & 23.71 & 77.35 & 56.11 & 80.53 & 91.67 & 5.00 & 3.67 & 4.00 & 4.50 & \underline{4.29} \\ 
    RAG($M$=1) & 2.43 & 0.04 & \underline{66.05} & 2.47 & 22.59 & 77.44 & 53.38 & 80.1 & 91.5 & \underline{2.00} & 4.00 & 6.00 & 8.50 & 5.12 \\ 
    RAG+Extr.($M$=4) & 1.91 & 0.02 & 66.99 & \textbf{2.41} & 23.5 & \textbf{77.87} & 56.82 & 80.31 & 91.69 & 4.00 & \textbf{2.00} & 3.00 & 6.00 & \textbf{3.75} \\ 
    RAG+Extr.($M$=1) & 2.72 & 0.06 & \textbf{65.76} & \underline{2.42} & 22.95 & \underline{77.79} & 53.85 & 80.39 & 91.64 & \textbf{1.00} & \underline{3.00} & 5.00 & 6.00 & \textbf{3.75} \\ 
    RAG+Abst. & 3.20 & 0.02 & 69.49 & 2.64 & 22.23 & 76.66 & \textbf{59.15} & 79.36 & 90.67 & 9.00 & 7.67 & \textbf{1.00} & 10.50 & 7.04 \\ 
    \hline
    \multicolumn{15}{c}{Qwen2.5-72B-Instruct} \\ 
    \hline
    Direct & \textbf{0.38} & \textbf{0.03} & 71.67 & 6.49 & 24.82 & 78.33 & 48.53 & 79.68 & 92.86 & 8.00 & \underline{5.00} & 8.00 & 9.00 & 7.50 \\ 
    CoT & \underline{0.42} & \underline{0.04} & 70.51 & 6.64 & 24.66 & 78.77 & 47.09 & 80.12 & 93.16 & 7.00 & 5.33 & 10.00 & 8.00 & 7.58 \\ 
    Reflextion & 2.39 & 1.62 & 85.38 & 8.08 & 25.37 & 77.54 & 49.15 & 79.65 & 92.14 & 10.00 & 7.67 & 7.00 & 10.00 & 8.67 \\ 
    MAC & 1.10 & 2.21 & 70.08 & \textbf{5.74} & 23.65 & 76.48 & 43.3 & \underline{81.28} & 90.00 & 6.00 & 7.33 & 11.00 & 6.50 & 7.71 \\ 
    MAD & 3.30 & 1.49 & 87.47 & 9.46 & \textbf{26.53} & 77.75 & 47.49 & 80.23 & 91.21 & 11.00 & 7.00 & 9.00 & 9.00 & 9.00 \\ 
    \hdashline[0.5pt/1pt]
    RAG($M$=8) & 3.41 & 0.11 & 69.39 & \underline{6.35} & 23.48 & 78.4 & \underline{55.15} & \textbf{81.67} & 93.29 & 5.00 & 6.00 & \underline{2.00} & \underline{3.00} & 4.00 \\ 
    RAG($M$=4) & 3.15 & 0.11 & 68.77 & 6.54 & 24.06 & \textbf{79.08} & 53.62 & 81.26 & \textbf{93.86} & 3.00 & \underline{5.00} & 4.00 & \textbf{2.00} & \textbf{3.50} \\ 
    RAG($M$=1) & 2.58 & 0.09 & \textbf{68.07} & 6.89 & 25.27 & 78.48 & 51.62 & 81.07 & 93.4 & \textbf{1.00} & 5.33 & 6.00 & 4.50 & 4.21 \\ 
    RAG+Extr.($M$=4) & 3.46 & 0.16 & 69.03 & 6.49 & 24.33 & \underline{79.03} & 54.79 & 81.21 & \underline{93.81} & 4.00 & \textbf{4.33} & 3.00 & \underline{3.00} & \underline{3.58} \\ 
    RAG+Extr.($M$=1) & 3.16 & 0.20 & \underline{68.29} & 6.73 & \underline{25.59} & 78.33 & 52.36 & 80.92 & 93.42 & \underline{2.00} & \underline{5.00} & 5.00 & 4.50 & 4.12 \\ 
    RAG+Abst. & 3.78 & 0.17 & 72.4 & 7.32 & 25.11 & 78.04 & \textbf{56.14} & 80.5 & 93.29 & 9.00 & 7.33 & \textbf{1.00} & 6.00 & 5.83 \\
    \hline
    \multicolumn{15}{c}{DeepSeek-R1} \\ 
    \hline
    Direct & \textbf{0.55} & \textbf{0.01} & 68.78 & \textbf{3.07} & 22.94 & 76.68 & 50.27 & 80.68 & 91.7 & 7.00 & 4.67 & 7.00 & 6.50 & 6.29 \\ 
    \hdashline[0.5pt/1pt]
    RAG($M$=8) & 2.31 & 0.03 & \textbf{65.31} & 3.94 & \underline{23.30} & \underline{77.21} & \underline{53.87} & \textbf{82.47} & 92.66 & \textbf{1.00} & \underline{3.00} & \underline{2.00} & \textbf{2.00} & \textbf{2.00} \\ 
    RAG($M$=4) & 1.62 & 0.03 & 66.76 & 3.78 & \textbf{23.37} & 76.95 & 53.02 & \underline{82.00} & 92.45 & 5.00 & \textbf{2.67} & 4.00 & 4.00 & 3.92 \\ 
    RAG($M$=1) & \underline{1.23} & 0.04 & 66.42 & \underline{3.73} & 22.97 & \textbf{77.37} & 51.82 & 81.54 & \underline{92.89} & 3.00 & \underline{3.00} & 6.00 & 3.50 & 3.88 \\ 
    RAG+Extr.($M$=4) & 2.00 & 0.03 & 66.45 & 4.04 & 23.28 & 76.94 & 53.69 & \underline{82.00} & 92.66 & 4.00 & 4.33 & 3.00 & \underline{2.50} & 3.46 \\ 
    RAG+Extr.($M$=1) & 1.25 & 0.03 & \underline{66.19} & 3.76 & 23.23 & 76.94 & 51.95 & 81.66 & \textbf{93.02} & \underline{2.00} & 3.67 & 5.00 & \underline{2.50} & \underline{3.29} \\ 
    RAG+Abst. & 2.04 & \underline{0.02} & 67.57 & 4.96 & 23.16 & 76.63 & \textbf{54.28} & 80.5 & 92.6 & 6.00 & 6.33 & \textbf{1.00} & 6.00 & 4.83 \\ 
    \bottomrule
    \end{tabular}}
    \vspace{-15pt}
    \label{table:main}
\end{table*}

In \appref{appendix:metrics}, we further elaborate details about the metrics, and validate that LLM evaluators aligns with humans well.
To provide a more transparent depiction of the overall effectiveness, we supplement five rank-based metrics:
$R_S$, $R_T$, $R_P$, $R_R$ and $R_C$, which respectively denote the performance rank of methods from spatial, temporal, POI semantic and query relevance dimensions, and a comprehensive view averaging all aspects.
\section{Experiment}


\subsection{Baselines}
\label{section:baseline}

\noindent\textbf{Travel Planning Methods.}
We evaluate two categories of travel agents:
\textbf{(1) Ground-up Travel Agents} which include
\textit{Direct}, \textit{Chain of Thought (CoT)}~\cite{wei2022chain}, \textit{Reflextion}~\cite{shinn2023reflexion} and two multi-agent frameworks: 
\textit{Multi-Agent Collaboration (MAC)} which applies a divide-and-conquer solution~\cite{zhang2025planning}, and \textit{Multi-Agent Debate (MAD)} that facilitates a discussion session for plan refinement~\cite{ni2024planning};
\textbf{(2) Retrieval-augmented Travel Agents} implemented by the RAG strategy utilizing trajectory-level knowledge.
\textit{RAG($M$=$m$)} denotes the retrieval-augmented method using $m$ trajectories.
To test agents' capabilities for explicit context utilization, 
we consider two simple post-retrieval techniques~\cite{xu2024recomp}: extractive compression method \textit{RAG+Extr.($M$=$m$)} and abstractive compression method \textit{RAG+Abst.}, which intentionally purify the retrieved content for enhancements.

\noindent\textbf{Base Models.}
The core of agent-based planning methods lies in the LLM.
Therefore, we evaluate various advanced LLMs including GPT-4o~\cite{openai2024gpt4o}, Qwen2.5-72B-Instruct~\cite{yang2024qwen2}, LLaMA3.3-70B-Instruct~\cite{llama70b}, as well as DeepSeek-R1~\cite{guo2025deepseek}.
More details of the baselines and evaluation setups are provided in \appref{appendix:setup}.

\subsection{Main Results}
\label{section:main}

In this section, we discuss the performances of various methods and models on our benchmark as presented in \tabref{table:main}.
Due to the space limit, we leave the results on LLaMA3.3 in \appref{appendix:llama3.3}.
Our critical observations summarized below:

\noindent\textbf{Advanced LLM agents struggle with spatiotemporal travel planning.}
Cutting-edge strategies (\ie CoT, Reflextion and multi-agent techniques) underperform direct prompting in holistic ranking $R_C$, revealing complex task decomposition or answer reflection can lead to error accumulation and degeneration of spatiotemporal reasoning.

\noindent\textbf{Trajectory knowledge holds potential for enhanced travel planning.}
According to $R_C$, retrieval-augmented planning methods generally outperform ground-up ones, highlighting the value of trajectory knowledge.
The performance gain primarily stems from spatial and POI semantic dimensions, while some models (\eg GPT-4o) exhibit slight declines in temporal and relevance aspects, which can be attributed to agents’ limited capacity to resolve confusing and verbose contexts.

\noindent\textbf{Sophisticated methods compromise agents' commmonsense awareness.}
A notable increase of inaccessible or repetitive POIs is observed within both advanced LLM agents and retrieval-augmented ones, suggesting that overly complex methods tend to perplex LLM agents since long-context and complicated inputs.

\noindent\textbf{Reasoning-optimized LLMs are not as desired in travel planning.}
Despite architectural advances, specialized reasoning models like Deepseek-R1 fail to show a remarkable advantage over other foundational models in spatiotemporal travel planning, even lagging in the temporal dimension.

\noindent\textbf{Retrieval-augmented planning methods lack stability.}
Knowledge richness and post-processing techniques show inconsistent benefits across different base models, indicating context sensitivity.
We leave the in-depth analysis in \secref{section:in-depth}.

\noindent\textbf{Performances in distinct metrics are inconsistent.}
No single solution dominates all evaluation dimensions.
Trade-offs between different metrics are commonly observed in most cases, illustrating the complexity of our multi-objective spatiotemporal travel planning task.

Refer to \appref{appendix:sub-query} for results on sub-datasets across various query categories, which generally align with our full-scale experiments in \tabref{table:main}.

\subsection{In-depth Analysis}
\label{section:in-depth}

In this section, we conduct a thorough examination of retrieval-augmented methods in terms of universality, planning mechanisms and robustness.
We implement our analysis experiments using Qwen2.5-72B-Instruct as an example.

\noindent\textbf{Universality Analysis.} 
To explore whether the retrieved knowledge is always necessary, we calculate the win rates of retrieval-augmented methods compared to the Direct baseline, achieving an average of 87.41\% across seven spatiotemporal metrics. 
This demonstrates that trajectory references are not universally efficacious, as evidenced by a 12.59\% failure gap. 
Refer to \appref{appendix:universality} for the detailed statistics of win rates.

\begin{figure}[ht]
    \centering
    \vspace{-15pt}
    \subfloat[Similarity by trajectory \\position.]{\includegraphics[width=0.49\linewidth]{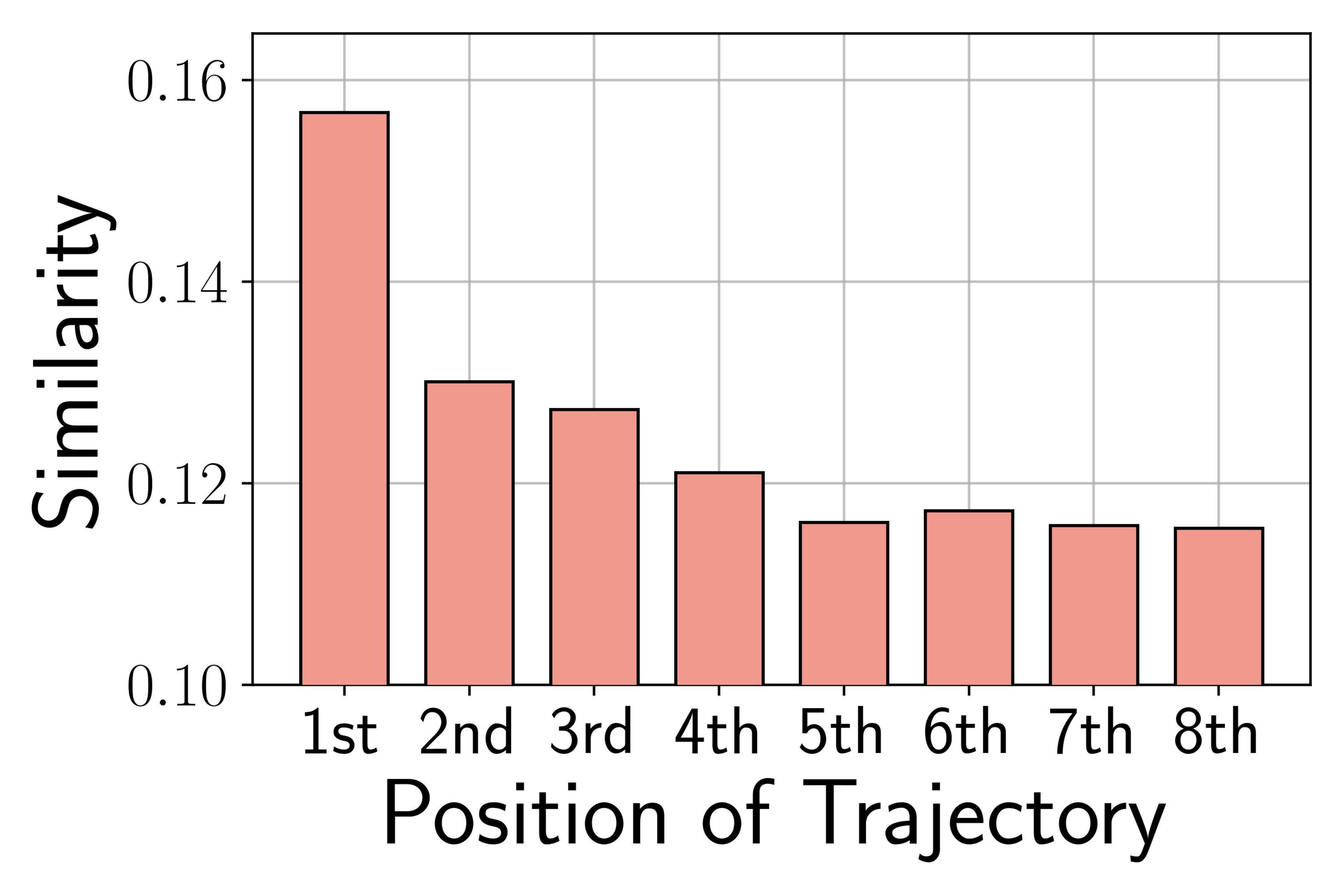}\label{figure:sim_dist_position}}
    \subfloat[Similarity by trajectory similarity rank.]{\includegraphics[width=0.49\linewidth]{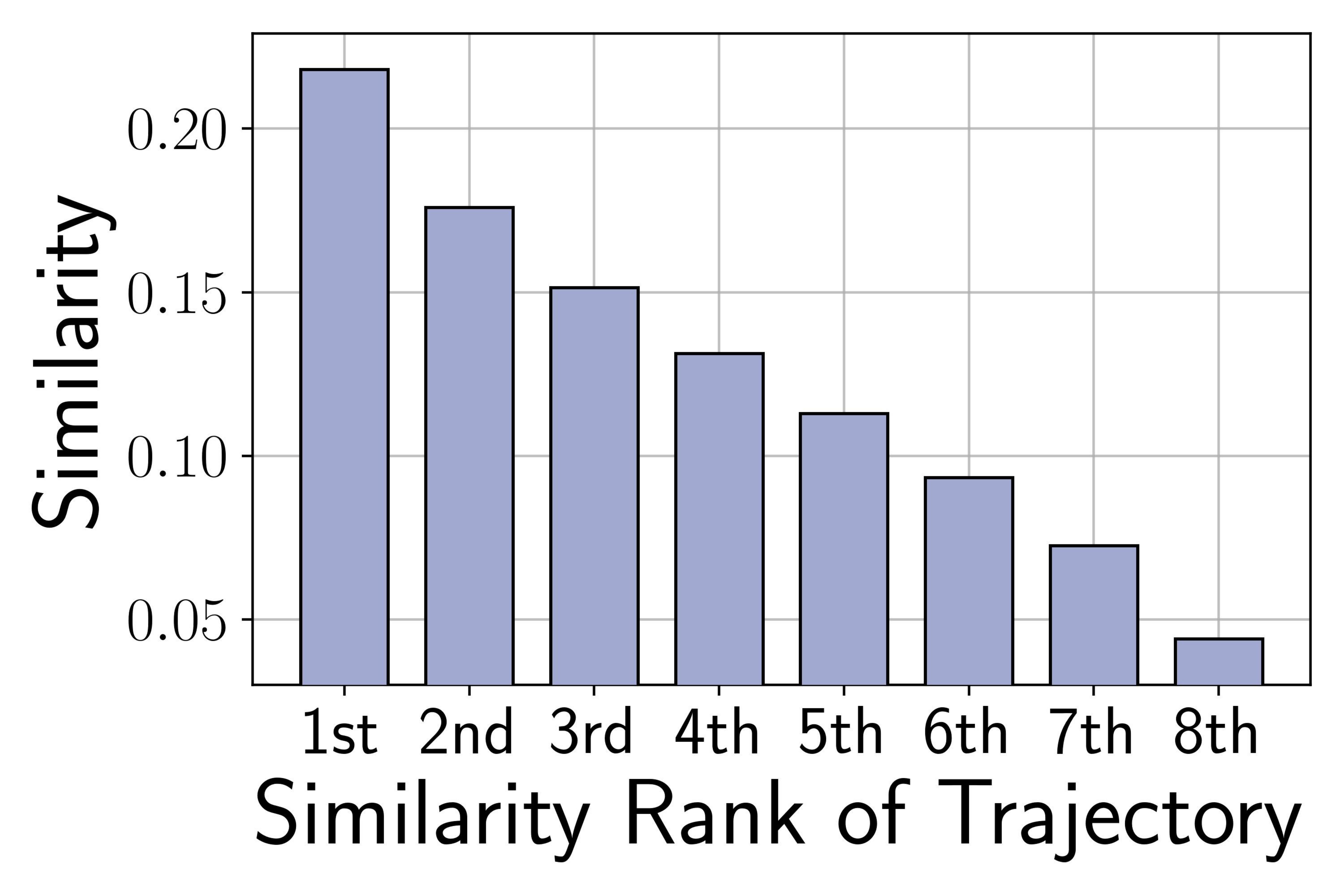}
    \label{figure:sim_dist_rank}}
    \vspace{-8pt}
    \caption{Similarities between plans and trajectories.}
    \vspace{-10pt}
    \label{figure:sim_dist}
\end{figure}

\noindent\textbf{Utilization Analysis.}
To delve into how LLM agents utilize trajectory knowledge, we employ the similarity $\sigma(s,t)$ (detailed in \appref{appendix:utilization}) between the plan $s$ generated by RAG($M$=8) and associated trajectories $T$, as a proxy indicating the extent of LLMs' utilization of the trajectory.
\figref{figure:sim_dist_position} unravels the similarities between the plan and trajectories at different positions within the prompt context, indicating a preference for information at the beginning, which is more relevant to the travel query.
When we reorder the distribution by descending reference similarity, the results in \figref{figure:sim_dist_rank} elucidate that LLM agents tend to selectively use several references rather than assimilate all trajectories.

\begin{figure}[ht]
    \centering
    \vspace{-15pt}
    \subfloat[Kendall Tau coefficients.]{\includegraphics[width=0.49\linewidth]{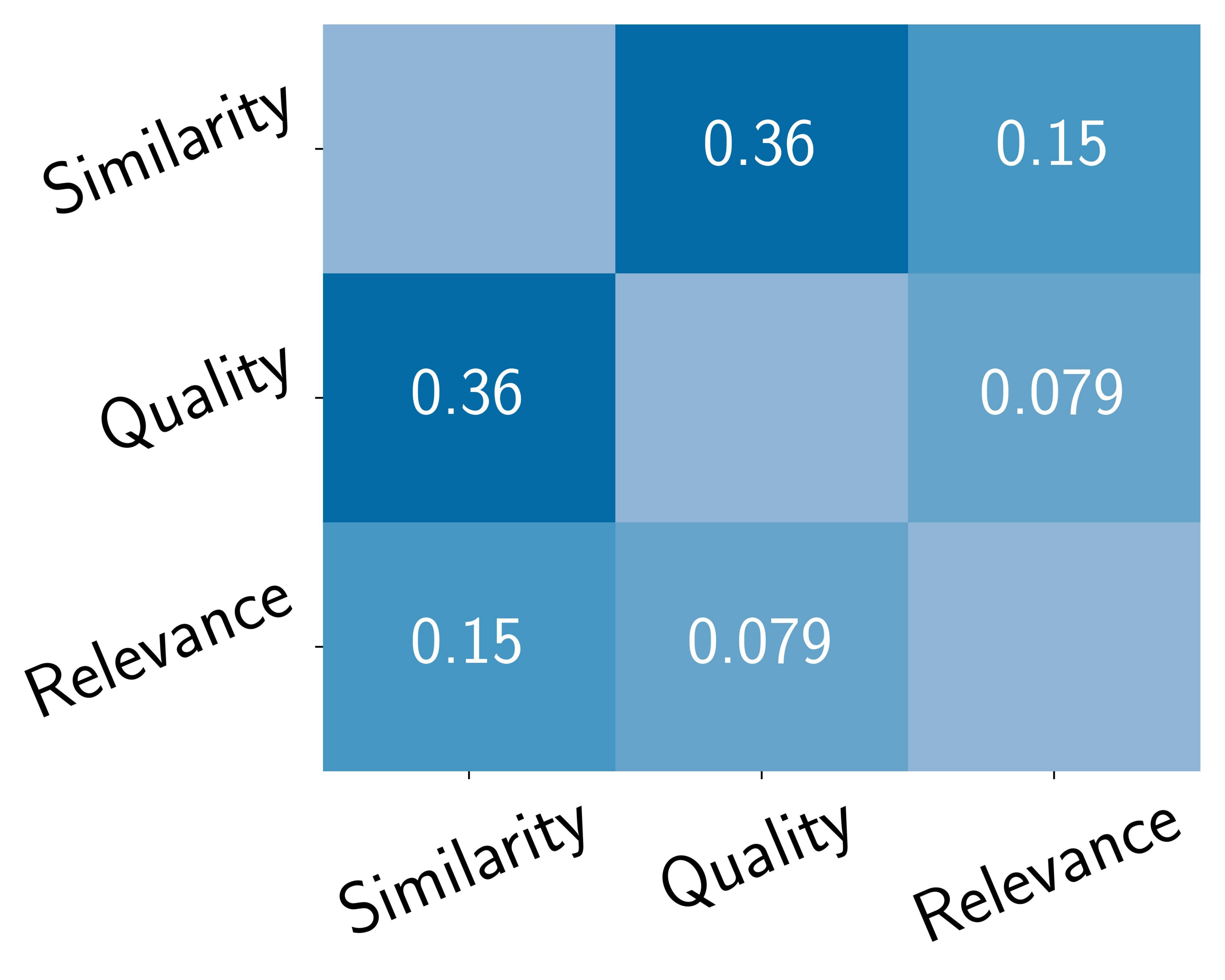}\label{figure:corr_tau}}
    \subfloat[Jaccard coefficients.]{\includegraphics[width=0.49\linewidth]{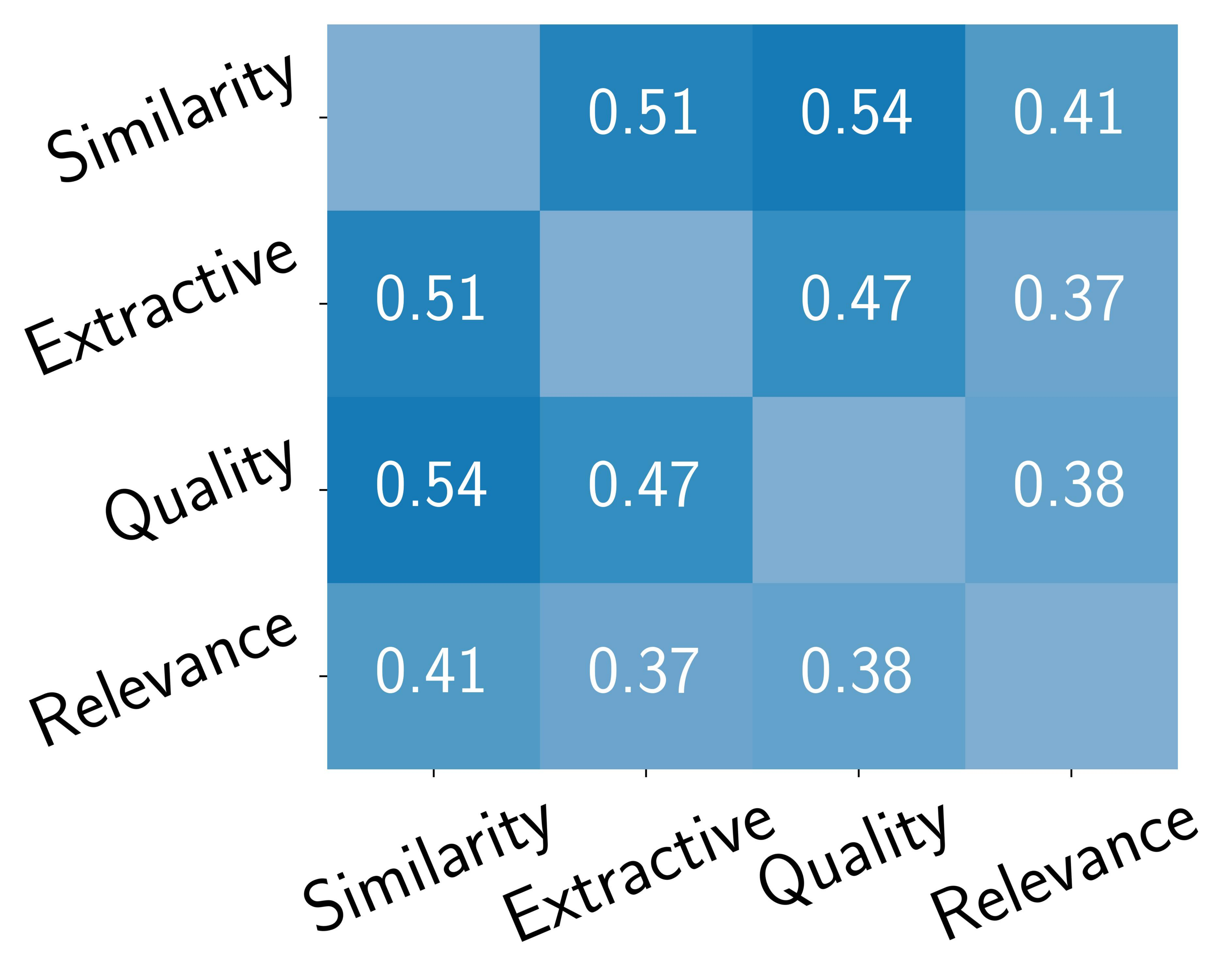}\label{figure:corr_jaccard}}
    \vspace{-8pt}
    \caption{Correlation analysis.}
    \vspace{-10pt}
    \label{figure:corr}
\end{figure}

Furthermore, to probe why retrieval-augmented methods have positive effects, we analyze the ordinal correlations among three variables: (1) extent of utilization for the trajectory (\ie similarity $\sigma$), (2) quality of the trajectory (detailed in \appref{appendix:utilization}), (3) query relevance of the trajectory.
\figref{figure:corr_tau} reveals a significant consistency between similarities and quality, demonstrating LLM agents' abilities to identify high-quality knowledge for travel planning.
Besides, we contrast implicit and explicit extractive utilization methods (\ie RAG($M$=8) and RAG+Extr.($M$=4)) in \figref{figure:corr_jaccard}, as detailed in \appref{appendix:utilization}.
We discern that these two extractive ways are distinct in reference utilization with only 0.51 Jaccard consistency, while the implicit method gains a slight edge in quality alignment (\ie 0.54 versus 0.47).

\begin{figure}[ht]
    \centering
    \vspace{-5pt}
    \includegraphics[width=\linewidth]{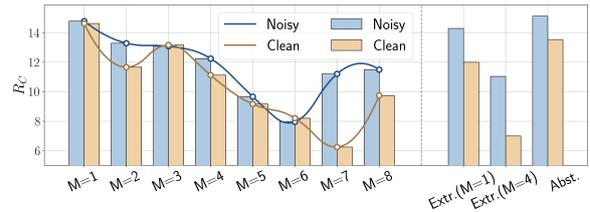}
    \vspace{-22pt}
    \caption{The sensitivity analysis of retrieval-augmented methods with different retrieval quantity, based on noisy and clean trajectory knowledge.}
    \vspace{-10pt}
    \label{figure:sensitivity}
\end{figure}

\noindent\textbf{Sensitivity Analysis.}
To investigate the sensitivity of retrieval-augmented planning methods with respect to the quantity and quality of retrieval data, we test retrieval-augmented strategies with varying number of trajectory references, and compare the results of using noisy and denoised trajectories (as mentioned in \secref{section:data}).
The analysis results in \figref{figure:sensitivity} depict that the $R_C$ performance reaches its peak by integrating 6 or 7 trajectories.
Reducing the volume of knowledge (\ie $M<6$) leads to a significant decline in efficacy, uncovering the necessity of diverse reference knowledge, and confirming the non-uniqueness of the travel planning problem.
Conversely, excessive knowledge (\ie $M=8$) also undermines the effectiveness, resulting in difficulties assimilating diverging references and filtering out low-quality insights, despite their relevance to user queries.
Moreover, the performance comparison between noisy and clean (\ie denoised) trajectories highlights the negative impact of the POI noise on the spatiotemporal validity, particularly for post-processing methods.

\begin{tcolorbox}[
    colback=gray!10, 
    colframe=black, 
    boxsep=3pt, 
    boxrule=0.5pt,
    arc=3pt, 
    left=0pt, 
    right=0pt, 
]
\fontsize{9.8}{0}
\textbf{Takeaways.}
\textit{
(1) LLM agents effectively enhance the spatiotemporal rationality of travel planning aided by retrieved trajectories. 
(2) LLM agents utilize trajectories extractively, in a manner that aligns with reference quality.
(3) Retrieval-augmented agents face challenges in uniformly promoting spatiotemporal travel planning across all queries and evaluation metrics, and robustly integrating conflicting and noisy references.
}
\end{tcolorbox}

\section{EvoRAG}

\begin{figure}[ht]
    \centering
    \vspace{-8pt}
    \includegraphics[width=\linewidth]{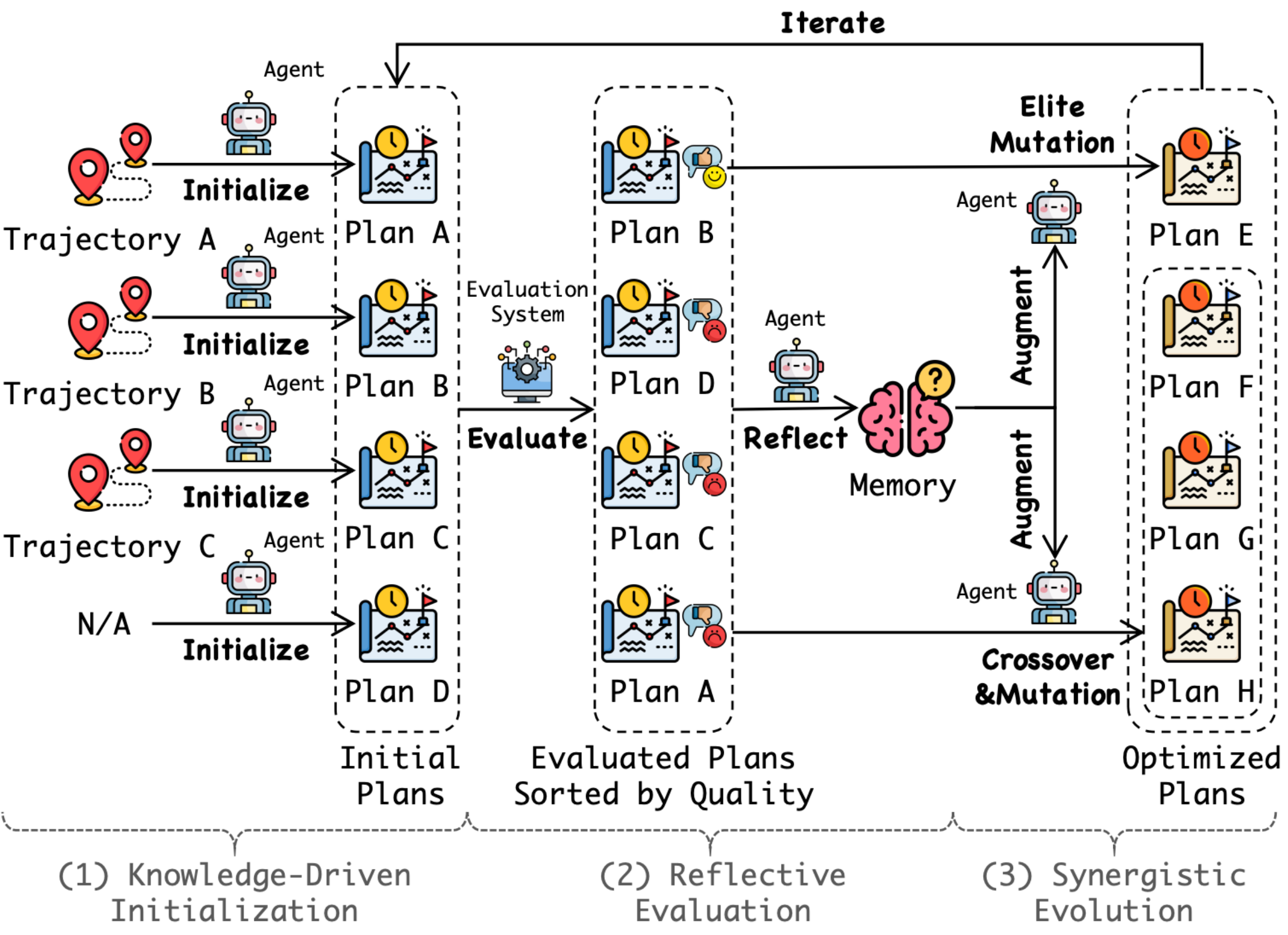}
    \vspace{-15pt}
    \caption{The workflow of EvoRAG.}
    \vspace{-10pt}
    \label{figure:method_workflow}
\end{figure}

\noindent\textbf{Method.}
To address the aforementioned issues, we propose \textit{EvoRAG}, a knowledge-evolution optimization framework, as illustrated in \figref{figure:method_workflow}. 
It includes three procedures: \textit{(1) knowledge-driven initialization}; \textit{(2) reflective evaluation}; and \textit{(3) synergistic evolution}. 
Based on initialization, the plans are iteratively optimized by alternating cycles of evaluation and evolution.
For a detailed algorithm workflow, see \appref{appendix:method}.

\noindent\textit{(1) Knowledge-Driven Initialization.}
To incorporate divergent retrievable knowledge, we curate LLM agents to generate initial plans, individually based on $|T|$ tourist trajectories.
These plans are independent and poised for evolutionary optimization, bypassing the deficiencies of LLM agents in assimilating discordant references.
Additionally, we include an initial plan derived solely from LLM agents' intrinsic knowledge (implemented by the Direct baseline) to counter the limited universality of retrieval-augmented methods.

\noindent\textit{(2) Reflective Evaluation.}
In each iteration, to ensure that LLM agents can comprehend the optimization objectives, we evaluate the quality of plans across various metrics (detailed in \secref{section:evaluation}).
To further facilitate LLM agents in discerning how to improve solutions, we encourage them to deliberately analyze the evaluation results of different plans and reflect on their strengths and weaknesses.
This self-aware reflection process is managed by a memory module, which is iteratively updated to maintain the expertise of LLM agents learned from optimization and evaluation experiences.

\noindent\textit{(3) Synergistic Evolution.}
Based on evaluation feedback and reflective memory, the best $\alpha$ proportion of plans are retained for optimization by LLM agents, termed elite mutation.
Furthermore, to unify the advantages of trajectory knowledge from distinct perspectives, we selectively synthesize the plans (\ie crossover), avoiding knowledge isolation stemming from the separate initialization.
Specifically, we repeatedly select dissimilar plans to perform crossover and mutation to ensure diversity, until we obtain $(|T|+1)$ solutions. 

The framework enables the LLM agent to focus on plan optimization utilizing multifarious external and internal knowledge, while becoming more robust to noisy and nonsensical information. 
The prompts used are shown in \appref{appendix:prompt_method}.


\begin{figure}[ht]
    \centering
    \vspace{-10pt}
    \includegraphics[width=0.58\linewidth]{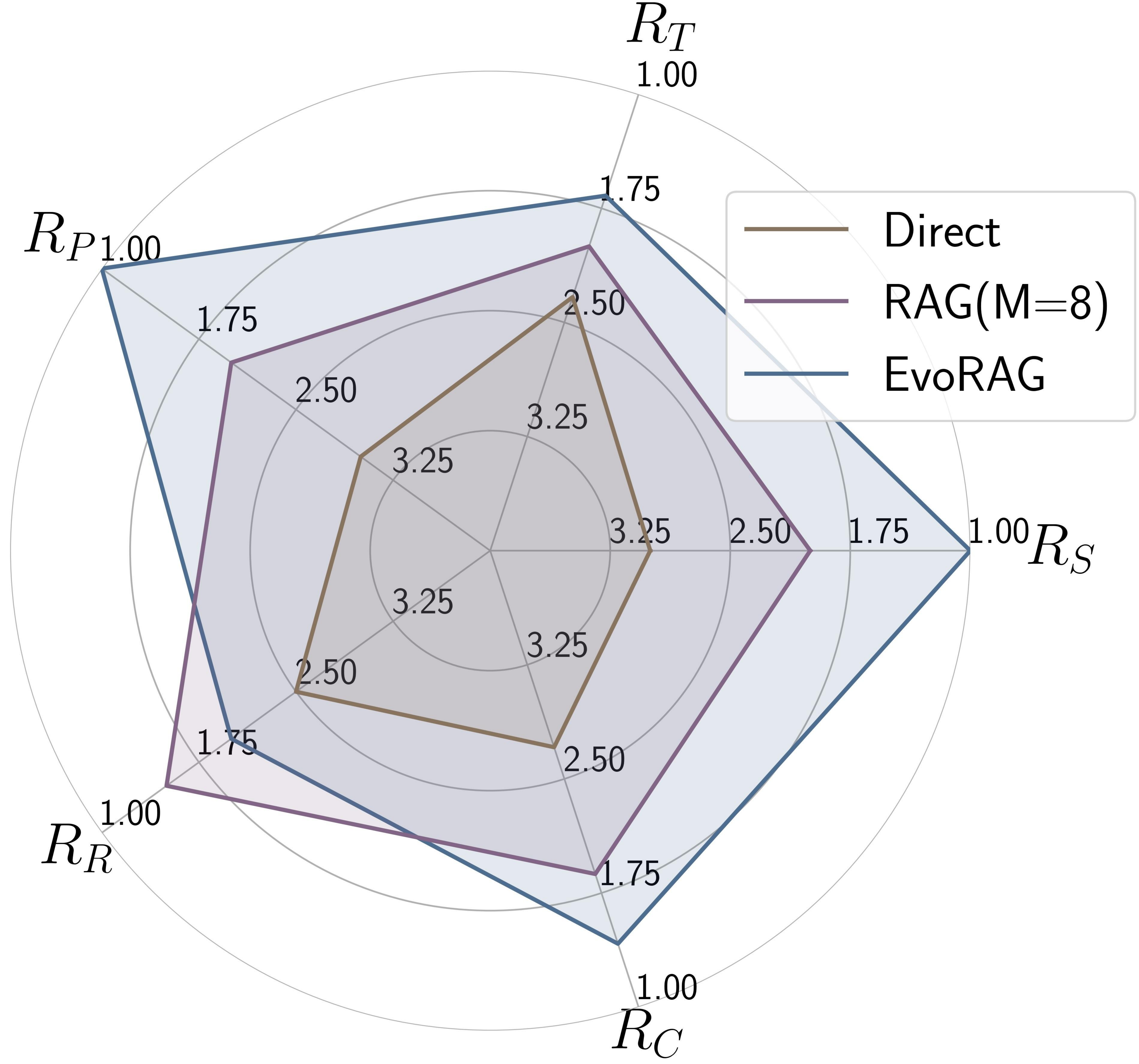}
    \vspace{-8pt}
    \caption{Comparison of EvoRAG and baselines.}
    \vspace{-10pt}
    \label{figure:method_radar}
\end{figure}

\noindent\textbf{Comparison Results.}
Based on Qwen2.5-72B-Instruct model, \textit{EvoRAG} generally surpasses both ground-up and retrieval-augmented planning methods across almost all dimensions (except for query relevance $R_R$), as illustrated in \figref{figure:method_radar}.
\textit{EvoRAG} also achieves notable reductions in commonsense failures with a 0.4\% POI failure rate (FR) and a 0.06\% POI repetition rate (RR), which are comparable to the \textit{Direct} baseline.
The detailed setups and complete experimental results are provided in \appref{appendix:method}.
This further underscores the proficiency of integrating trajectory-level knowledge with LLM-based optimization, paving the way for future advancements in LLM-driven travel agents.

\section{Conclusion}

This work investigates the role of online knowledge in improving LLM agents for spatiotemporal travel planning. 
We introduce \textit{TP-RAG}, a benchmark that integrates spatiotemporal POI characteristics and trajectory-level Web knowledge.
Experiments across advanced LLMs demonstrate that retrieval-augmented methods generally improve the fine-grained quality of plans, but they are not universally effective or robust.
Our proposed \textit{EvoTravel} framework counters these issues through evolutionary optimization, achieving state-of-the-art results by balancing divergent knowledge. 
This sets the stage for developing more powerful travel agents with exceptional spatiotemporal awareness.
\section*{Limitations}

\noindent\textbf{Planning Setup.}
The proposed \textit{TP-RAG} aims to examine the capabilities of LLM agents in spatiotemporal travel planning by utilizing online knowledge. 
Therefore, we specifically focus on attraction planning without the use of tools for agents. 
We believe that our benchmark can be expanded to include realistic planning scenarios encompassing meals, accommodations, and transportation, and enabling adaptive, tool-based information retrieval.

\noindent\textbf{Query Scenario.}
Our dataset is limited to search scenarios which feature concise user queries.
To address more complex queries, a viable approach involves decomposing the query and sourcing relevant online information for each segment, which we plan to explore in future work. 

\noindent\textbf{Data Source.}
Our dataset is built using the Baidu search engine, with a focus on sourcing documents in Chinese. 
Though there may be some regional biases due to limitations in Chinese cities, our construction pipeline, backed by advanced search engines, can be adapted for other regions globally. 
This adaptability contributes to opportunities of creating a more comprehensive dataset.

\noindent\textbf{Evaluation.}
In our paper, we conceptualize spatiotemporal travel planning as a multi-objective optimization problem. 
However, the intricate nature of these objectives complicates the calculation of Pareto fronts, which serve as the golden reference plan for our task. 
Furthermore, the lack of reliable ground truths precludes the consideration of fine-tuning strategies related to RAG or post-retrieval compression. 
Future research may focus on developing methods to annotate trustworthy ground truths for this task, ensuring that evaluations are free from multifaceted biases.

\noindent\textbf{Baseline.}
Due to the challenges in collecting ground-truth information, we have opted not to consider training-based baseline methods in our current work. 
Although exploring the potential of training a travel agent is indeed valuable, developing a reliable specialist model for the complex spatiotemporal travel planning methods is something we leave for future research.
\section*{Ethical Statement}

Our dataset is constructed using publicly accessible Web content retrieved through the Baidu search engine, strictly adhering to the platform's terms of service and data usage policies. 
To ensure privacy compliance, all collected trajectories undergo rigorous desensitization: sensitive personal information is removed, and non-trajectory content (e.g., user profiles, comments) is filtered via automated processes without retaining original documents.  
The POI corpus is derived from the Baidu Map platform and the Web (\eg names, geocoordinates), with hallucinated or duplicate entries systematically eliminated. 
Upon publication, we will release the sanitized dataset and code to foster reproducibility, while withholding raw Web content to respect source providers' rights. 
This aligns with ethical research practices in handling Web-derived data while ensuring user anonymity.

\bibliography{custom}

\clearpage
\onecolumn
\appendix
\section{Dataset Details}

In this section, we present further details of our dataset.
It is noteworthy that our original dataset was developed in Chinese based on the Baidu search engine, and we provide an English version of all queries, POI data, and trajectories to facilitate global research.

\subsection{Query Data}
\label{appendix:query}

We select the most popular Chinese travel cities to construct our query dataset.
In specific, we refer to \href{https://www.sohu.com/a/624120806_484968}{Sohu Travel 2022's rankings}, and select the top 30 cities in mainland China. Below is the city list: 
\textsl{Chongqing, Wuhan, Sanya, Luoyang, Beijing, Nanjing,  
Shanghai, Xi'an, Qingdao, Guiyang, Fuzhou, Xiamen,  
Hangzhou, Shaoxing, Guilin, Jinan, Zhaoqing, Foshan,  
Chengdu, Changchun, Suzhou, Rizhao, Yantai, Huangshan,  
Yangzhou, Zhangjiajie, Guangzhou, Nanning, Jilin, Binzhou}.

\begin{table*}[!ht]
    \centering
    \caption{Constraint taxonomy.}
    \resizebox{0.9\linewidth}{!}{
    \begin{tabular}{l|l}
    \toprule
   \textbf{ Constraint Category} & \textbf{Constraints} \\
    \hline
    Season & Spring,\ Summer,\ Autumn,\ Winter \\
    \hdashline[0.5pt/1pt]
    \multirow{2}{*}{Holiday} & Spring Festival,\ Qingming Festival,\ Labor Day,\\ 
    & Gragon Boat Festival,\ Mid-Autumn Festival,\ National Day \\
    \hdashline[0.5pt/1pt]
    \multirow{2}{*}{POI Categoty} & Natural Landscapes,\ Historical \& Cultural Heritage,\ Leisure \& Recreation Areas,\\
    & Art \& Technology Hubs,\ City Sightseeing,\ Religious \& Spiritual Sites \\
    \hdashline[0.5pt/1pt]
    Traveler Category & Senior,\ Single,\ Couple,\ Parent-child \\
    \hdashline[0.5pt/1pt]
    Trip Compactness & Special Forces-style \\
    \bottomrule
    \end{tabular}
    }
    \label{table:constraint}
\end{table*}

In \tabref{table:constraint}, we detail the personalized constraints according to their categories.
And we report the query data distribution in terms of the constraint type In \tabref{table:query}.
To diversify the travel duration, we generate queries specified with 3, 4, and 5 days.

\begin{table*}[!ht]
    \centering
    \caption{Query distribution according to constraints.}
    \resizebox{0.45\linewidth}{!}{
    \begin{tabular}{l|l|c}
    \toprule
    \textbf{Query Category} & \textbf{Constraint Category} & \textbf{\#Query} \\
    \hline
    Generic & - & 115 \\
    \hdashline[0.5pt/1pt]
    \multirow{5}{*}{Personal} & Season & 425 \\
    & Holiday & 608 \\
    & POI Category & 634 \\
    & Traveler Category & 451 \\
    & Trip Compactness & 115 \\
    & Total & 2233 \\
    \hdashline[0.5pt/1pt]
    Total & - & 2348 \\
    \bottomrule
    \end{tabular}
    }
    \label{table:query}
\end{table*}

\subsection{POI and Trajectory Data}

For each query, we retrieve 10, 5, 20 documents for POI collection, real-time POI refinement, and trajectory collection, respectively.
We associate 8 valid trajectories for each query and abandon instances that are insufficient in number.
To emulate the retrieval-augmented planning scenario for each query independently, it is essential that the POIs featured in the retrieved trajectory should be included in the candidate set. 
Thus, we locate the POIs that are newly present in the trajectories and add them to the candidate POI set.

\section{Metric Details}
\label{appendix:metrics}

In this section, we detail each evaluation metric as follows:

\begin{table}[!ht]
    \centering
    \vspace{-8pt}
    \caption{Metric taxonomy.}
    \vspace{-8pt}
    \resizebox{0.5\linewidth}{!}{
    \begin{tabular}{l|l|l}
    \toprule
    \textbf{Dimension} & \textbf{Metric} & \textbf{Description}\\
    \hline
    \multirow{2}{*}{Commonsense} & FR & Legitimacy of POIs \\
    & RR & Non-redundancy of POIs. \\
    \hdashline[0.5pt/1pt]
    Spatial & DMR & Route efficiency. \\
    \hdashline[0.5pt/1pt]
    \multirow{3}{*}{Temporal} & STR & Rationality of arrival time. \\
    & DUR & Rationality of visit duration. \\
    & TBR & Comfort level of schedule. \\
    \hdashline[0.5pt/1pt]
    POI Semantic & PP & POI popularity.\\
    \hdashline[0.5pt/1pt]
    \multirow{2}{*}{Query Relevance} & PR & POI relevance. \\
    & TSR & Time schedule relevance. \\
    \bottomrule
    \end{tabular}
    }
    \vspace{-8pt}
    \label{table:metric}
\end{table}

\begin{itemize}[leftmargin=*,itemsep=2pt,topsep=0pt,parsep=0pt]
\item {\textbf{Failure Rate (FR)}: 
This metric quantifies the number of attractions absent from the candidate set, which may indicate hallucinations by the LLM agents.
}
\item {\textbf{Repetition Rate (RR)}: 
We measure the frequency of POI repetition in plans to assess basic commonsense awareness of the agents.
}
\item {\textbf{Distance Margin Ratio (DMR)}: 
This metric evaluates the margin ratio between the total distance required to transfer between attractions in the generated plan and the optimal distance determined by the Traveling Salesman Problem (TSP) solver\footnote{\url{https://github.com/fillipe-gsm/python-tsp}}.
} 
\item {\textbf{Start Time Rationality (STR)}:
It is essential to determine whether the scheduled times for visiting attractions are appropriate. 
Due to the lack of uniform standard, we prompt LLMs to consult POI-level temporal information and verify the plausibility of scheduled arrival times by binary judgment (\ie yes or no), then calculating the total acceptable rate.
}
\item {\textbf{Duration Underflow Ratio (DUR)}:
This metric assesses how well the planned visit durations align with the expected time spans for each attraction. 
We directly used the visiting duration tags of POIs in our dataset, and then compute the average duration underflow ratio.
}
\item {\textbf{Time Buffer Ratio (TBR)}:
Given that overly tight schedules are generally inadvisable, this metric evaluates the flexibility of plans by estimating the proportion of buffer time available between attractions throughout the plan.
}
\item {\textbf{POI Popularity (PP)}:
Popular attractions are typically favored, thus, we curate LLMs to offer a golden popularity ranking based on the retrieved attraction leaderboard data. 
We then calculate the top $M$ recall, where $M$ represents the number of selected POIs in the plan.
}
\item {\textbf{POI Relevance (PR)}:
To check the alignment of the delivered plans with the soft constraints in personal queries, we nudge LLMs to judge the matchness of each POI in a binary manner.
}
\item {\textbf{Time Schedule Relevance (TSR)}:
Similarly, LLMs are elicited to gauge whether the time intervals arranged for attractions are consistent with the personalized requirements.
}
\end{itemize}

A summary of the proposed evaluation metrics is presented in \tabref{table:metric}. 
The prompts for LLM-based evaluation are showed in \appref{appendix:prompt_evaluation}.
Moreover, we conduct human evaluation to validate the proficiency of LLM-based evaluation.
We randomly sample 100 plans generated by GPT-4o, and test the performance of Qwen2.5-72B-Instruct in evaluating results across three LLM-based metrics: STR, PR, and TSR.
For each metric, we report three measurements: (1) agreement rate between the judgments of LLMs and humans; (2) Kendall Tau and (3) Spearman coefficient of the method rankings across all sampled queries.
\tabref{table:human} demonstrates the alignment between LLM and human evaluators on our metrics.

\begin{table}[!ht]
    \centering
    \caption{The alignment performance between LLM and human evalutors across three metrics (\%).}
    \resizebox{0.58\linewidth}{!}{
    \begin{tabular}{c|c|c|c}
    \toprule
    \textbf{Metric} & \textbf{Agreement Rate} & \textbf{Kendall Tau} & \textbf{Spearman Coefficient}\\
    \hline
    STR & 93.70 & 60.24 & 66.61\\
    \hdashline[0.5pt/1pt]
    PR & 95.54 & 68.07 & 71.72\\
    \hdashline[0.5pt/1pt]
    TSR & 97.55 & 74.34 & 77.66\\
    \bottomrule
    \end{tabular}
    }
    \vspace{-8pt}
    \label{table:human}
\end{table}

\section{Experiments}
\label{appendix:experiments}

\subsection{Experimental Setup}
\label{appendix:setup}

In this section, we provide some details of our experiments.
For stability, we set the temperature as 0 for all base models.
For the cases that agents fail to generate a grammatically correct answer, we retry several times until a success, because the delivery failure is not considered in our evaluation system.
Since the strong reasoning ability of DeepSeek-R1, we omit the implementation of complex ground-up planning strategies (\ie CoT, Reflextion, MAC, MAD) on the model.
The prompts designed for these baseline methods are presented in \appref{appendix:prompt_baseline}.
To assess our baselines, we adopt Qwen2.5-72B-Instruct as the LLM evaluator.

\subsection{Results on LLaMA3.3-70B-Instruct}
\label{appendix:llama3.3}

We present the experimental results implemented by LLaMA3.3-70B-Instruct in \tabref{table:llama}, which are generally consistent with the outcomes in \tabref{table:main}.

\begin{table*}[ht]
    \centering
    \caption{Main results (\%) of LLaMA3.3-70B-Instruct on our dataset.}
    \resizebox{\linewidth}{!}{
    \begin{tabular}{l|cc|ccccccc|cccc|c}
    \toprule
    \textbf{Method} & \textbf{FR$\downarrow$} & \textbf{RR$\downarrow$} & \textbf{DMR$\downarrow$} & \textbf{DUR$\downarrow$} & \textbf{TBR$\uparrow$} & \textbf{STR$\uparrow$} & \textbf{PP$\uparrow$} & \textbf{PR$\uparrow$} & \textbf{TSR$\uparrow$} & $R_S\downarrow$ & $R_T\downarrow$ & $R_P\downarrow$ & $R_R\downarrow$ & $R_C\downarrow$ \\ 
    \hline
    \multicolumn{15}{c}{LLaMA3.3-70B-Instruct} \\ 
    \hline
    Direct & \textbf{0.67} & \textbf{0.01} & 90.55 & 7.41 & 24.14 & \underline{77.07} & 46.27 & 79.75 & 89.32 & 8.00 & \underline{4.67} & 7.00 & 7.50 & 6.79 \\ 
    CoT & \underline{0.97} & \textbf{0.01} & 86.45 & 7.77 & 24.76 & \textbf{78.10} & 45.52 & \textbf{80.31} & 87.97 & 7.00 & \textbf{4.33} & 10.00 & 6.00 & 6.83 \\ 
    Reflextion & 1.37 & 0.22 & 94.71 & 8.19 & 26.40 & 76.73 & 45.73 & \underline{80.28} & \underline{89.96} & 10.00 & 7.00 & 9.00 & \textbf{2.00} & 7.00 \\ 
    MAC & 3.55 & 3.73 & 97.68 & 9.61 & \textbf{31.46} & 75.47 & 42.11 & 79.98 & 88.38 & 11.00 & 7.67 & 11.00 & 7.00 & 9.17 \\ 
    MAD & 1.37 & 0.16 & 91.77 & 7.81 & \underline{26.42} & 76.26 & 46.27 & 79.05 & \textbf{90.14} & 9.00 & 7.00 & 7.00 & 6.00 & 7.25 \\ 
    \hdashline[0.5pt/1pt]
    RAG($M$=8) & 3.95 & 0.08 & 84.92 & \textbf{5.99} & 20.80 & 76.94 & 56.00 & 79.91 & 89.09 & 5.00 & 5.33 & \textbf{1.00} & 7.00 & 4.58 \\ 
    RAG($M$=4) & 2.81 & 0.07 & 83.67 & 6.10 & 21.02 & 76.77 & 53.60 & 79.99 & 89.68 & 4.00 & 6.33 & 4.00 & \underline{3.00} & \underline{4.33} \\ 
    RAG($M$=1) & 2.53 & \underline{0.06} & 85.36 & 6.56 & 22.68 & 77.00 & 50.43 & 79.79 & 89.55 & 6.00 & \underline{4.67} & 6.00 & 5.50 & 5.54 \\ 
    RAG+Extr.($M$=4) & 3.33 & 0.09 & 83.60 & \underline{6.05} & 21.06 & 76.70 & \textbf{54.66} & 79.73 & 89.30 & 3.00 & 6.67 & \underline{2.00} & 8.50 & 5.04 \\ 
    RAG+Extr.($M$=1) & 3.18 & 0.09 & \underline{82.42} & 6.62 & 22.57 & 76.77 & 52.23 & 79.83 & 89.38 & \underline{2.00} & 6.67 & 5.00 & 6.00 & 4.92 \\ 
    RAG+Abst. & 3.97 & 0.22 & \textbf{81.96} & 6.60 & 22.91 & 76.86 & \underline{54.44} & 79.36 & 89.42 & \textbf{1.00} & 5.33 & 3.00 & 7.50 & \textbf{4.21} \\ 
    \bottomrule
    \end{tabular}}
    \label{table:llama}
\end{table*}

\subsection{Results on Different Query Categories}
\label{appendix:sub-query}

\begin{figure*}[!ht]
    \centering
    \subfloat[Generic.]{\includegraphics[width=0.25\textwidth]{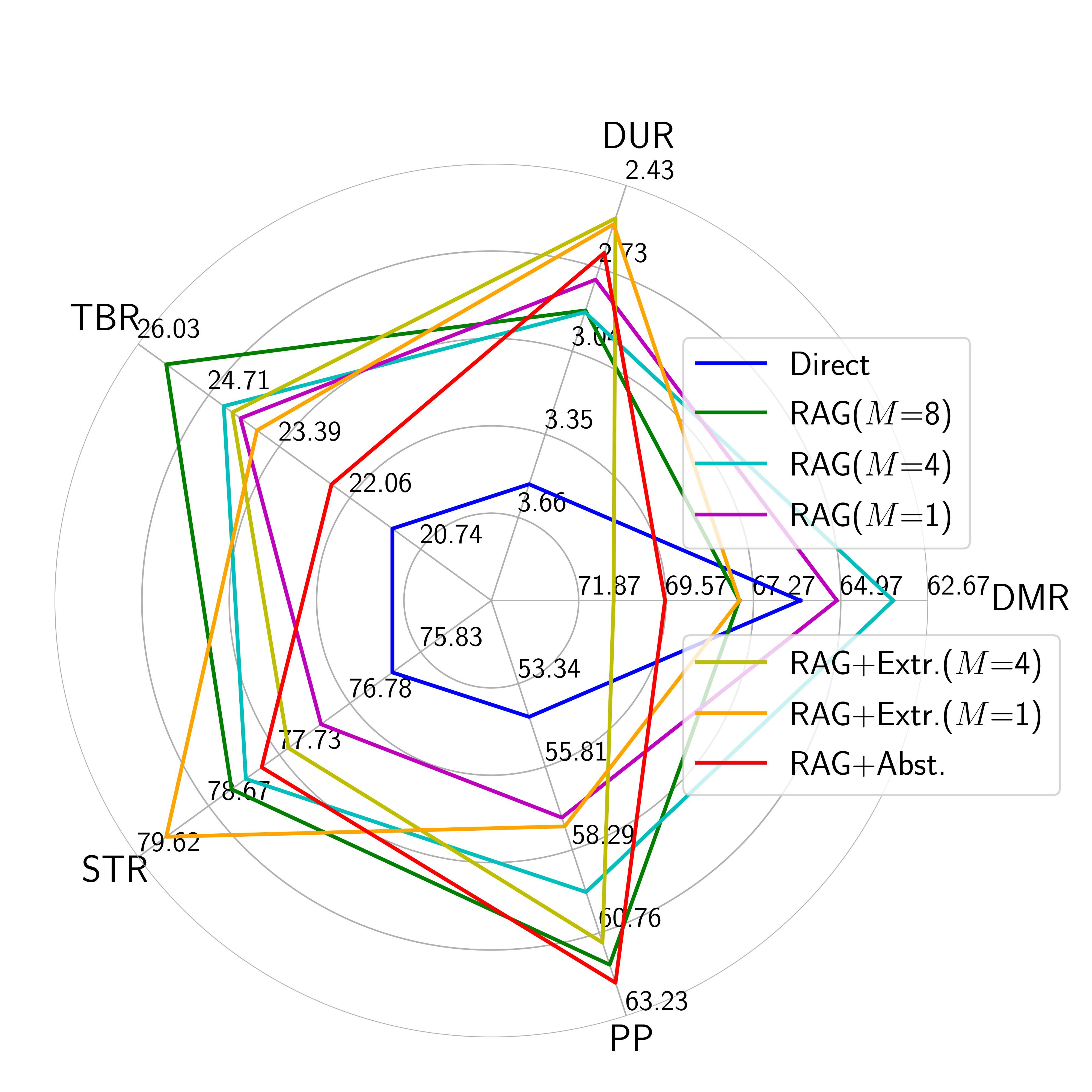}}
    \subfloat[Season.]{\includegraphics[width=0.25\textwidth]{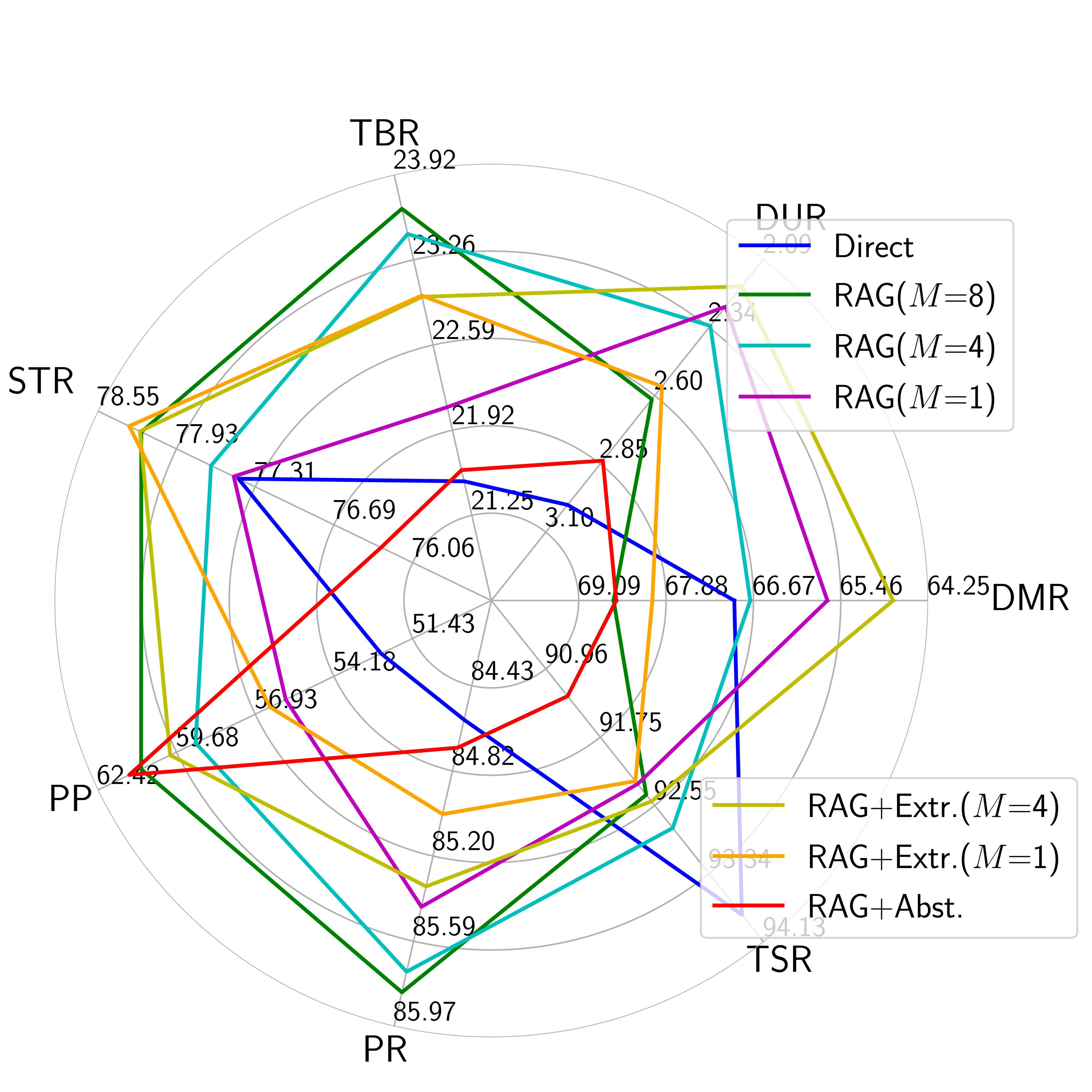}}
    \subfloat[Holiday.]{\includegraphics[width=0.25\textwidth]{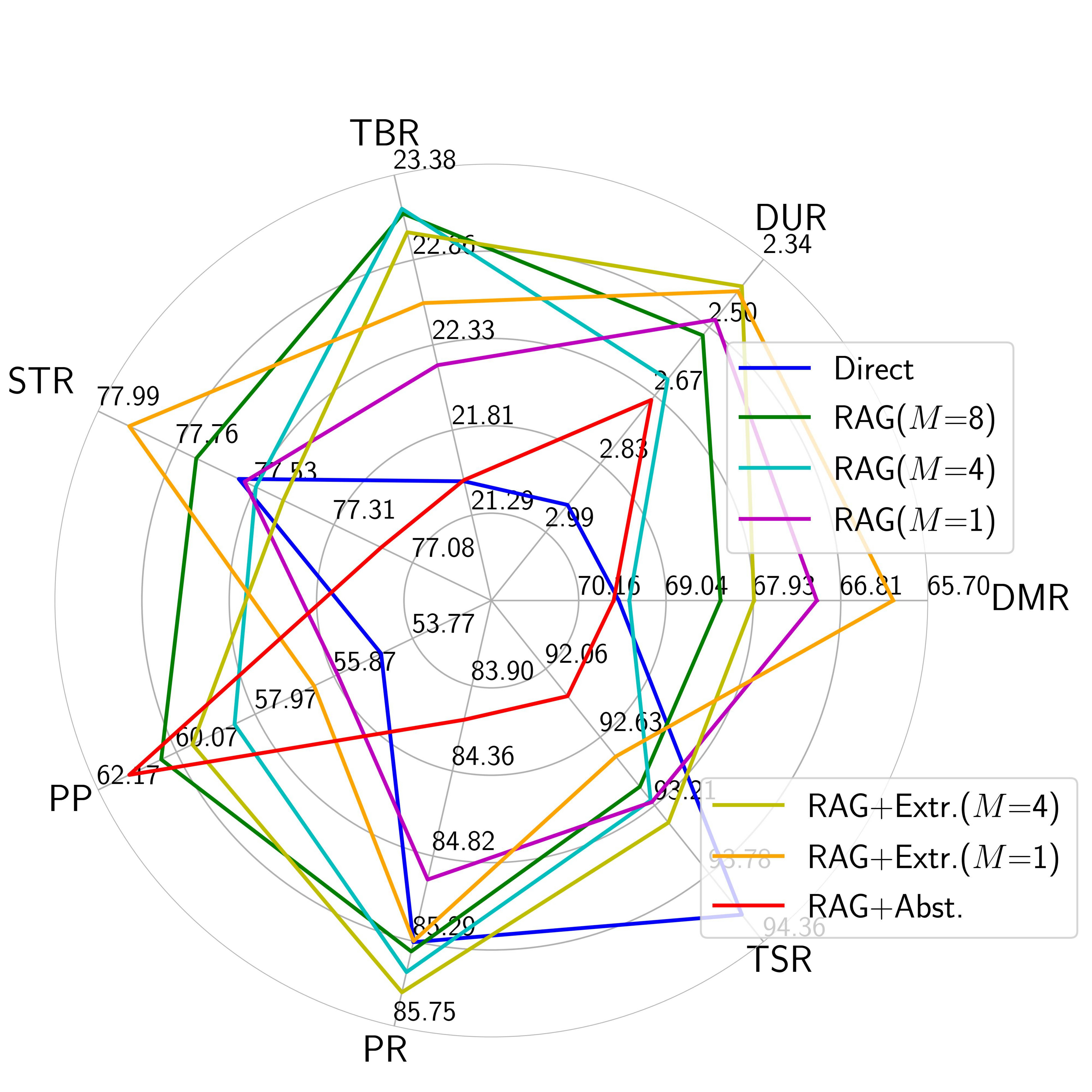}}
    \vspace{0.1pt}
    \subfloat[POI Category.]{\includegraphics[width=0.25\textwidth]{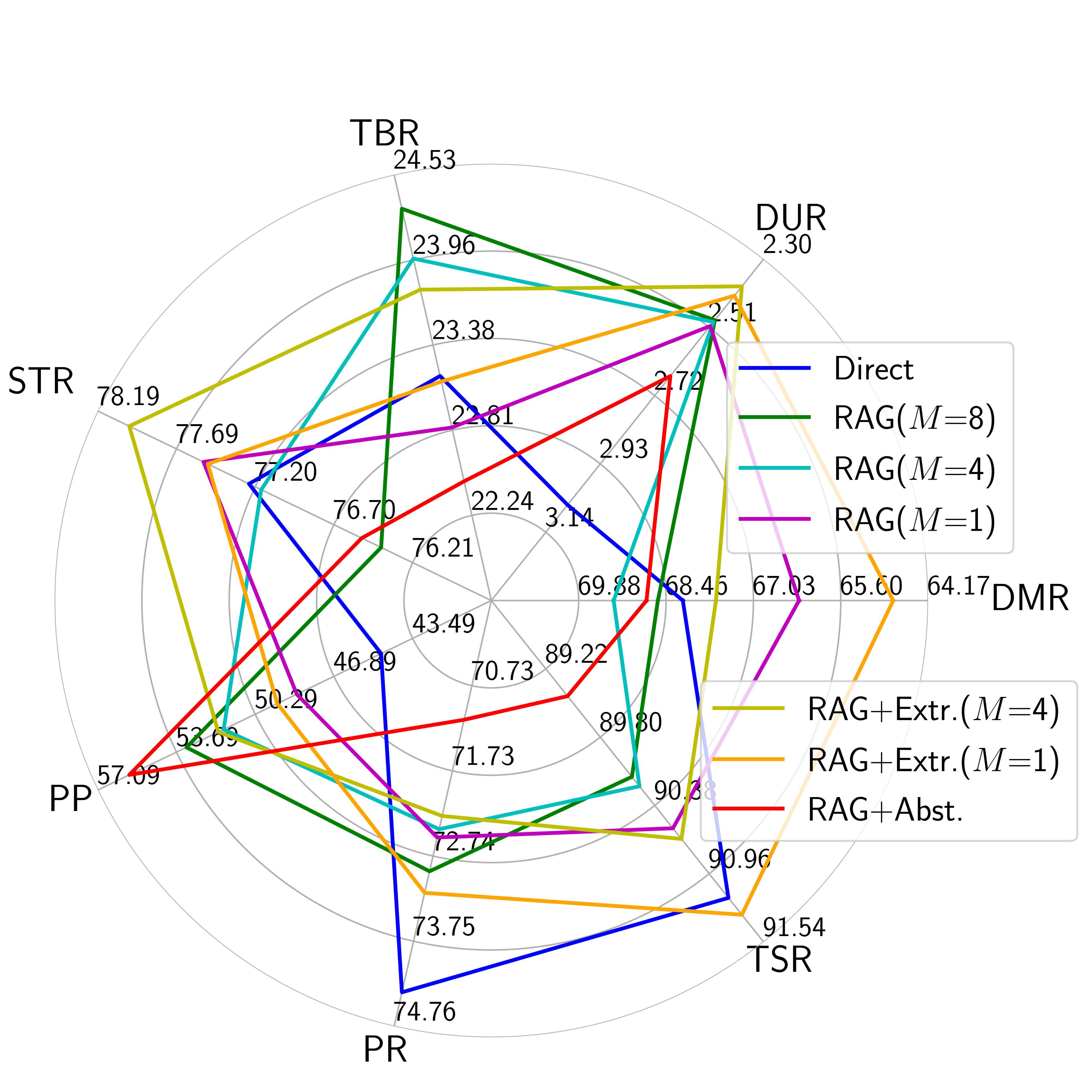}}
    \subfloat[Traveler Category.]{\includegraphics[width=0.25\textwidth]{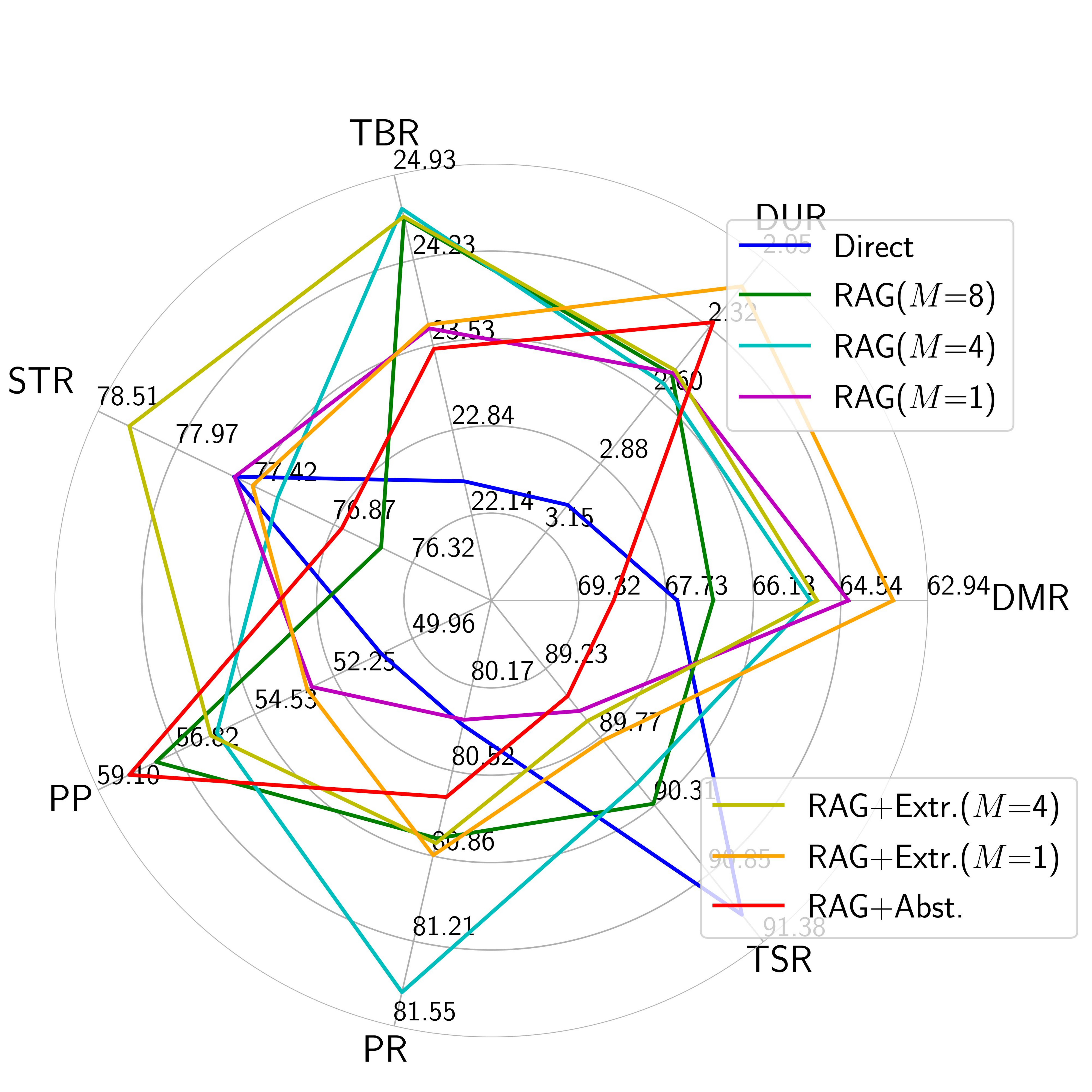}}
    \subfloat[Trip Compactness.]{\includegraphics[width=0.25\textwidth]{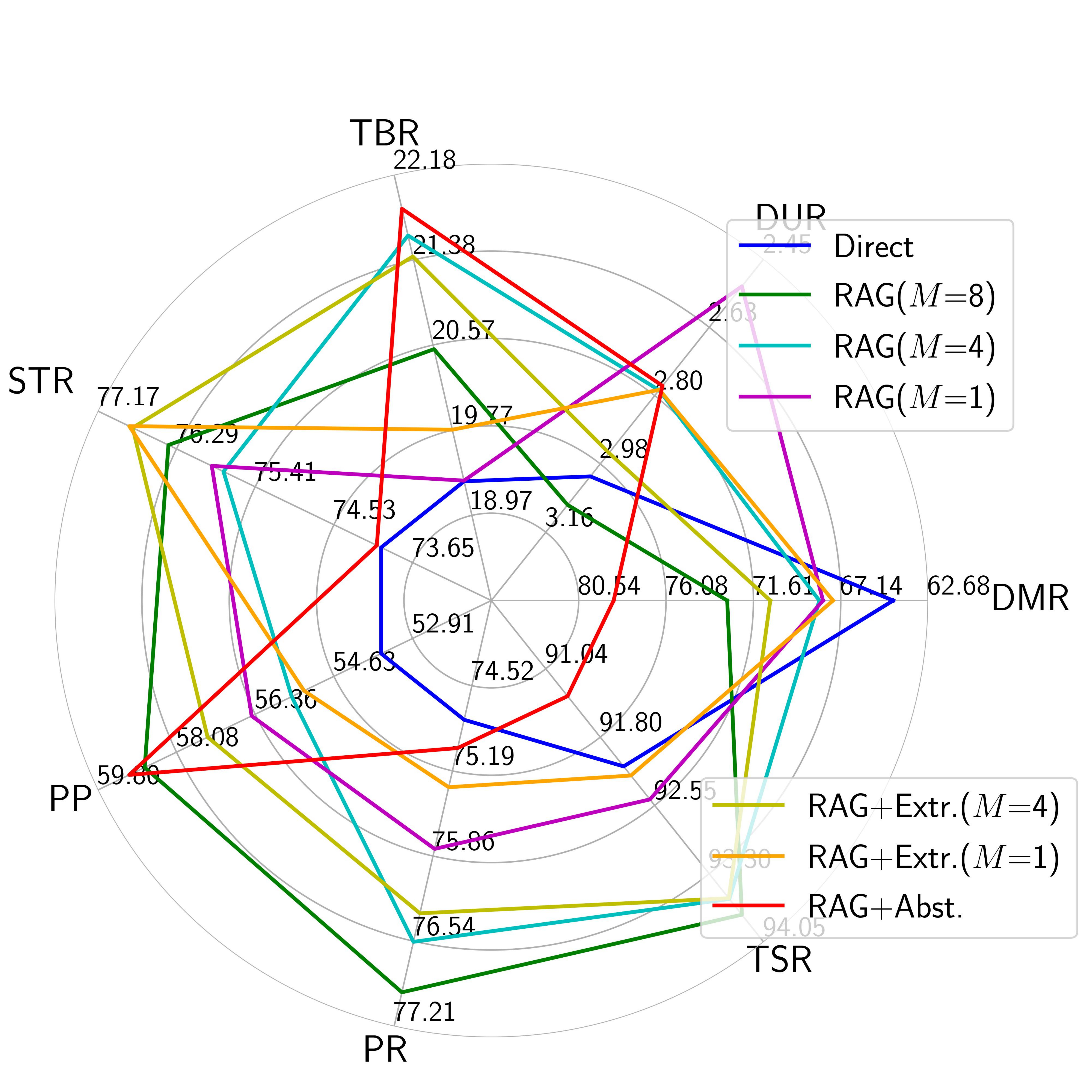}}
    \caption{Comparison between Direct and RAG strategies on query data with various categories, implemented by GPT-4o.}
    \vspace{-20pt}
    \label{figure:radar_gpt}
\end{figure*}

\begin{figure*}[!ht]
    \centering
    \subfloat[Generic.]{\includegraphics[width=0.25\textwidth]{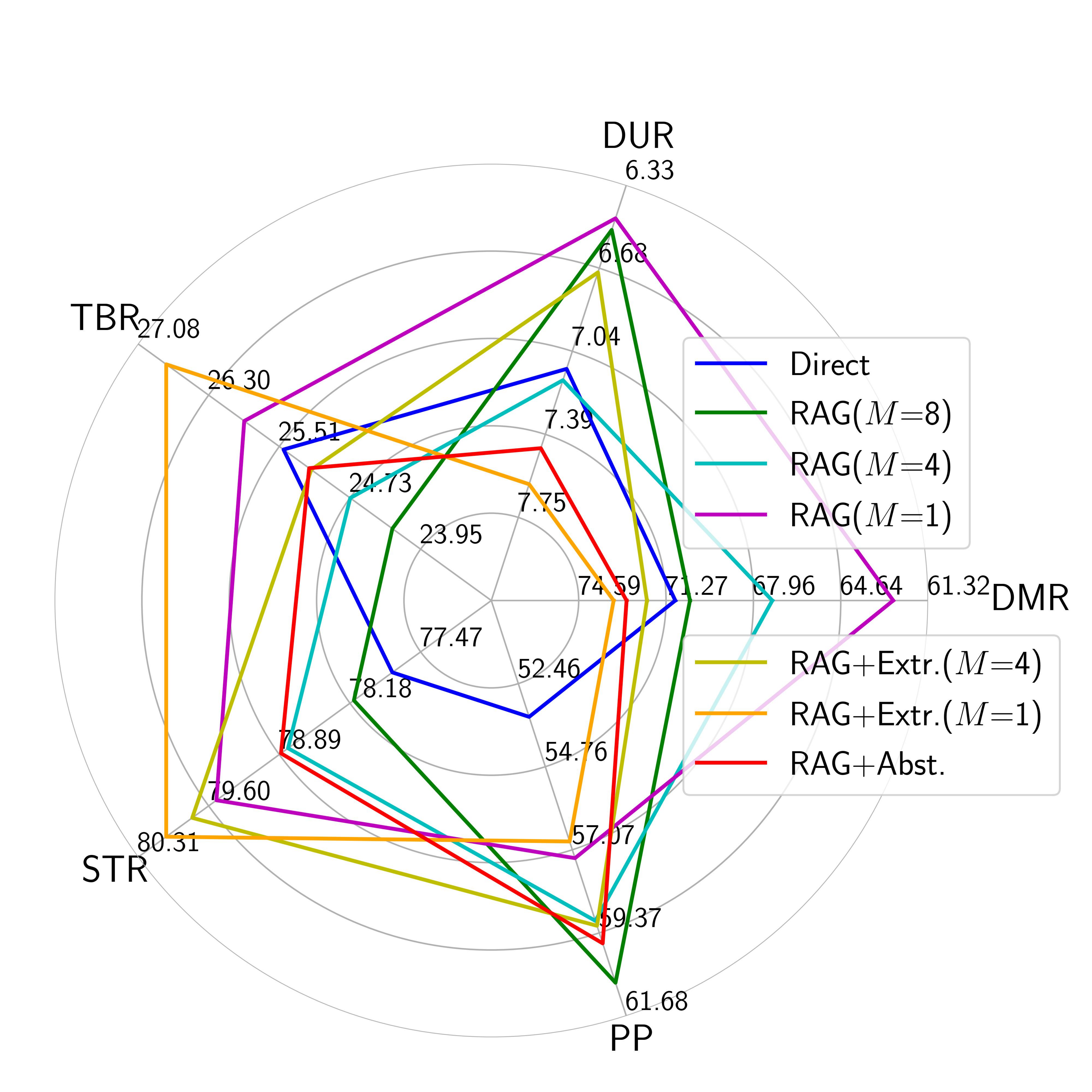}}
    \subfloat[Season.]{\includegraphics[width=0.25\textwidth]{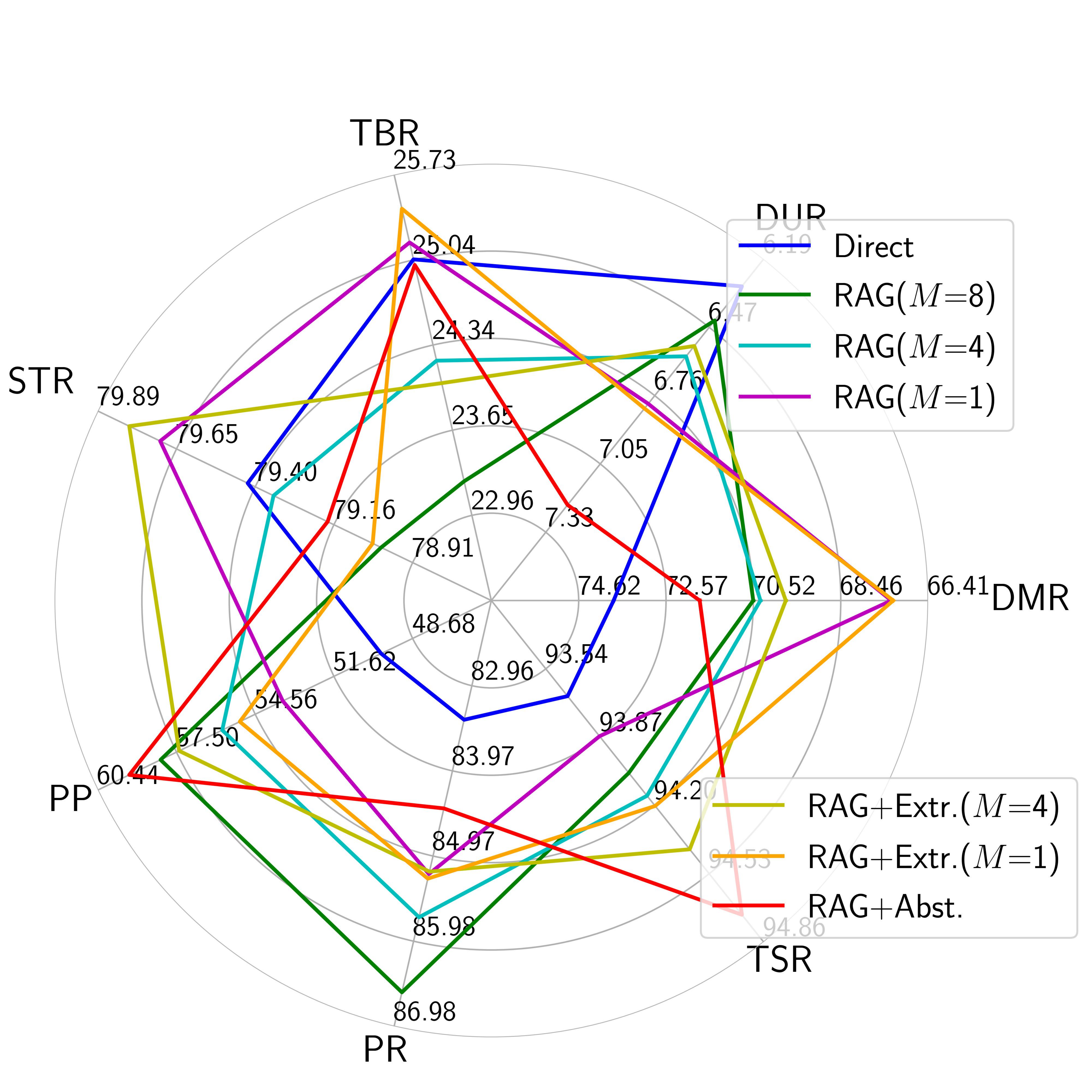}}
    \subfloat[Holiday.]{\includegraphics[width=0.25\textwidth]{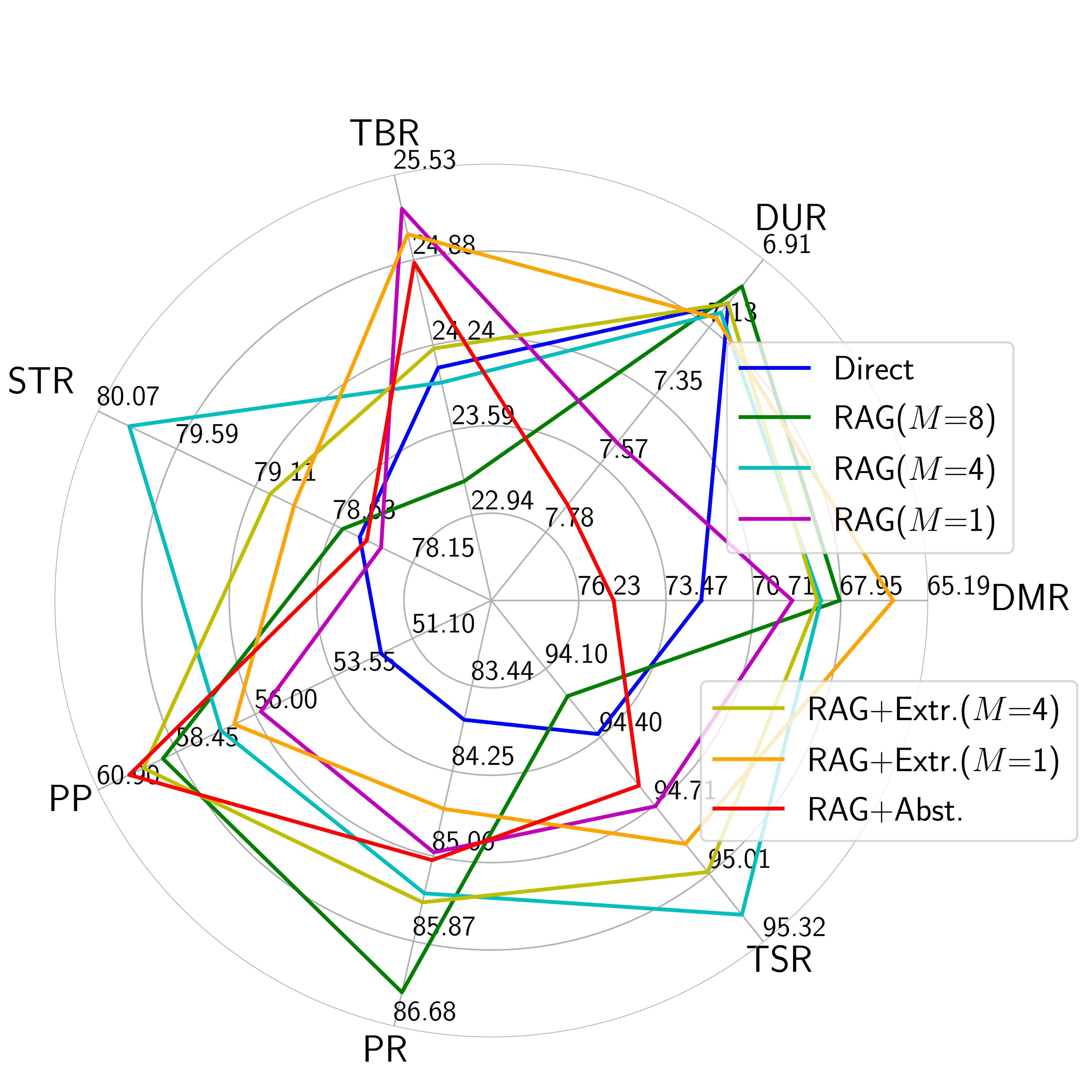}}
    \vspace{0.1pt}
    \subfloat[POI Category.]{\includegraphics[width=0.25\textwidth]{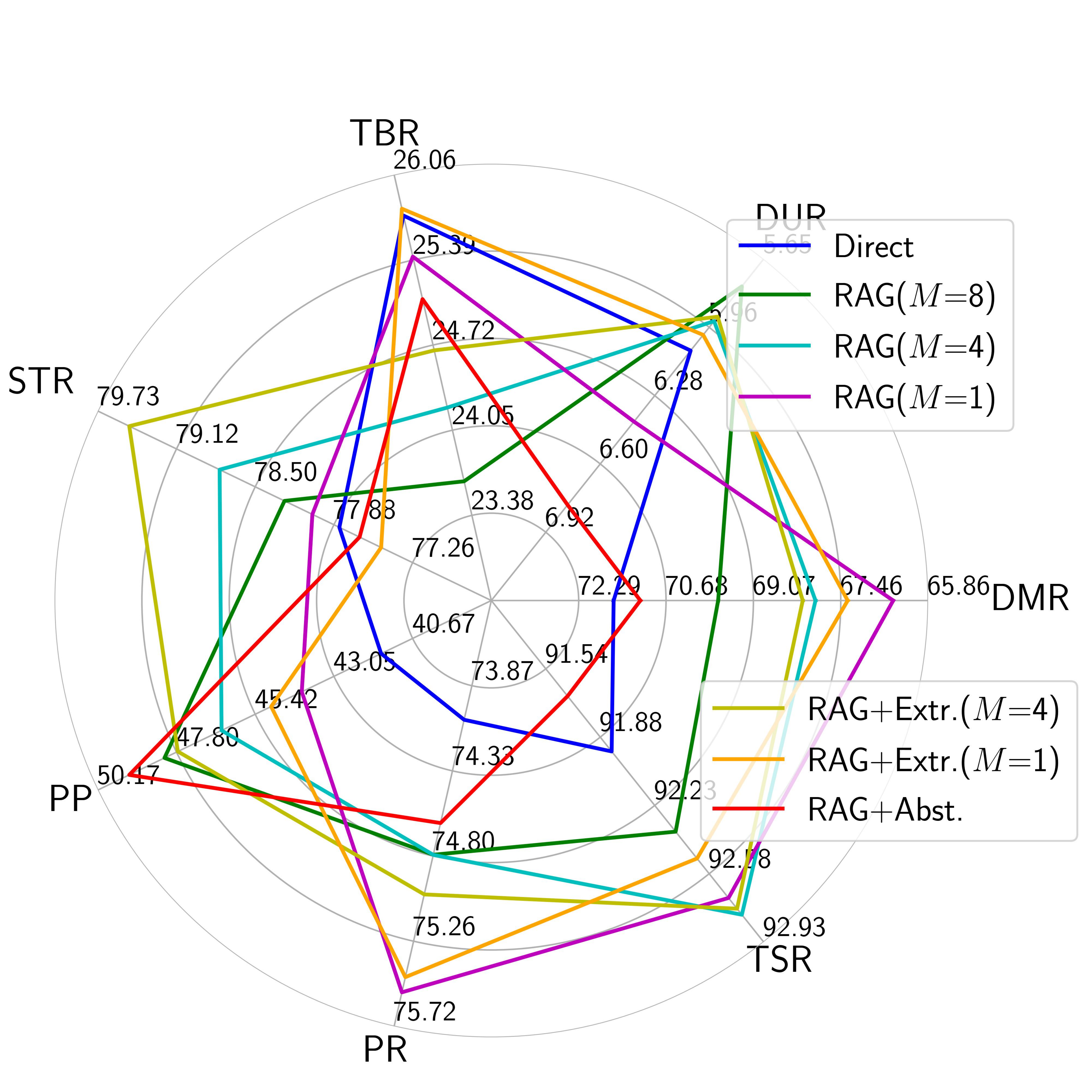}}
    \subfloat[Traveler Category.]{\includegraphics[width=0.25\textwidth]{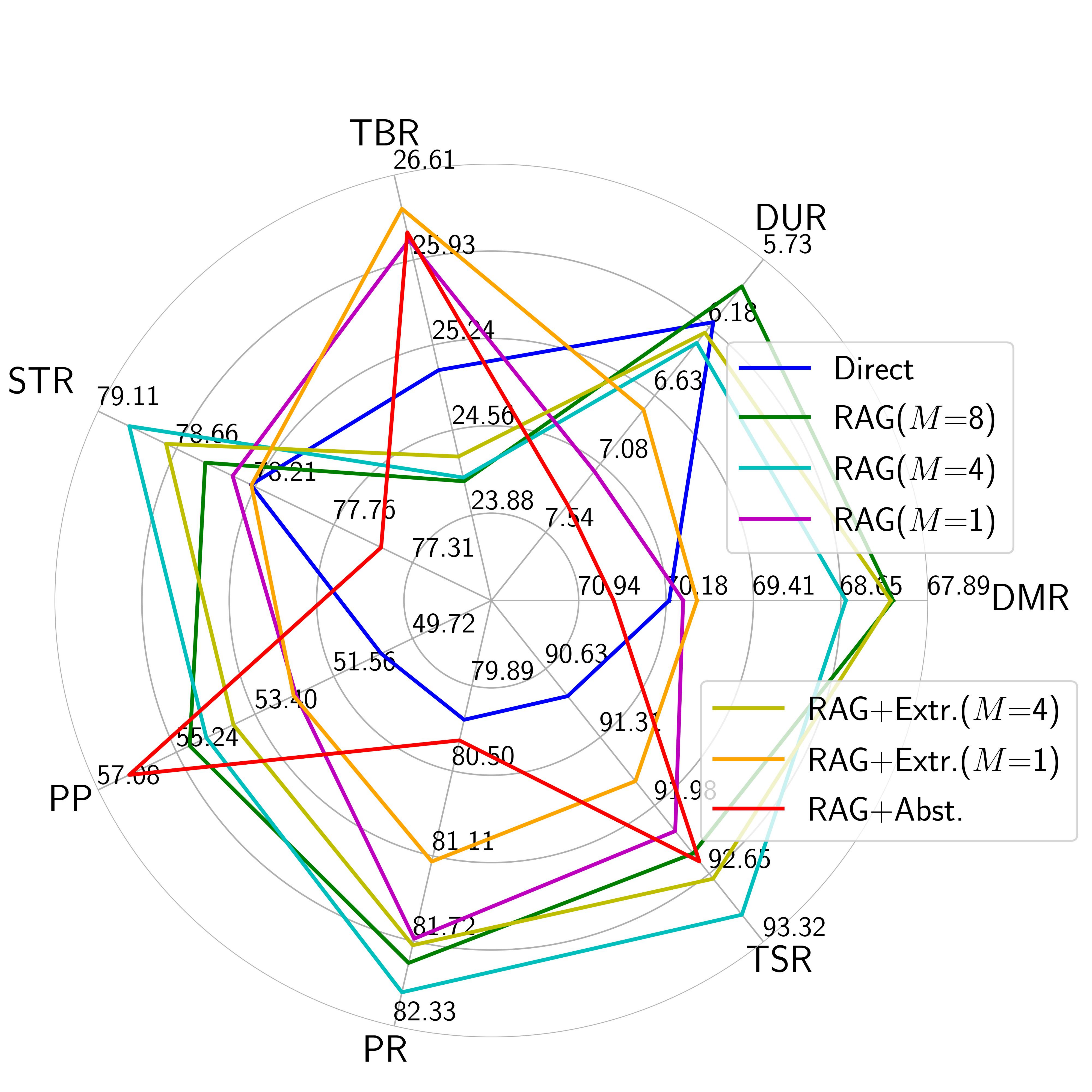}}
    \subfloat[Trip Compactness.]{\includegraphics[width=0.25\textwidth]{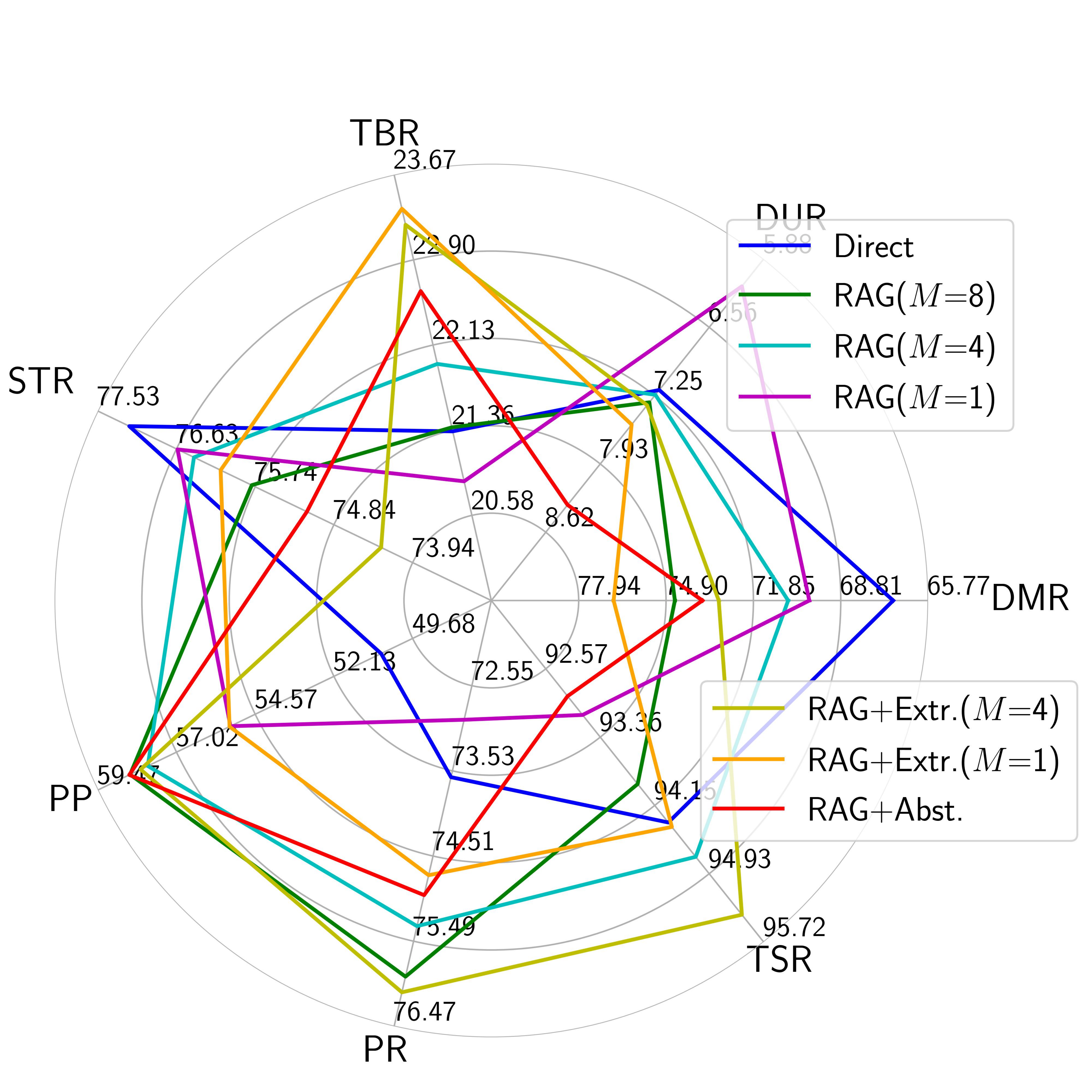}}
    \caption{Comparison between Direct and RAG strategies on query data with various categories, implemented by Qwen2.5-72B-Instruct.}
    \vspace{-20pt}
    \label{figure:radar_qwen}
\end{figure*}

\begin{figure*}[!ht]
    \centering
    \subfloat[Generic.]{\includegraphics[width=0.25\textwidth]{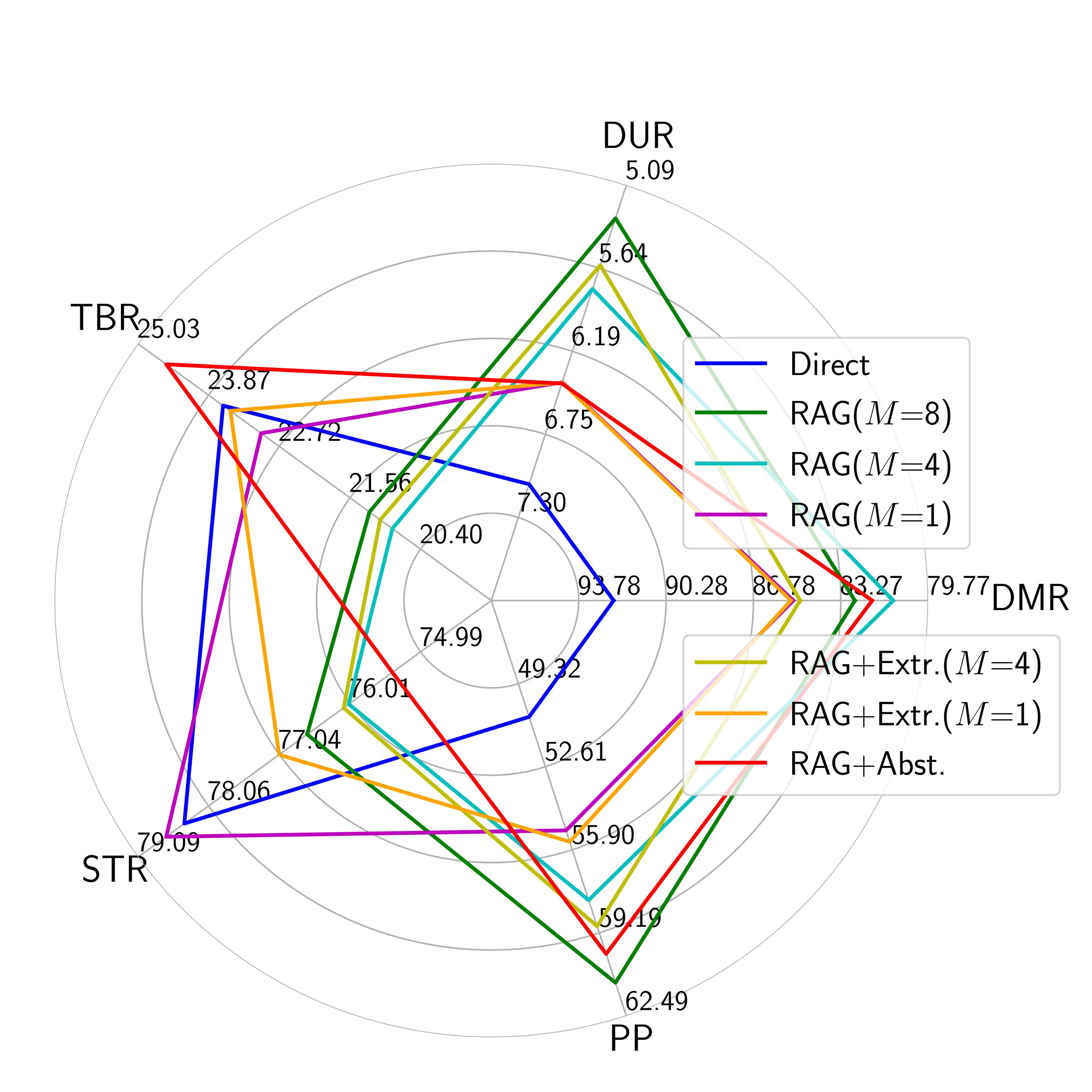}}
    \subfloat[Season.]{\includegraphics[width=0.25\textwidth]{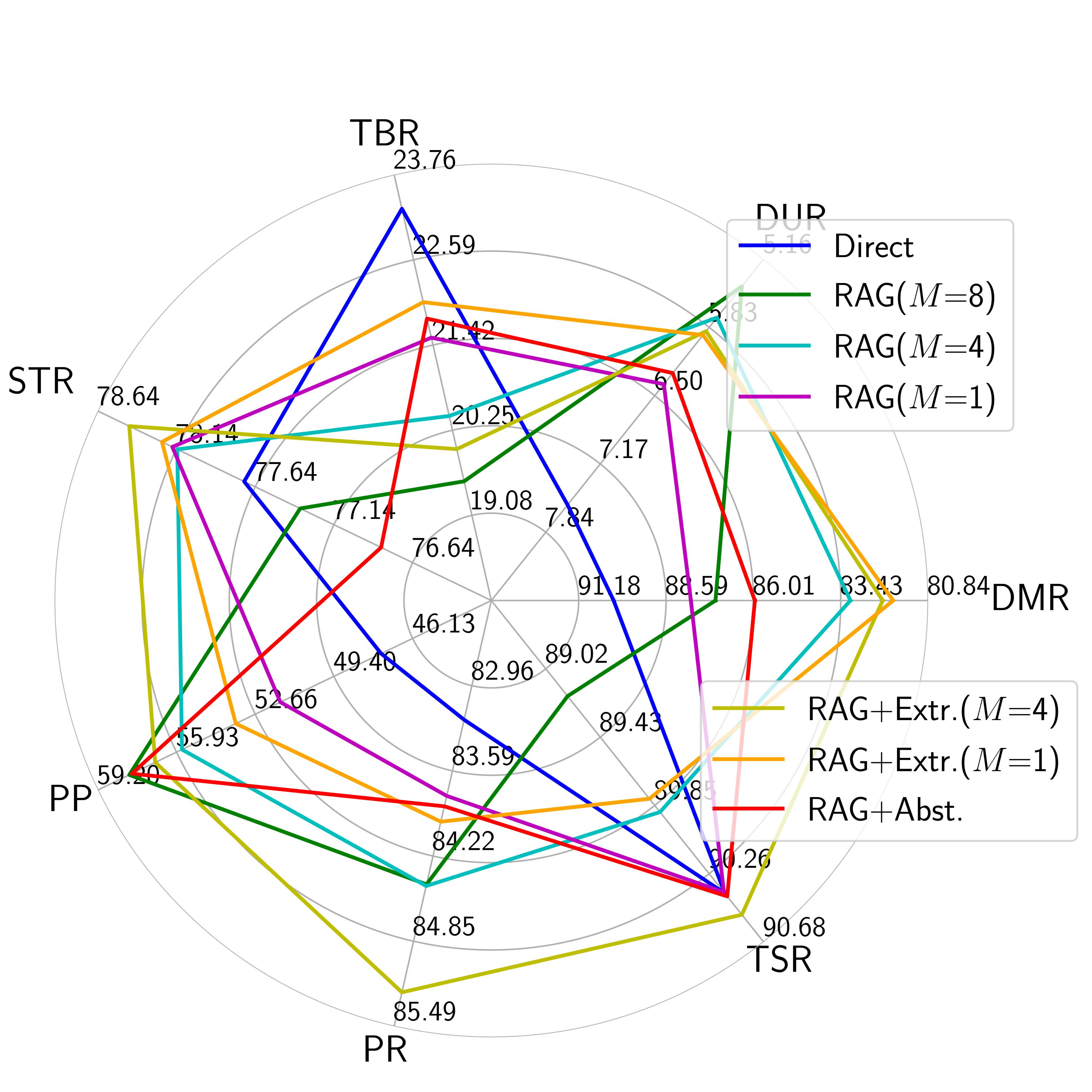}}
    \subfloat[Holiday.]{\includegraphics[width=0.25\textwidth]{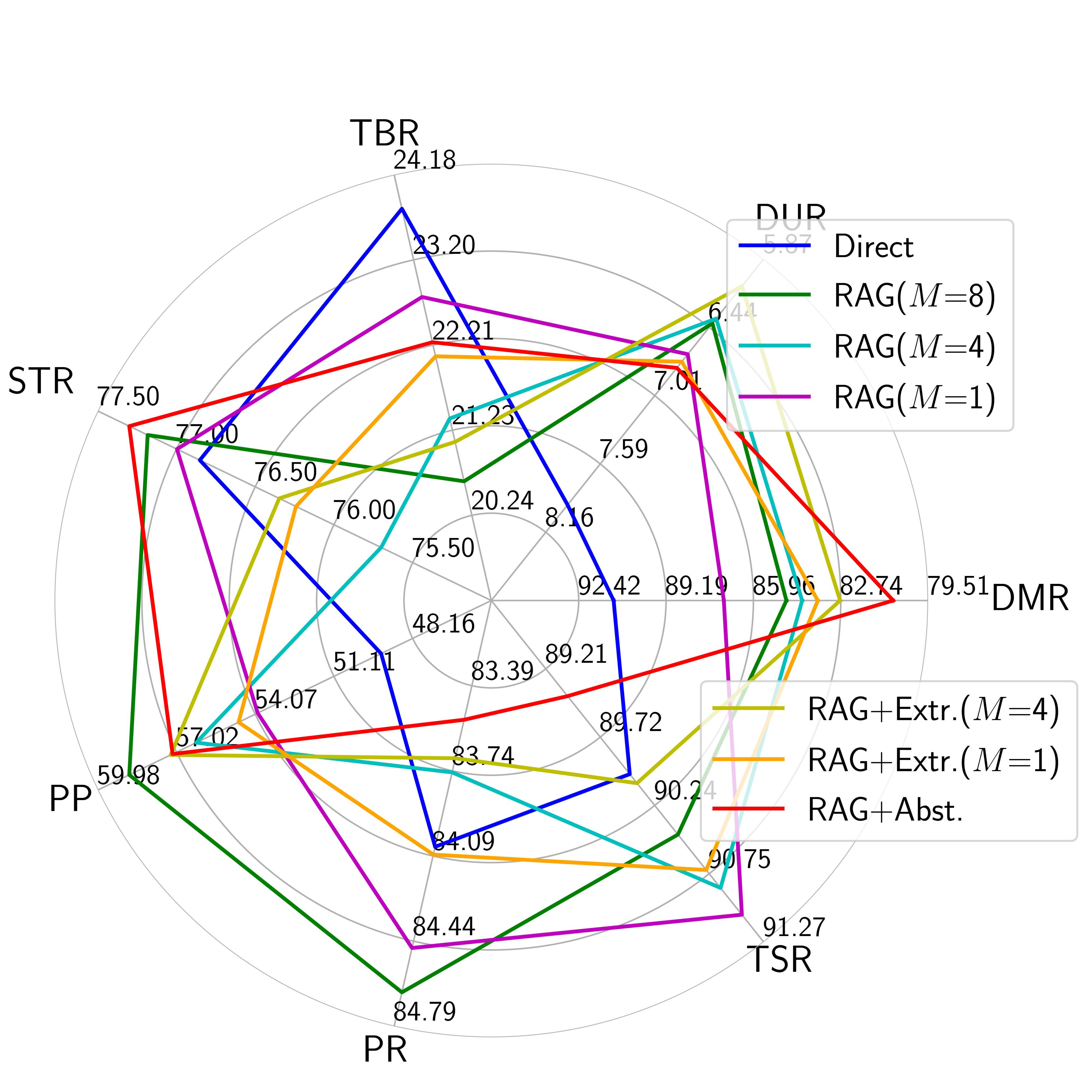}}
    \vspace{0.1pt}
    \subfloat[POI Category.]{\includegraphics[width=0.25\textwidth]{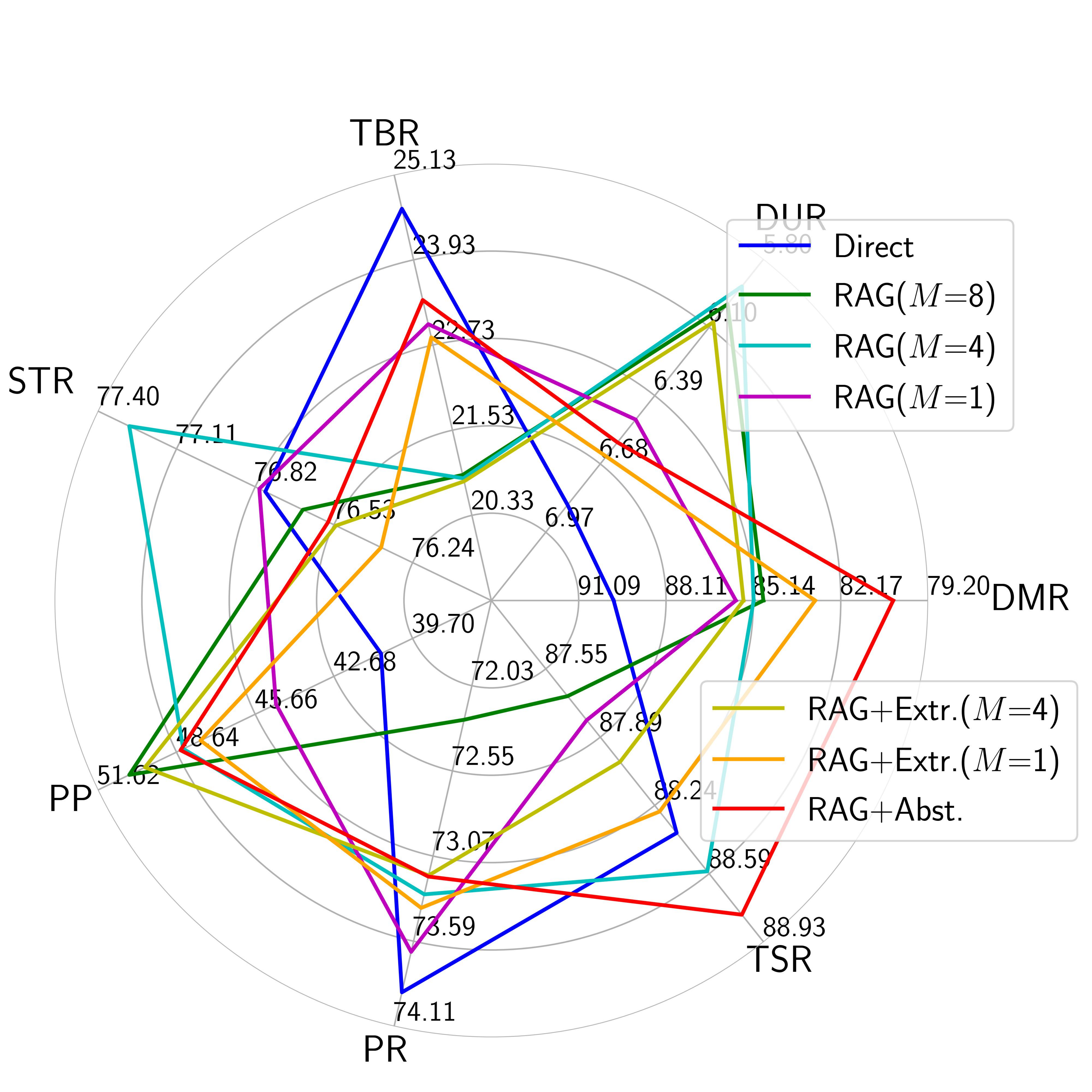}}
    \subfloat[Traveler Category.]{\includegraphics[width=0.25\textwidth]{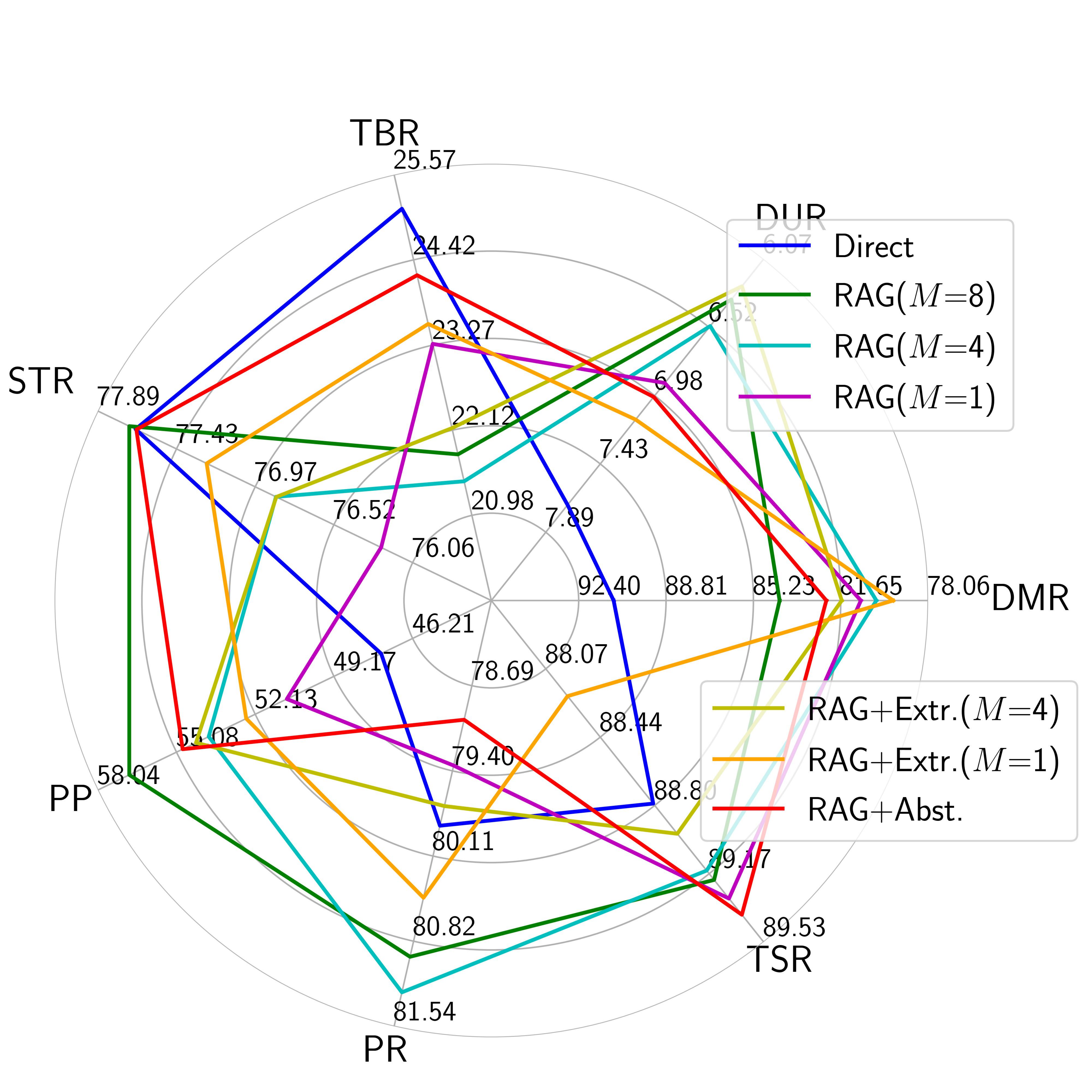}}
    \subfloat[Trip Compactness.]{\includegraphics[width=0.25\textwidth]{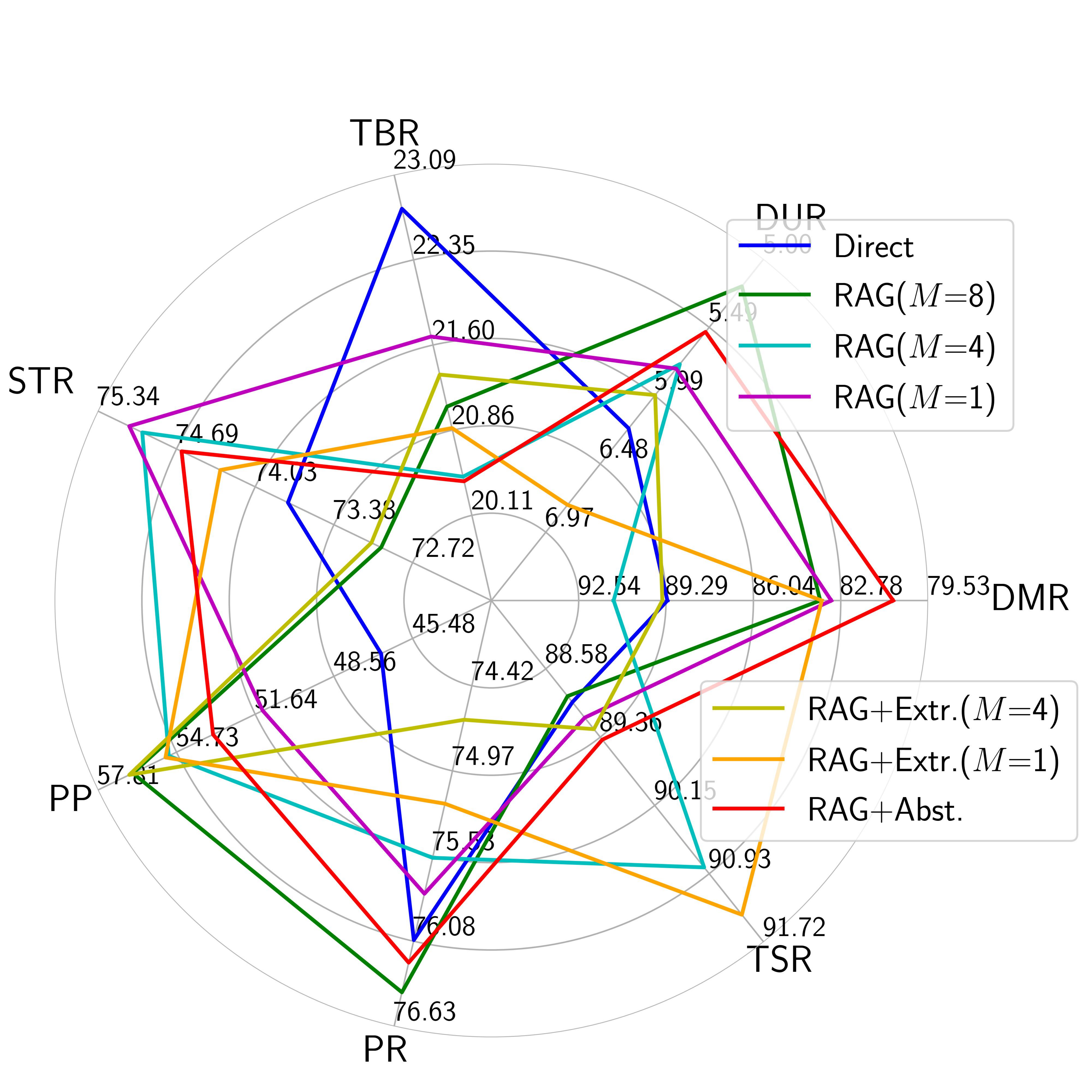}}
    \caption{Comparison between Direct and RAG strategies on query data with various categories, implemented by LLaMA3.3-70B-Instruct.}
    \vspace{-20pt}
    \label{figure:radar_llama}
\end{figure*}

\begin{figure*}[!ht]
    \centering
    \subfloat[Generic.]{\includegraphics[width=0.25\textwidth]{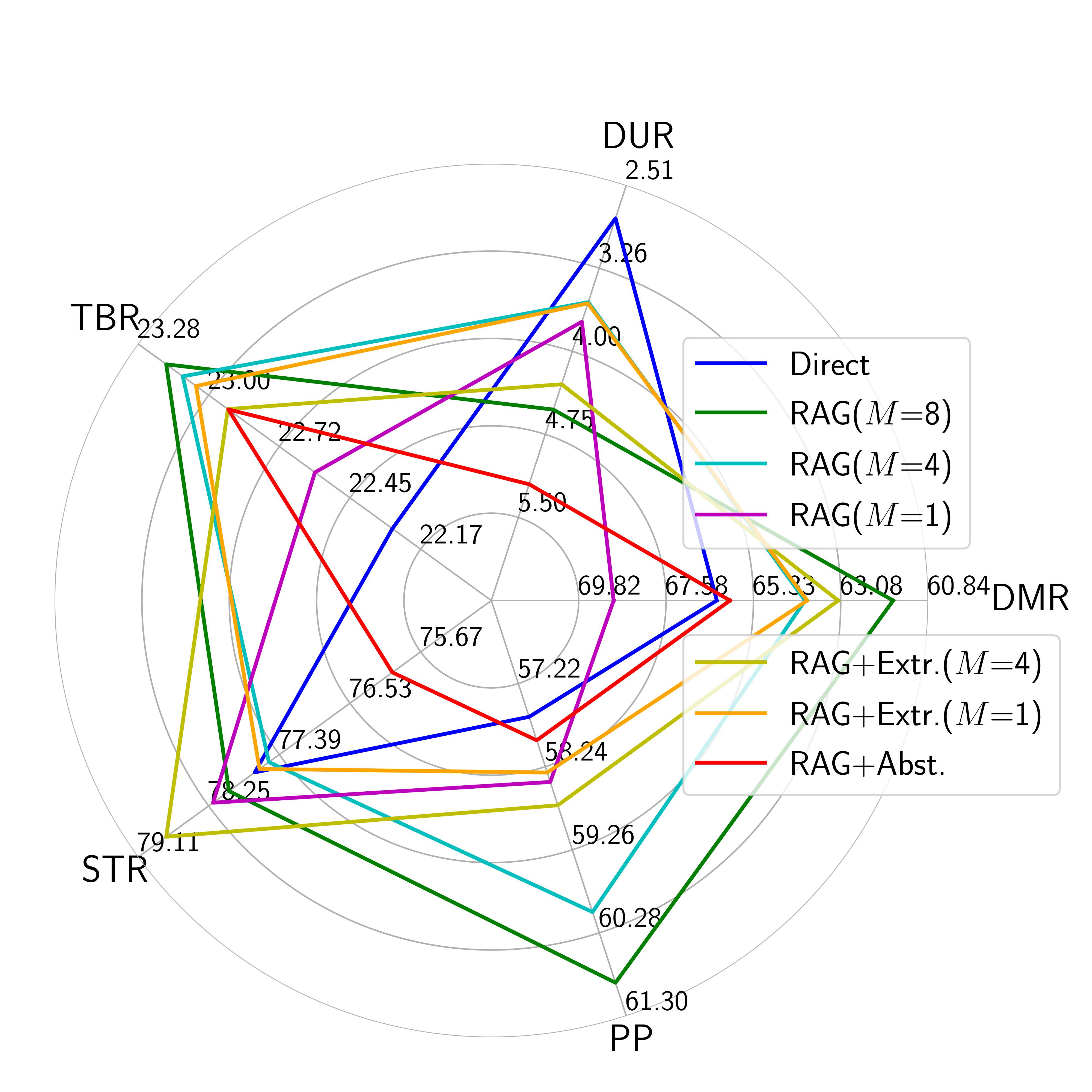}}
    \subfloat[Season.]{\includegraphics[width=0.25\textwidth]{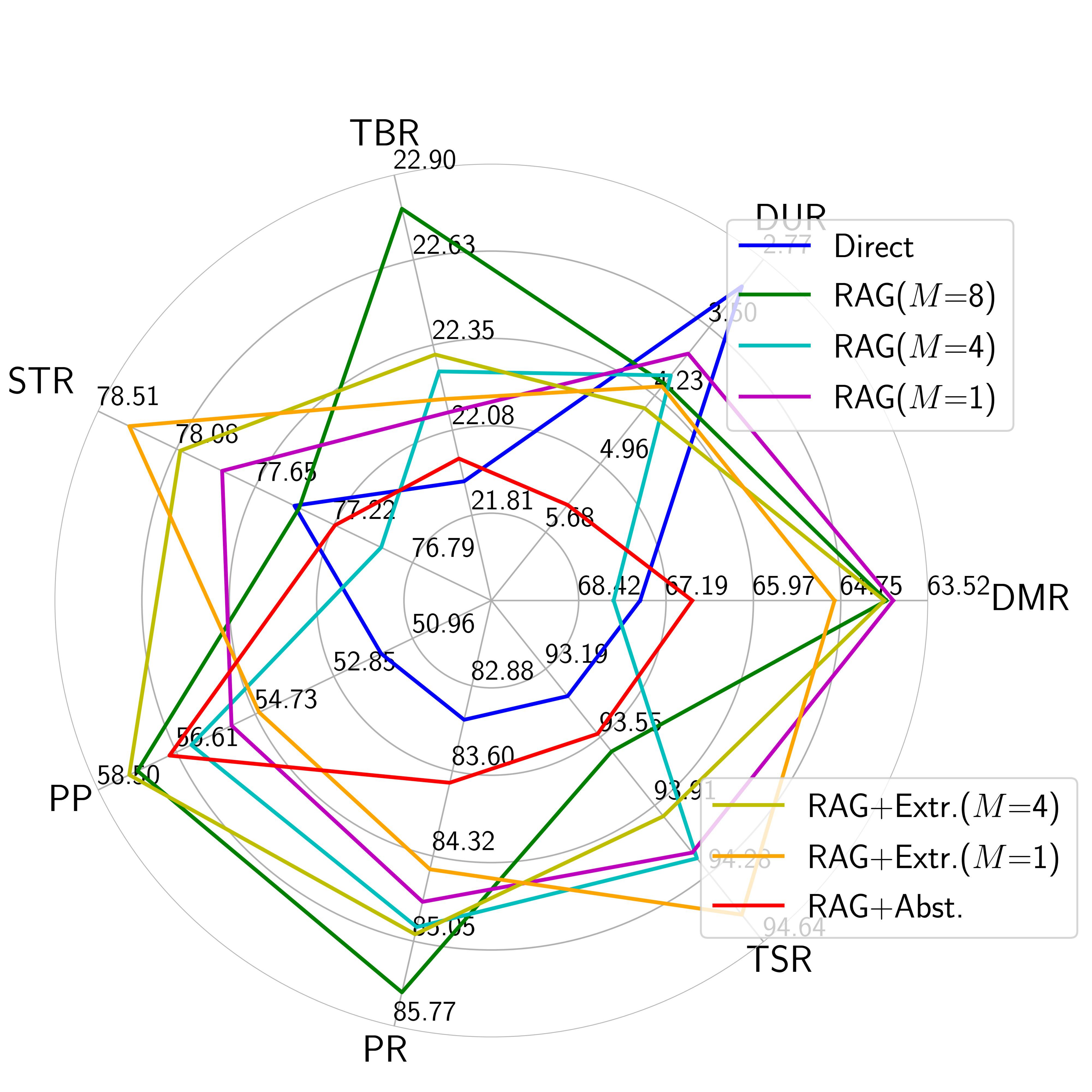}}
    \subfloat[Holiday.]{\includegraphics[width=0.25\textwidth]{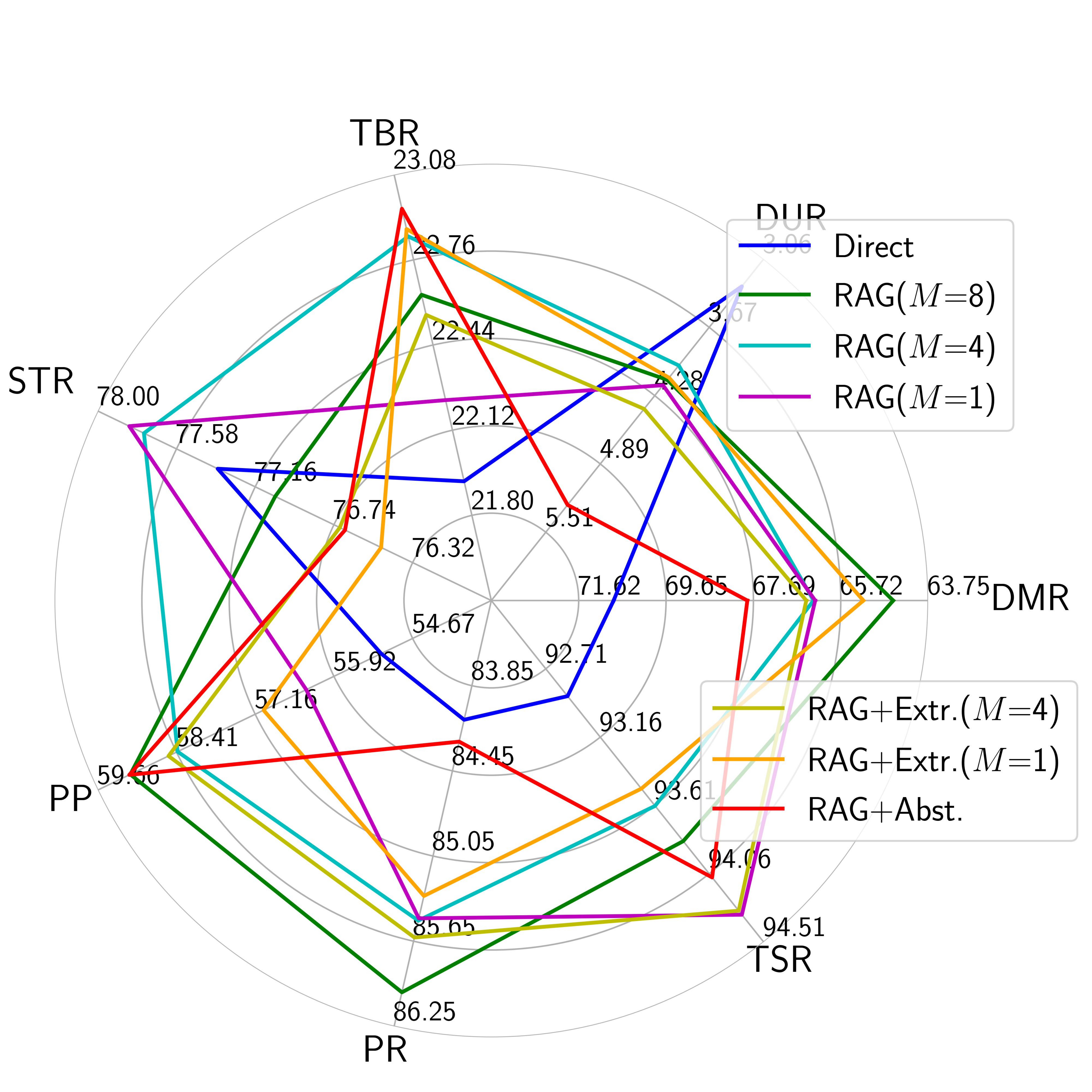}}
    \vspace{0.1pt}
    \subfloat[POI Category.]{\includegraphics[width=0.25\textwidth]{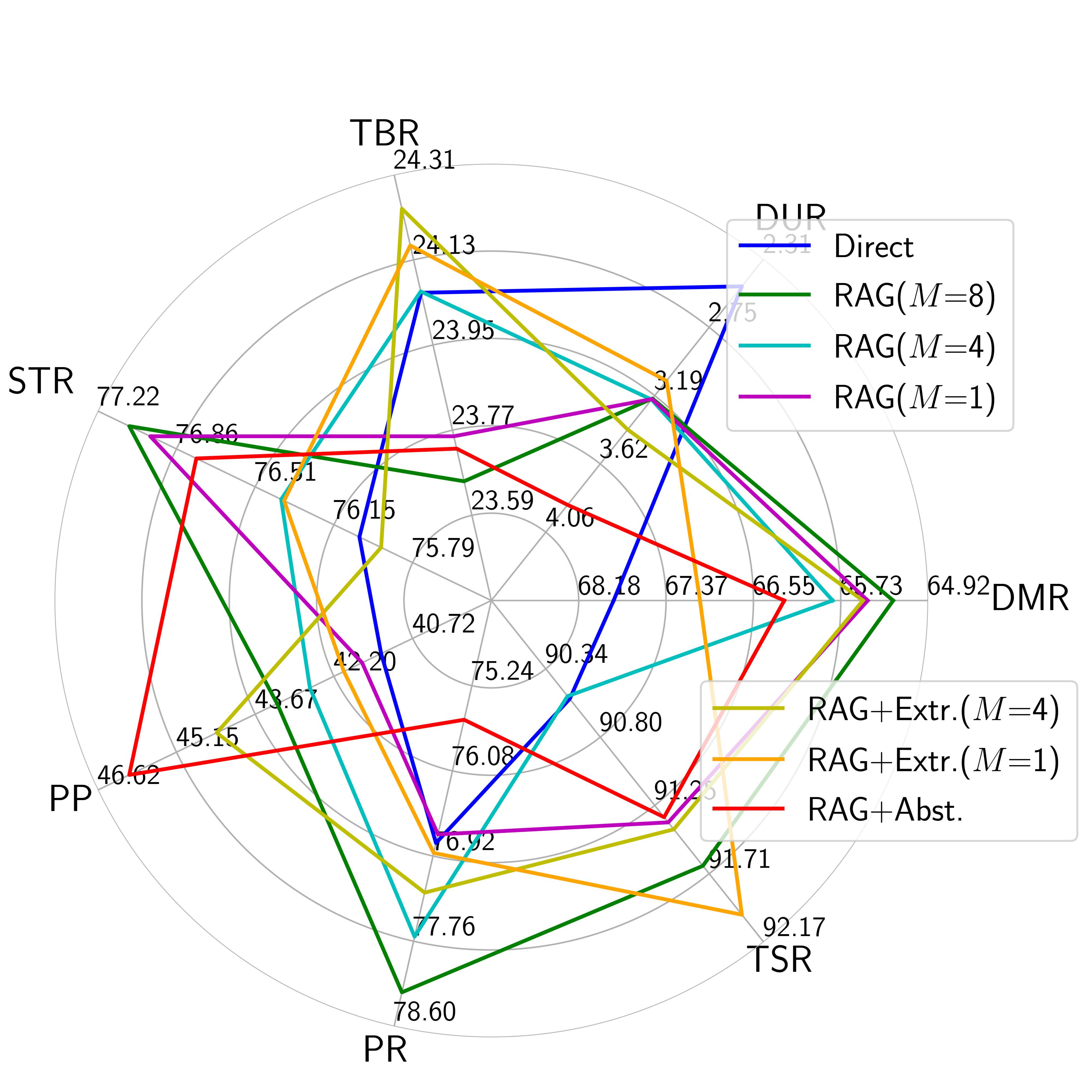}}
    \subfloat[Traveler Category.]{\includegraphics[width=0.25\textwidth]{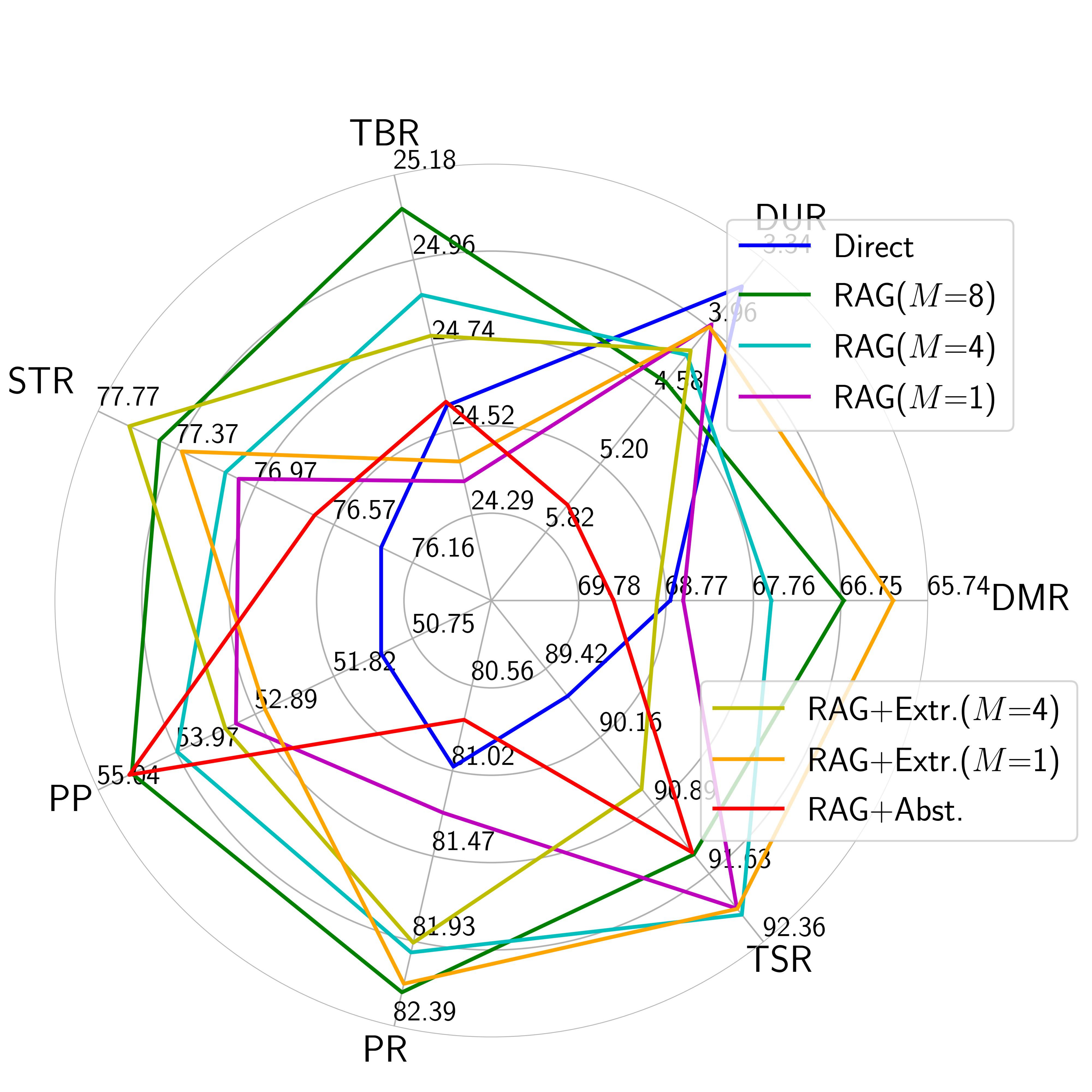}}
    \subfloat[Trip Compactness.]{\includegraphics[width=0.25\textwidth]{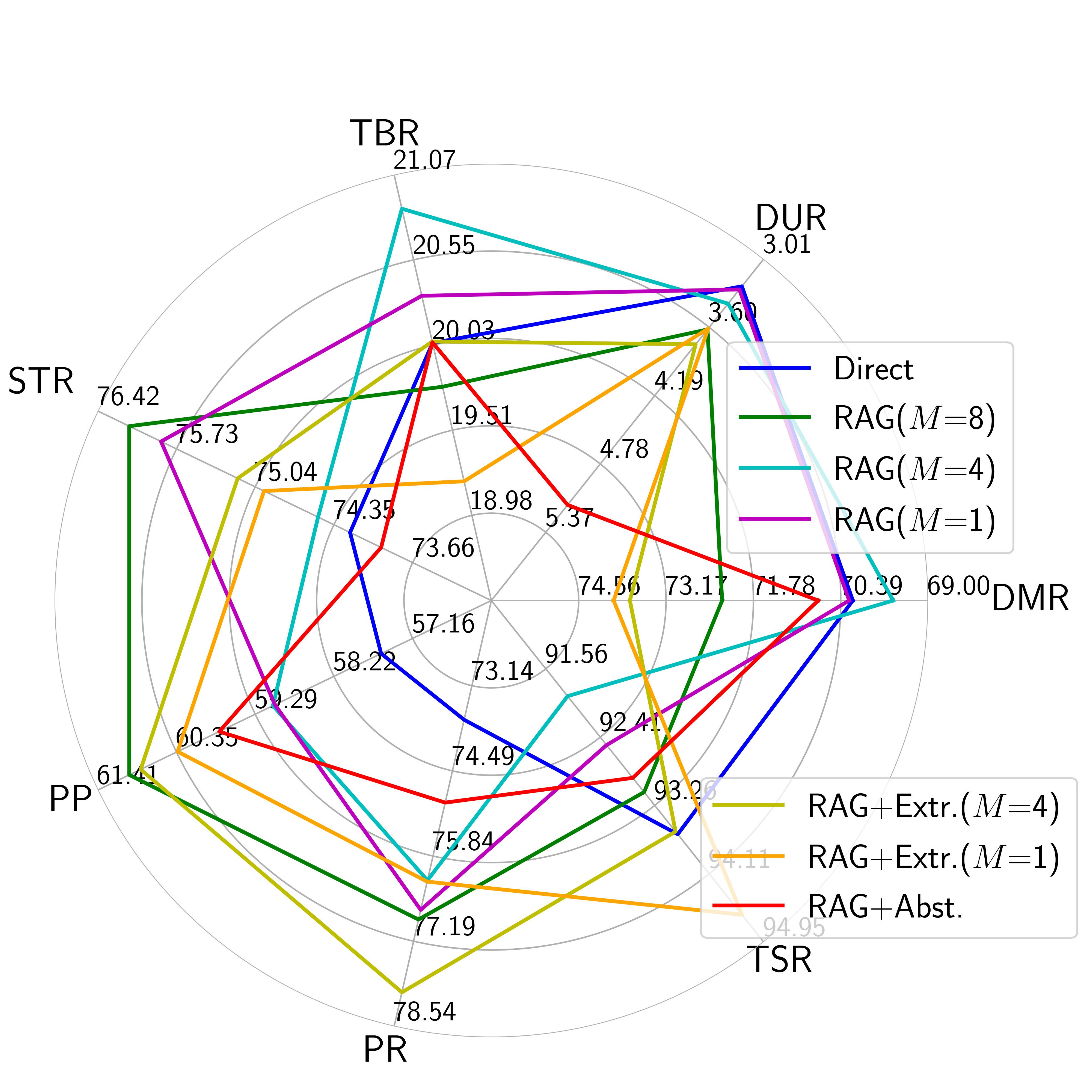}}
    \caption{Comparison between Direct and RAG strategies on query data with various categories, implemented by DeepSeek-R1-671B.}
    \vspace{-20pt}
    \label{figure:radar_deepseek}
\end{figure*}

We report our benchmark results on separate query data via radar charts, as illustrated in \figref{figure:radar_gpt}, \figref{figure:radar_qwen}, \figref{figure:radar_llama} and \figref{figure:radar_deepseek}.
The analysis reveals that retrieval-augmented strategies significantly outperform ground-up methods, such as direct prompting, in most evaluation metrics. 
This observation aligns with the findings of our full-scale experiments detailed in \secref{section:main}.

\subsection{Universality Analysis}
\label{appendix:universality}

We detail the statistics of win rates of retrieved-augmented planning methods over the Direct baseline in \tabref{table:necessity}, which is implemented with Qwen2.5-72B-Instruct model.

\begin{table}[H]
    \centering
    \vspace{-10pt}
    \caption{Win rate statistics of retrieval-augmented agents exceeding the Direct one, across various metrics.}
    \resizebox{0.525\linewidth}{!}{
    \begin{tabular}{c|ccccccc}
    \toprule
    \textbf{Metric} & \textbf{DMR} & \textbf{DUR} & \textbf{TBR} & \textbf{STR} & \textbf{PP} & \textbf{PR} & \textbf{TSR} \\ 
    \hline
    Win Rate & 80.96 & 83.30 & 81.37 & 87.69 & 91.35 & 89.57 & 97.67\\
    \bottomrule
    \end{tabular}}
    \label{table:necessity}
\end{table}

\subsection{Utilization Analysis}
\label{appendix:utilization}

To investigate how LLM agents utilize the trajectory knowledge, we design a similarity function $\sigma(s,t)=\beta\cdot \sigma_{\text{POI}}(s,t)+(1-\beta)\cdot \sigma_{\text{order}}(s,t)$ as a proxy that implies the extent of utilization for each trajectory.
$\sigma_{\text{POI}}$ denotes the Jaccard similarity of the POIs contained in the plan and trajectory, while $\sigma_{\text{order}}$ is the Kendall Tau coefficient between the ranks of common POIs in the plan and trajectory.
$\beta$ controls the balance between similarities of POI sets and POI order.

To represent the quality variable of the methods, we use POI semantic metric as a proxy since it shows the greatest improvements according to \tabref{table:main}, which facilitates the exploration into the mechanisms behind the superiority of retrieval-augmented methods.
Since the explicit extractive method RAG + Extr. ($M$=4)) directly determines the selected trajectory references, where the utilization indicator like similarity is inaccessible.
Thus we select the top-4 trajectory set according to the ranks of three variables: similarity, quality, and relevance, so as to contrast them with the set extracted by RAG + Extr. ($M$=4)).

\section{Methodology}
\label{appendix:method}

In this section, we elaborate the workflow of our method \textit{EvoRAG} in \algoref{algorithm:method}.
For efficiency, we set $G$ as 1.
The evaluate function $E$ is implemented by our evaluation system introduced in \secref{section:evaluation}.
And the complete experiment results are exhibited in \tabref{table:method}.

\begin{algorithm*}[H]
\SetAlgoNoLine  
\small
    \caption{The workflow of EvoRAG.}
    \KwIn{query $q$, POI candidates $P^q$, trajectories $T^q$; LLM agents $A_p,A_r,A_u^{mo},A_u^{cm}$ respectively for plan initialization, evaluation reflection, mutation-only plan updating and crossover-mutation plan updating, evaluation function $\mathcal{E}$; maximum number of optimization iterations $G$, mutation-only ratio $\alpha$;}
    \KwOut{The optimal plan $I^*$.}
    Initialize population $\mathcal{I}_0=\{I_{(0,i)}\}_{i=0}^{|T^q|}$ via $I_{(0,0)}=A_p(q,P^q)$, $I_{(0,i)}=\{A_p(q,P^q,t_i^q)\},\ i=1,2,...,|T^q|$; \\
    Initialize planning reflection memory $R_0$; \\
    Initialize $g \leftarrow 0$; \\
    \While{$g<G$}{
        1. Evaluate and rank the plans $E_g=[(I_{(g,i)}^o,e_{(g,i)})]_{i=0}^{|T^q|}=\mathcal{E}(\{I_{(g,i)}\}_{i=0}^{|T^q|})$ in the descending order of efficacy, where $e$ denotes the evaluation details; \\
        2. Reflect about the evaluation results and update the memory $R_{g+1}=A_r(R_g,E_g)$; \\
        3. Perform mutation for the top $\alpha$ proportion of population and generate $\mathcal{I}_{g+1,a}=\{I_{(g+1,i)})\}_{i=0}^{\alpha\cdot |T^q|}=A_u^{mo}(q,P^q,R_{g+1},[(I_{(g,i)}^o,e_{(g,i)})]_{i=0}^{\alpha\cdot |T^q|})$; \\
        4. Perform crossover and mutation for population and generate $\mathcal{I}_{g+1,b}=\{I_{(g+1,i)})\}_{i=0}^{(1-\alpha)\cdot |T^q|}=A_u^{cm}(q,P^q,R_{g+1},E_g)$; \\
        5. $\mathcal{I}_{g+1}=\text{Union}(\mathcal{I}_{g+1,a},\mathcal{I}_{g+1,b})$; \\
        6. $g=g+1$;\\
    }
    Evaluate and rank the plans $E_G=[(I_{(G,i)}^o,e_{(G,i)})]_{i=0}^{|T^q|}=\mathcal{E}(\{I_{(G,i)}\}_{i=0}^{|T^q|})$ in the descending order of efficacy; \\
    Evaluate and rank the best plans from all the iterations $E^*=[(s_i^*,e_i^*)]_{i=0}^G=\mathcal{E}(\{I_{(i,0)}^o\}_{i=0}^G)$;\\
    \Return $I^*=I_0^*$.\\
    \label{algorithm:method}
\end{algorithm*}

\begin{table*}[h!]
    \centering
    \caption{Comparison between EvoRAG and baseline methods on our dataset implemented by Qwen2.5-72B-Instruct. 
    }
    \vspace{-8pt}
    \resizebox{\linewidth}{!}{
    \begin{tabular}{l|cc|ccccccc|cccc|c}
    \toprule
    \textbf{Method} & \textbf{FR$\downarrow$} & \textbf{RR$\downarrow$} & \textbf{DMR$\downarrow$} & \textbf{DUR$\downarrow$} & \textbf{TBR$\uparrow$} & \textbf{STR$\uparrow$} & \textbf{PP$\uparrow$} & \textbf{PR$\uparrow$} & \textbf{TSR$\uparrow$} & $R_S\downarrow$ & $R_T\downarrow$ & $R_P\downarrow$ & $R_R\downarrow$ & $R_C\downarrow$ \\ 
    \hline
    Direct & \textbf{0.38} & \textbf{0.03} & 71.67 & 6.49 & 24.82 & 78.33 & 48.53 & 79.68 & 92.86 & 9.00 & 5.67 & 9.00 & 9.50 & 8.29 \\
    CoT & 0.42 & \underline{0.04} & 70.51 & 6.64 & 24.66 & 78.77 & 47.09 & 80.12 & 93.16 & 8.00 & 6.33 & 11.00 & 8.50 & 8.46 \\
    Reflextion & 2.39 & 1.62 & 85.38 & 8.08 & 25.37 & 77.54 & 49.15 & 79.65 & 92.14 & 11.00 & 8.67 & 8.00 & 11.00 & 9.67 \\
    MAC & 1.10 & 2.21 & 70.08 & \textbf{5.74} & 23.65 & 76.48 & 43.3 & 81.28 & 90.00 & 7.00 & 8.00 & 12.00 & 7.50 & 8.62 \\
    MAD & 3.30 & 1.49 & 87.47 & 9.46 & \underline{26.53} & 77.75 & 47.49 & 80.23 & 91.21 & 12.00 & 8.00 & 10.00 & 10.00 & 10.00 \\
    \hdashline[0.5pt/1pt]
    RAG($M$=8) & 3.41 & 0.11 & 69.39 & \underline{6.35} & 23.48 & 78.40 & 55.15 & \underline{81.67} & 93.29 & 6.00 & 6.67 & 3.00 & \underline{3.50} & 4.79 \\
    RAG($M$=4) & 3.15 & 0.11 & 68.77 & 6.54 & 24.06 & \underline{79.08} & 53.62 & 81.26 & \textbf{93.86} & 4.00 & 5.67 & 5.00 & \textbf{2.50} & \underline{4.29} \\
    RAG($M$=1) & 2.58 & 0.09 & \underline{68.07} & 6.89 & 25.27 & 78.48 & 51.62 & 81.07 & 93.40 & \underline{2.00} & 6.33 & 7.00 & 5.00 & 5.08 \\
    RAG+Extr.($M$=4) & 3.46 & 0.16 & 69.03 & 6.49 & 24.33 & 79.03 & 54.79 & 81.21 & \underline{93.81} & 5.00 & \underline{5.00} & 4.00 & \underline{3.50} & 4.38 \\
    RAG+Extr.($M$=1) & 3.16 & 0.20 & 68.29 & 6.73 & 25.59 & 78.33 & 52.36 & 80.92 & 93.42 & 3.00 & 6.00 & 6.00 & 5.00 & 5.00 \\
    RAG+Abst. & 3.78 & 0.17 & 72.40 & 7.32 & 25.11 & 78.04 & \underline{56.14} & 80.50 & 93.29 & 10.00 & 8.33 & \underline{2.00} & 6.50 & 6.71 \\
    EvoRAG & \underline{0.40} & 0.06 & \textbf{44.45} & 6.54 & \textbf{28.22} & \textbf{81.63} & \textbf{58.05} & \textbf{85.82} & 92.19 & \textbf{1.00} & \textbf{2.33} & \textbf{1.00} & 5.00 & \textbf{2.33} \\
    \bottomrule
    \end{tabular}}
    \label{table:method}
\end{table*}

\section{Prompt Templates}

\tcbset{
    width=\textwidth, 
    colback=gray!10, 
    colframe=black, 
    coltitle=white, 
    fonttitle=\bfseries, 
    arc=3mm, 
    colbacktitle=gray!120, 
    boxsep=5pt, 
    boxrule=0.8pt, 
    left=0pt, 
    right=0pt, 
    breakable,
    enhanced jigsaw,
    fontupper=\ttfamily, 
}

\subsection{Data Generation}
\label{appendix:prompt_data_generation}

\begin{tcolorbox}[title=Query Formulation]
\small
Based on the seed query examples, please create a standard for query formulation, \ie which fundamental elements the query may include, as well as which potential words can represent these elements.\\

<SEED QUERIES>\\

Answer Format in JSON:\\
\texttt{\{"Element 1": ["Potential Word 1", "Potential Word 2", \ldots], "Element 2": ["Potential Word 1", "Potential Word 2", \ldots], \ldots\}}
\end{tcolorbox}

\begin{tcolorbox}[title=Query Generation]
\small
Based on the selected popular cities, please generate <NUMBER OF QUERIES> new natural language queries, that adhere to the standard of query formulation.\\

<POPULAR CITIES>\\

<QUERY FORMULATION>\\

Answer Format in JSON:\\
\texttt{["Query 1", "Query 2", \ldots]}
\end{tcolorbox}

\begin{tcolorbox}[title=POI-level Query Rewriting]
\small
Based on the user query <QUERY>, please generate <NUMBER OF SUB-QUERIES> sub-queries that retain the meaning of the original query while facilitate retrieving Web documents about POI recommendations via the search engine.
In addition, the generated sub-queries must be diverse.\\

Answer Format in JSON:\\
\texttt{["Sub-query 1", "Sub-query 2", \ldots]}
\end{tcolorbox}

\begin{tcolorbox}[title=POI Extraction]
\small
Based on the following <NUMBER OF DOCUMENTS> retrieved documents, please identify and extract all related tourist attraction Points of Interest (POIs).\\

Tourist Attraction POI Definition:\\
Tourist attraction POIs include natural scenic areas, historical sites, cultural landmarks, parks, museums, commercial streets, resorts, theme parks, amusement parks, zoos and botanical gardens, specialty malls, cinemas, temples, palaces, \etc.\\

POI Extraction Requirements:\\
1. Extraction: Use your knowledge and reasoning abilities to identify and extract all tourist attraction-related POIs from the documents.\\
2. Explanation: For each extracted POI, provide a brief explanation of why it is considered a tourist attraction POI.\\

Special Notes:\\
1. Ensure the extraction is thorough, and avoid missing any potential POIs. Include every tourist attraction POI mentioned in the documents. 
Extract at least 10 and up to 30 POIs.\\
2. All POIs must come from the documents. Do not fabricate any POIs. 
Specify the exact source by referencing the relevant document ID.\\
3. Do not extract streets, roads, public facilities, broad geographic regions or city names.\\
4. Each POI name should represent a single tourist attraction. 
Avoid connecting multiple POIs with a hyphen ("-") or extracting duplicates or overlapping POIs.\\

<DOCUMENTS>\\

Answer Format in JSON:\\
\texttt{[\{"Extraction reason": "xxx", "Source": ["Document ID", \ldots], "POI name": "xxx"\}, \ldots]}
\end{tcolorbox}

\begin{tcolorbox}[title=POI Inquiry Construction]
\small
Based on the user query <QUERY>, please generate <NUMBER OF QUERIES> queries for a specific attraction POI <POI>, that are relevant to the original query while facilitate retrieving real-time Web documents related to this POI via the search engine.
In addition, the generated sub-queries must be diverse.\\

Answer Format in JSON:\\
\texttt{["Query 1", "Query 2", \ldots]}
\end{tcolorbox}

\begin{tcolorbox}[title=Temporal and Semantic Tagging]
\small
Based on the retrieved documents, please analyze several things for the POI <POI>: (1) Opening Hours; (2) Recommended Visit Time; (3) Expected Visit Duration; (4) POI Description with Special Notes.\\

Requirements:
1. Provide intermediate analysis and reasoning steps.\\
2. Opening hours should specify the specific opening hours of the attraction in 24-hour format, such as "9:00-14:00" or "0:00-24:00" for all-day opening.
Recommended visit time should offer specific recommended arrival times, also in 24-hour format. 
If there are no time restrictions, indicate "Open All Day." 
Expected visit duration should be in hours, such as "3" and "4.5".
Description must be brief, not exceeding 50 words.\\

<DOCUMENTS>\\

Answer Format in JSON:\\
\texttt{\{"Reasoning steps": "xxx", "Opening hours": "xxx", "Recommended visit time": "xxx", "Expected visit duration": "xxx", "POI description": "xxx"\}}
\end{tcolorbox}

\begin{tcolorbox}[title=Trajectory-level Query Rewriting]
\small
Based on the user query <QUERY>, please generate <NUMBER OF SUB-QUERIES> sub-queries that retain the meaning of the original query while facilitate retrieving Web documents about travel guides via the search engine.
In addition, the generated sub-queries must be diverse.\\

Answer Format in JSON:\\
\texttt{["Sub-query 1", "Sub-query 2", \ldots]}
\end{tcolorbox}

\begin{tcolorbox}[title=Trajectory Extraction]
\small
Based on the given document below, please identify and extract the exact tourist trajectory, which consists of attraction POIs and is relevant to the user query <QUERY>.\\

Requirements:\\
1. Determine whether the given document contains a clear travel route. If not, respond with "None".\\
2. Organize the trajectory according to the number of travel days, with subtitles like "Day 1", "Day 2", and "Day 3".
For each day, the itinerary should not be empty, especially the first and last day. 
If there are multiple trajectories in the given plan, choose one to extract and do not mix multiple solutions together.\\
3. In each extracted trajectory, the tourist attraction POIs and the order of visit must strictly follow the document. 
Do not hallucinate!\\
4. The attractions in the trajectory should be answered using the standardized POI names from the given POI reference list.
If a POI is not listed in the reference list, it should still be included but with an special note.\\
5. Verify newly added attractions to ensure they are genuine tourist attractions, excluding non-POI items such as "return journey", "flag-raising ceremony", "free time", \etc.\\
6. Summarize additional information about the attraction mentioned in the trajectory in a remark section. 
If there is no additional information, leave it blank. 
Do not extract information from the POI reference list for the remarks; it must be from the document.\\

<DOCUMENT>\\

<POI REFERENCE LIST>\\

Answer Format in JSON:\\
\texttt{\{"Day 1": [\{"POI name": "xxx", "Whether in the reference list": "xxx", "Remark": "xxx"\}, \ldots], \ldots\} or \{"None": "None\}}
\end{tcolorbox}

\begin{tcolorbox}[title=POI Quality Control]
\small
Given the initial list of tourist attraction POIs, please purify these POIs.\\

POI Purification Requirements:\\
1. Check whether the attraction POI is included in the user-specified city <CITY>.
If not, please remove them.\\
2. Based on your understanding of these tourist attraction POIs, recheck if they are all indeed tourist attractions. 
If not, please remove non-tourist attraction POIs such as restaurants, hotels, specific leisure and entertainment venues (arcades, spas, cinemas, \etc).\\ 
3. Remove duplicate tourist attraction POIs, keeping only one instance. 
For example, if "Tiananmen" and "Tiananmen Square" both appear, keep "Tiananmen". 
If "Wangfujing" and "Wangfujing Street" both appear, keep "Wangfujing". 
If "National Aquatics Center" and "National Aquatics Center - North Entrance" both appear, keep "National Aquatics Center".\\
4. Some POIs may have a hierarchical relationship, such as "Summer Palace" and "Kunming Lake". 
In such cases, only keep "Summer Palace" and remove "Kunming Lake".\\
5. Filter the list from the initial tourist attraction POI list, and refrain from fabricating attractions.\\

<INITIAL POI LIST>\\

Please provide a textual explanation for the discarded tourist attraction POIs and respond in JSON format with the final processed list of purified POIs.\\

Answer Format in JSON:\\
\texttt{["POI name 1", "POI name 2", \ldots]}
\end{tcolorbox}

\begin{tcolorbox}[title=Trajectory Quality Control]
\small
There is a tourist trajectory with a lot of noise (false scenic POIs, incomplete or incorrect names of scenic POIs). 
Please denoise the existing trajectory based on the standard name list of POIs.\\

Requirements:\\
1. If the existing trajectory contains non-attraction POIs that do not exist in the standard name list, consider making modifications or deletions:\\
(1) POI Modification: The names of POIs in the existing planning may be incomplete, non-standard, or inaccurate. 
Please find a standard and accurate name from the standard name list as a replacement. 
For example, change “Aosen Park” to “Olympic Forest Park”.\\
(2) POI Deletion: If there is no suitable replacement in the standard name list, please directly delete the POI from the current trajectory.\\
2. Ensure that the names of POIs in the modified trajectory are all in the standard name list. Prohibit the addition of new unrelated scenic POIs during the modification process.\\

<STANDARD POI INFORMATION>\\

<CURRENT TRAJECTORY>\\

Answer Format in JSON:\\
\texttt{\{"Day 1": [\{"POI name": "xxx", "Remark": "xxx"\}, \ldots], \ldots\}}
\end{tcolorbox}

\subsection{Evaluation}
\label{appendix:prompt_evaluation}

\begin{tcolorbox}[title=Start Time Rationality (STR) Evaluation]
\small
Given the user query <QUERY>, associated reference information, and the planned start visit times for attraction POIs, evaluate whether the arrival time is reasonable. 

Requirements:
1. Judge whether the start visit time falls into the opening hours.\\
2. Judge whether the start visit time conforms to the recommended arrival time.\\
3. Answer "Appropriate" or "Inappropriate" for each POI.\\
4. Do not omit any POI.\\

<REFERENCE TEMPORAL INFORMATION OF POIS>\\

<PLANNED START VISIT TIMES OF POIS>\\

Please give brief textual explanations of why the time slots for the attractions that are deemed inappropriate are not suitable, and provide the final evaluated results.\\

Answer Format in JSON:\\
\texttt{\{"POI name 1": "xxx", "POI name 2": "xxx", \ldots\}}
\end{tcolorbox}

\begin{tcolorbox}[title=POI Relevance (PR) Evaluation]
\small
Given the user query <QUERY> and the planned attraction POIs, evaluate whether the POI satisfies the personalized demands in the query. 
Answer "Satisfied" or "Unsatisfied" for each POI.
Do not omit any POI.\\

<PLANNED POIS>\\

Please give brief textual explanations for POIs that are deemed unsatisfied, and provide the final evaluated results.\\

Answer Format in JSON:\\
\texttt{\{"POI name 1": "xxx", "POI name 2": "xxx", \ldots\}}
\end{tcolorbox}

\begin{tcolorbox}[title=Time Scheduling (TSR) Evaluation]
\small
Given the user query <QUERY> and the planned time slots of attraction POIs, evaluate whether the scheduled time slot satisfies the personalized demands in the query. 
Answer "Satisfied" or "Unsatisfied" for each POI.
Do not omit any POI.\\

<PLANNED TIME SLOTS of POIS>\\

Please give brief textual explanations for POIs' time slots that are deemed unsatisfied, and provide the final evaluated results.\\

Answer Format in JSON:\\
\texttt{\{"POI name 1": "xxx", "POI name 2": "xxx", \ldots\}}
\end{tcolorbox}

\subsection{Baseline}
\label{appendix:prompt_baseline}

\begin{tcolorbox}[title=Direct]
\small
Our task is to generate a travel plan based on the query <QUERY> and associated POI references.\\

Basic Requirements:\\
1. Structure the article according to the number of days, such as "Day 1", "Day 2", and "Day 3". 
If the query does not specify the number of days, use your knowledge and the attractions list to deduce the duration of the travel plan.\\
2. Plan the visit to attractions POI in the order of scheduled visit times. 
Select attractions only from the provided reference list, and do not include attractions outside the list.\\
3. Plan specific start and end times of visit in 24-hour format for each POI. 
Ensure no overlap in visit times for different POIs, leave gaps between activities.\\

More Planning Requirements:\\
a. General Requirements:\\
1. Spatial: Consider the clustering of attraction POIs based on their geographical locations. 
Assign attraction POIs with close geographical locations to the same day and those with distant locations to different days. 
Ensure attractions on the same day are not too far apart.\\
2. Time:\\
(1) Each POI must be visited during its opening hours. Prioritize the recommended start times for attractions and ensure sufficient time for each visit (based on the expected duration of the attraction).\\
(2) In general, the total travel schedule for each day should not be too tight, ensuring the overall travel time is not too long, the number of attractions visited is not too large, and there is enough free time for meals, accommodation, and transportation.\\
3. Attractions Semantics: In general, prioritize popular and unique attractions that reflect the city’s characteristics.\\
b. Personalized Requirements:\\
1. If there are additional personalized constraints in the query, understand and summarize these requirements. 
When selecting attractions and planning the plan, consider these personalized constraints.\\
2. Some personalized requirements may conflict with general requirements. 
In such cases, prioritize the personalized requirements.
For example, if the query is "Special Forces-style Tourist", the overall itinerary time, number of attractions visited per day, free time, duration of attraction visits, and start times of visits may not have specific constraints, allowing for a more compact itinerary.
If the query is related to specific demands (\eg "Natural Landscape Tourism", "Family Travel", "Autumn Travel", "Chinese New Year Travel", \etc.), just choose the most popular attractions that meet the query constraints.\\

<POI REFERENCE LIST>\\

Answer Format in JSON:\\
\texttt{\{"Day 1": [\{"POI name": "xxx", "Start visit time": "xxx", "End visit time": "xxx"\}, \ldots], \ldots\}, \ldots}
\end{tcolorbox}

\begin{tcolorbox}[title=Chain of Thought (CoT)]
\small
Our task is to generate a travel plan based on the query <QUERY> and associated POI references.\\

Basic Requirements:\\
1. Structure the article according to the number of days, such as "Day 1", "Day 2", and "Day 3". 
If the query does not specify the number of days, use your knowledge and the attractions list to deduce the duration of the travel plan.\\
2. Plan the visit to attractions POI in the order of scheduled visit times. 
Select attractions only from the provided reference list, and do not include attractions outside the list.\\
3. Plan specific start and end times of visit in 24-hour format for each POI. 
Ensure no overlap in visit times for different POIs, leave gaps between activities.\\

More Planning Requirements:\\
a. General Requirements:\\
1. Spatial: Consider the clustering of attraction POIs based on their geographical locations. 
Assign attraction POIs with close geographical locations to the same day and those with distant locations to different days. 
Ensure attractions on the same day are not too far apart.\\
2. Time:\\
(1) Each POI must be visited during its opening hours. Prioritize the recommended start times for attractions and ensure sufficient time for each visit (based on the expected duration of the attraction).\\
(2) In general, the total travel schedule for each day should not be too tight, ensuring the overall travel time is not too long, the number of attractions visited is not too large, and there is enough free time for meals, accommodation, and transportation.\\
3. Attractions Semantics: In general, prioritize popular and unique attractions that reflect the city’s characteristics.\\
b. Personalized Requirements:\\
1. If there are additional personalized constraints in the query, understand and summarize these requirements. 
When selecting attractions and planning the plan, consider these personalized constraints.\\
2. Some personalized requirements may conflict with general requirements. 
In such cases, prioritize the personalized requirements.
For example, if the query is "Special Forces-style Tourist", the overall itinerary time, number of attractions visited per day, free time, duration of attraction visits, and start times of visits may not have specific constraints, allowing for a more compact itinerary.
If the query is related to specific demands (\eg "Natural Landscape Tourism", "Family Travel", "Autumn Travel", "Chinese New Year Travel", \etc.), just choose the most popular attractions that meet the query constraints.\\

<POI REFERENCE LIST>\\

Please provide a step-by-step plan to solve the problem, and then present the final plan.\\

Answer Format in JSON:\\
\texttt{\{"Day 1": [\{"POI name": "xxx", "Start visit time": "xxx", "End visit time": "xxx"\}, \ldots], \ldots\}, \ldots}
\end{tcolorbox}

For plan initialization of Reflextion method, we directly utilize the template of Direct.

\begin{tcolorbox}[title=Reflextion (Feedback)]
\small
Our task is to generate a travel plan based on the query <QUERY> and associated POI references.\\

Planning Requirements:\\
a. General Requirements:\\
1. Spatial: Consider the clustering of attraction POIs based on their geographical locations. 
Assign attraction POIs with close geographical locations to the same day and those with distant locations to different days. 
Ensure attractions on the same day are not too far apart.\\
2. Time:\\
(1) Each POI must be visited during its opening hours. Prioritize the recommended start times for attractions and ensure sufficient time for each visit (based on the expected duration of the attraction).\\
(2) In general, the total travel schedule for each day should not be too tight, ensuring the overall travel time is not too long, the number of attractions visited is not too large, and there is enough free time for meals, accommodation, and transportation.\\
3. Attractions Semantics: In general, prioritize popular and unique attractions that reflect the city’s characteristics.\\
b. Personalized Requirements:\\
1. If there are additional personalized constraints in the query, understand and summarize these requirements. 
When selecting attractions and planning the plan, consider these personalized constraints.\\
2. Some personalized requirements may conflict with general requirements. 
In such cases, prioritize the personalized requirements.
For example, if the query is "Special Forces-style Tourist", the overall itinerary time, number of attractions visited per day, free time, duration of attraction visits, and start times of visits may not have specific constraints, allowing for a more compact itinerary.
If the query is related to specific demands (\eg "Natural Landscape Tourism", "Family Travel", "Autumn Travel", "Chinese New Year Travel", \etc.), just choose the most popular attractions that meet the query constraints.\\

There is an initial plan in place, please review whether this plan meets the requirements for attraction planning and provide specific feedback for modifications.\\

<POI REFERENCE LIST>\\

<INITIAL PLAN>
\end{tcolorbox}

\begin{tcolorbox}[title=Reflextion (Refinement)]
\small
Our task is to generate a travel plan based on the query <QUERY> and associated POI references.\\

Basic Requirements:\\
1. Structure the article according to the number of days, such as "Day 1", "Day 2", and "Day 3". 
If the query does not specify the number of days, use your knowledge and the attractions list to deduce the duration of the travel plan.\\
2. Plan the visit to attractions POI in the order of scheduled visit times. 
Select attractions only from the provided reference list, and do not include attractions outside the list.\\
3. Plan specific start and end times of visit in 24-hour format for each POI. 
Ensure no overlap in visit times for different POIs, leave gaps between activities.\\

More Planning Requirements:\\
a. General Requirements:\\
1. Spatial: Consider the clustering of attraction POIs based on their geographical locations. 
Assign attraction POIs with close geographical locations to the same day and those with distant locations to different days. 
Ensure attractions on the same day are not too far apart.\\
2. Time:\\
(1) Each POI must be visited during its opening hours. Prioritize the recommended start times for attractions and ensure sufficient time for each visit (based on the expected duration of the attraction).\\
(2) In general, the total travel schedule for each day should not be too tight, ensuring the overall travel time is not too long, the number of attractions visited is not too large, and there is enough free time for meals, accommodation, and transportation.\\
3. Attractions Semantics: In general, prioritize popular and unique attractions that reflect the city’s characteristics.\\
b. Personalized Requirements:\\
1. If there are additional personalized constraints in the query, understand and summarize these requirements. 
When selecting attractions and planning the plan, consider these personalized constraints.\\
2. Some personalized requirements may conflict with general requirements. 
In such cases, prioritize the personalized requirements.
For example, if the query is "Special Forces-style Tourist", the overall itinerary time, number of attractions visited per day, free time, duration of attraction visits, and start times of visits may not have specific constraints, allowing for a more compact itinerary.
If the query is related to specific demands (\eg "Natural Landscape Tourism", "Family Travel", "Autumn Travel", "Chinese New Year Travel", \etc.), just choose the most popular attractions that meet the query constraints.\\

There is an initial travel plan in place, as well as feedback for modifications to the plan. 
Please generate a revised plan in the same format as the original plan.\\

<POI REFERENCE LIST>\\

<INITIAL PLAN>\\

<FEEDBACK>\\

Answer Format in JSON:\\
\texttt{\{"Day 1": [\{"POI name": "xxx", "Start visit time": "xxx", "End visit time": "xxx"\}, \ldots], \ldots\}}
\end{tcolorbox}

Multi-Agent Collaboration (MAC) applies a divide-and-conquer paradigm. 
First, a manager agent decomposes the planning problem into several sub-problems.
The executor agents strive to address these sub-problems independently and the sub-solutions are finally summarized by the manager to directly answer the original question.

\begin{tcolorbox}[title=Multi-Agent Collaboration (MAC) - Manager Agent (Decomposition)]
\small
Our task is to generate a travel plan based on the query <QUERY> and associated POI references.\\

Basic Requirements:\\
1. Structure the article according to the number of days, such as "Day 1", "Day 2", and "Day 3". 
If the query does not specify the number of days, use your knowledge and the attractions list to deduce the duration of the travel plan.\\
2. Plan the visit to attractions POI in the order of scheduled visit times. 
Select attractions only from the provided reference list, and do not include attractions outside the list.\\
3. Plan specific start and end times of visit in 24-hour format for each POI. 
Ensure no overlap in visit times for different POIs, leave gaps between activities.\\

More Planning Requirements:\\
a. General Requirements:\\
1. Spatial: Consider the clustering of attraction POIs based on their geographical locations. 
Assign attraction POIs with close geographical locations to the same day and those with distant locations to different days. 
Ensure attractions on the same day are not too far apart.\\
2. Time:\\
(1) Each POI must be visited during its opening hours. Prioritize the recommended start times for attractions and ensure sufficient time for each visit (based on the expected duration of the attraction).\\
(2) In general, the total travel schedule for each day should not be too tight, ensuring the overall travel time is not too long, the number of attractions visited is not too large, and there is enough free time for meals, accommodation, and transportation.\\
3. Attractions Semantics: In general, prioritize popular and unique attractions that reflect the city’s characteristics.\\
b. Personalized Requirements:\\
1. If there are additional personalized constraints in the query, understand and summarize these requirements. 
When selecting attractions and planning the plan, consider these personalized constraints.\\
2. Some personalized requirements may conflict with general requirements. 
In such cases, prioritize the personalized requirements.
For example, if the query is "Special Forces-style Tourist", the overall itinerary time, number of attractions visited per day, free time, duration of attraction visits, and start times of visits may not have specific constraints, allowing for a more compact itinerary.
If the query is related to specific demands (\eg "Natural Landscape Tourism", "Family Travel", "Autumn Travel", "Chinese New Year Travel", \etc.), just choose the most popular attractions that meet the query constraints.\\

Please do not directly answer this question, but carefully consider how to break down the problem and plan the execution sequence.\\
1. Break down the original planning problem into several subproblems (up to four), and detail the planning requirements for each subproblem.\\
2. The names of the subproblems correspond to the order of execution, with later subproblems taking the outputs of the previous subproblems as inputs.\\

Answer Format in JSON:\\
\texttt{\{"Sub-problem 1": \{"Sub-problem description": "xxx", "Planning requirements": ["xxx", "xxx", \ldots]\}, \ldots\}}
\end{tcolorbox}

\begin{tcolorbox}[title=Multi-Agent Collaboration (MAC) - Executor Agent]
\small
Our task is to generate a travel plan based on the query <QUERY> and associated POI references.\\

<POI REFERENCE LIST>\\

Please do not directly answer this question, but carefully solve one sub-problem about it as follows:\\
<SUB-PROBLEM INFORMATION>\\

The outputs of previous subproblem:\\
<PREVIOUS OUTPUTS>\\

Please answer according to the requirements of the subproblem (answer in JSON format, keep it brief, not exceeding 500 characters).
\end{tcolorbox}

\begin{tcolorbox}[title=Multi-Agent Collaboration (MAC) - Manager Agent (Summarization)]
\small
Our task is to generate a travel plan based on the query <QUERY> and associated POI references.\\

Basic Requirements:\\
1. Structure the article according to the number of days, such as "Day 1", "Day 2", and "Day 3". 
If the query does not specify the number of days, use your knowledge and the attractions list to deduce the duration of the travel plan.\\
2. Plan the visit to attractions POI in the order of scheduled visit times. 
Select attractions only from the provided reference list, and do not include attractions outside the list.\\
3. Plan specific start and end times of visit in 24-hour format for each POI. 
Ensure no overlap in visit times for different POIs, leave gaps between activities.\\

More Planning Requirements:\\
a. General Requirements:\\
1. Spatial: Consider the clustering of attraction POIs based on their geographical locations. 
Assign attraction POIs with close geographical locations to the same day and those with distant locations to different days. 
Ensure attractions on the same day are not too far apart.\\
2. Time:\\
(1) Each POI must be visited during its opening hours. Prioritize the recommended start times for attractions and ensure sufficient time for each visit (based on the expected duration of the attraction).\\
(2) In general, the total travel schedule for each day should not be too tight, ensuring the overall travel time is not too long, the number of attractions visited is not too large, and there is enough free time for meals, accommodation, and transportation.\\
3. Attractions Semantics: In general, prioritize popular and unique attractions that reflect the city’s characteristics.\\
b. Personalized Requirements:\\
1. If there are additional personalized constraints in the query, understand and summarize these requirements. 
When selecting attractions and planning the plan, consider these personalized constraints.\\
2. Some personalized requirements may conflict with general requirements. 
In such cases, prioritize the personalized requirements.
For example, if the query is "Special Forces-style Tourist", the overall itinerary time, number of attractions visited per day, free time, duration of attraction visits, and start times of visits may not have specific constraints, allowing for a more compact itinerary.
If the query is related to specific demands (\eg "Natural Landscape Tourism", "Family Travel", "Autumn Travel", "Chinese New Year Travel", \etc.), just choose the most popular attractions that meet the query constraints.\\

<POI REFERENCE LIST>\\

You have already thought about how to break down this problem and got answers to each sub-problem as follows:\\
<SUB-PROBLEM OUTPUTS>\\

Now please solve the original planning problem.\\

Answer Format in JSON:\\
\texttt{\{"Day 1": [\{"POI name": "xxx", "Start visit time": "xxx", "End visit time": "xxx"\}, \ldots], \ldots\}}
\end{tcolorbox}

Multi-Agent Debate (MAD) launches a discussion session allowing criticism agents with different perspectives to give feedback for the initial plan generated by a planner agent.
The planner agent collects the feedback and try to make an enhanced travel plan.
For the planner agent, we use the templates of Direct and Reflextion (Refinement) for plan initialization and refinement respectively.

\begin{tcolorbox}[title=Multi-Agent Debate (MAD) - Spatial-Perspective Criticism Agent]
\small
Our task is to generate a travel plan based on the query <QUERY> and associated POI references.\\

Spatial Requirements: Consider the clustering of attraction POIs based on their geographical locations. 
Assign attraction POIs with close geographical locations to the same day and those with distant locations to different days. 
Ensure attractions on the same day are not too far apart.\\

There is an initial plan in place, please review whether this plan meets the spatial requirements for attraction planning and provide specific feedback for modifications.\\

<POI REFERENCE LIST>\\

<INITIAL PLAN>
\end{tcolorbox}

\begin{tcolorbox}[title=Multi-Agent Debate (MAD) - Temporal-Perspective Criticism Agent]
\small
Our task is to generate a travel plan based on the query <QUERY> and associated POI references.\\

Temporal Requirements:\\
1. Each POI must be visited during its opening hours. Prioritize the recommended start times for attractions and ensure sufficient time for each visit (based on the expected duration of the attraction).\\
2. In general, the total travel schedule for each day should not be too tight, ensuring the overall travel time is not too long, the number of attractions visited is not too large, and there is enough free time for meals, accommodation, and transportation.\\

There is an initial plan in place, please review whether this plan meets the temporal requirements for attraction planning and provide specific feedback for modifications.\\

<POI REFERENCE LIST>\\

<INITIAL PLAN>
\end{tcolorbox}

\begin{tcolorbox}[title=Multi-Agent Debate (MAD) - Semantic-Perspective Criticism Agent]
\small
Our task is to generate a travel plan based on the query <QUERY> and associated POI references.\\

Semantic Requirements: In general, prioritize popular and unique attractions that reflect the city’s characteristics.\\

There is an initial plan in place, please review whether this plan meets the semantic requirements for attraction planning and provide specific feedback for modifications.\\

<POI REFERENCE LIST>\\

<INITIAL PLAN>
\end{tcolorbox}

\begin{tcolorbox}[title=Multi-Agent Debate (MAD) - Relevance-Perspective Criticism Agent]
\small
Our task is to generate a travel plan based on the query <QUERY> and associated POI references.\\

Personalized Requirements:\\
1. If there are additional personalized constraints in the query, understand and summarize these requirements. 
When selecting attractions and planning the plan, consider these personalized constraints.\\
2. Some personalized requirements may conflict with general requirements. 
In such cases, prioritize the personalized requirements.
For example, if the query is "Special Forces-style Tourist", the overall itinerary time, number of attractions visited per day, free time, duration of attraction visits, and start times of visits may not have specific constraints, allowing for a more compact itinerary.
If the query is related to specific demands (\eg "Natural Landscape Tourism", "Family Travel", "Autumn Travel", "Chinese New Year Travel", \etc.), just choose the most popular attractions that meet the query constraints.\\

There is an initial plan in place, please review whether this plan meets the personal requirements for attraction planning and provide specific feedback for modifications.\\

<POI REFERENCE LIST>\\

<INITIAL PLAN>
\end{tcolorbox}

\begin{tcolorbox}[title=Retrieval-Augmented Planning]
\small
Our task is to generate a travel plan based on the query <QUERY> and associated POI references and retrieved trajectories.\\

Basic Requirements:\\
1. Structure the article according to the number of days, such as "Day 1", "Day 2", and "Day 3". 
If the query does not specify the number of days, use your knowledge and the attractions list to deduce the duration of the travel plan.\\
2. Plan the visit to attractions POI in the order of scheduled visit times. 
Select attractions only from the provided reference list, and do not include attractions outside the list.\\
3. Plan specific start and end times of visit in 24-hour format for each POI. 
Ensure no overlap in visit times for different POIs, leave gaps between activities.\\

More Planning Requirements:\\
a. General Requirements:\\
1. Spatial: Consider the clustering of attraction POIs based on their geographical locations. 
Assign attraction POIs with close geographical locations to the same day and those with distant locations to different days. 
Ensure attractions on the same day are not too far apart.\\
2. Time:\\
(1) Each POI must be visited during its opening hours. Prioritize the recommended start times for attractions and ensure sufficient time for each visit (based on the expected duration of the attraction).\\
(2) In general, the total travel schedule for each day should not be too tight, ensuring the overall travel time is not too long, the number of attractions visited is not too large, and there is enough free time for meals, accommodation, and transportation.\\
3. Attractions Semantics: In general, prioritize popular and unique attractions that reflect the city’s characteristics.\\
b. Personalized Requirements:\\
1. If there are additional personalized constraints in the query, understand and summarize these requirements. 
When selecting attractions and planning the plan, consider these personalized constraints.\\
2. Some personalized requirements may conflict with general requirements. 
In such cases, prioritize the personalized requirements.
For example, if the query is "Special Forces-style Tourist", the overall itinerary time, number of attractions visited per day, free time, duration of attraction visits, and start times of visits may not have specific constraints, allowing for a more compact itinerary.
If the query is related to specific demands (\eg "Natural Landscape Tourism", "Family Travel", "Autumn Travel", "Chinese New Year Travel", \etc.), just choose the most popular attractions that meet the query constraints.\\

<POI REFERENCE LIST>\\

<TRAJECTORIES>\\

Answer Format in JSON:\\
\texttt{\{"Day 1": [\{"POI name": "xxx", "Start visit time": "xxx", "End visit time": "xxx"\}, \ldots], \ldots\}}
\end{tcolorbox}

For post-retrieval methods, we first apply compression on the raw trajectories and then use the Retrieval-Augmented Planning template for travel planning.

\begin{tcolorbox}[title=Extractive Trajectories Compression]
\small
Given several reference tourist trajectories about the query <QUERY>. Please select the <EXTRACTIVE NUMBER> schemes that best meet the requirements below, and answer with the trajectory ID with explanations.\\

Extractive Requirements:\\
a. General Requirements:\\
1. Spatial: Consider the clustering of attraction POIs based on their geographical locations. 
Assign attraction POIs with close geographical locations to the same day and those with distant locations to different days. 
Ensure attractions on the same day are not too far apart.\\
2. Time:\\
(1) Each POI must be visited during its opening hours. Prioritize the recommended start times for attractions and ensure sufficient time for each visit (based on the expected duration of the attraction).\\
(2) In general, the total travel schedule for each day should not be too tight, ensuring the overall travel time is not too long, the number of attractions visited is not too large, and there is enough free time for meals, accommodation, and transportation.\\
3. Attractions Semantics: In general, prioritize popular and unique attractions that reflect the city’s characteristics.\\
b. Personalized Requirements:\\
1. If there are additional personalized constraints in the query, understand and summarize these requirements. 
When selecting attractions and planning the plan, consider these personalized constraints.\\
2. Some personalized requirements may conflict with general requirements. 
In such cases, prioritize the personalized requirements.
For example, if the query is "Special Forces-style Tourist", the overall itinerary time, number of attractions visited per day, free time, duration of attraction visits, and start times of visits may not have specific constraints, allowing for a more compact itinerary.
If the query is related to specific demands (\eg "Natural Landscape Tourism", "Family Travel", "Autumn Travel", "Chinese New Year Travel", \etc.), just choose the most popular attractions that meet the query constraints.\\

<POI REFERENCE LIST>\\

<TRAJECTORIES>\\

Answer Format in JSON:\\
\texttt{\{"Explanation": "xxx", "Extractive IDs": ["x", "x", \ldots]\}}
\end{tcolorbox}

\begin{tcolorbox}[title=Abstractive Trajectories Compression]
\small
Given several reference tourist trajectories about the query <QUERY>, please summarize, generalize, merge, and compress these information into a single trajectory according to the given requirements, with specific explanation and remarks.\\

Summarization Requirements:\\
a. General Requirements:\\
1. Spatial: Consider the clustering of attraction POIs based on their geographical locations. 
Assign attraction POIs with close geographical locations to the same day and those with distant locations to different days. 
Ensure attractions on the same day are not too far apart.\\
2. Time:\\
(1) Each POI must be visited during its opening hours. Prioritize the recommended start times for attractions and ensure sufficient time for each visit (based on the expected duration of the attraction).\\
(2) In general, the total travel schedule for each day should not be too tight, ensuring the overall travel time is not too long, the number of attractions visited is not too large, and there is enough free time for meals, accommodation, and transportation.\\
3. Attractions Semantics: In general, prioritize popular and unique attractions that reflect the city’s characteristics.\\
b. Personalized Requirements:\\
1. If there are additional personalized constraints in the query, understand and summarize these requirements. 
When selecting attractions and planning the plan, consider these personalized constraints.\\
2. Some personalized requirements may conflict with general requirements. 
In such cases, prioritize the personalized requirements.
For example, if the query is "Special Forces-style Tourist", the overall itinerary time, number of attractions visited per day, free time, duration of attraction visits, and start times of visits may not have specific constraints, allowing for a more compact itinerary.
If the query is related to specific demands (\eg "Natural Landscape Tourism", "Family Travel", "Autumn Travel", "Chinese New Year Travel", \etc.), just choose the most popular attractions that meet the query constraints.\\

<POI REFERENCE LIST>\\

<TRAJECTORIES>\\

Answer Format in JSON:\\
\texttt{\{"Explanation": "xxx", "Results": \{"Day 1": [\{"POI name": "xxx", "Remark": "xxx"\}, \ldots], \ldots\}\}}
\end{tcolorbox}

\subsection{Methodology}
\label{appendix:prompt_method}

For plan initialization, we adopt the prompt templates of Direct and Retrieval-Augmented Planning.

\begin{tcolorbox}[title=Evaluation Reflection]
\small
Our task is to generate a travel plan based on the query <QUERY> and associated POI references.\\

<POI REFERENCE LIST>\\

You have previously generated a batch of planning results and evaluated them:\\

1. Planning results are as follows (sorted in descending order of optimality):\\
<PREVIOUS PLANNING RESULTS>\\

2. Evaluation criteria (\ie optimization objectives) are as follows:\\
<CRITERIA \& OBJECTIVES>\\

Please analyze the differences in these plans based on the evaluation results. Considering the optimization objectives, reflect on what makes a good plan and how to achieve an even better plan.\\

You have already considered the following:\\
<PREVIOUS REFLECTION>\\

Now, please refine your previous reflection, providing a concise analysis for each optimization objective.
Just provide your final reflection, no need to output the analysis process.
\end{tcolorbox}

\begin{tcolorbox}[title=Plan Updating - Mutation Only]
\small
Our task is to generate a travel plan based on the query <QUERY> and associated POI references.\\

<POI REFERENCE LIST>\\

You have previously generated a batch of planning results and evaluated them:\\

1. Planning results are as follows (sorted in descending order of optimality):\\
<PREVIOUS PLANNING RESULTS>\\

2. Evaluation criteria (\ie optimization objectives) are as follows:\\
<CRITERIA \& OBJECTIVES>\\

Based on the evaluation results, you have the following considerations:\\
<REFLECTION>\\

Now, improve and optimize these plans. 
The improved set of <NUMBER OF PLANS> new plans should be distinct from each other and from the previous plans, but they should be more optimal than the previous results across all evaluation criteria.\\

Answer Format in JSON:\\
\texttt{[\{"Day 1": [\{"POI name": "xxx", "Start visit time": "xxx", "End visit time": "xxx"\}, \ldots], \ldots\}, \ldots]}
\end{tcolorbox}

\begin{tcolorbox}[title=Plan Updating - Crossover \& Mutation]
\small
Our task is to generate a travel plan based on the query <QUERY> and associated POI references.\\

<POI REFERENCE LIST>\\

You have previously generated a batch of planning results and evaluated them:\\

1. Planning results are as follows (sorted in descending order of optimality):\\
<PREVIOUS PLANNING RESULTS>\\

2. Evaluation criteria (\ie optimization objectives) are as follows:\\
<CRITERIA \& OBJECTIVES>\\

Based on the evaluation results, you have the following considerations:\\
<REFLECTION>\\

Please follow the steps below to generate new plans:\\
1. Selection: Choose two plans from the previous results that are less similar.\\
2. Crossover: Merge and process these two plans to create a new plan that combines the strengths of the original two plans.\\
3. Mutation: Based on your thoughts, further improve and optimize the new plan after the crossover.\\
Iterate and repeat the above steps until you generate <NUMBER OF PLANS> new plans. 
These new plans should be distinct from each other and from the previous plans, but they should be more optimal than the previous results across all evaluation criteria.\\

Answer Format in JSON:\\
\texttt{[\{"Day 1": [\{"POI name": "xxx", "Start visit time": "xxx", "End visit time": "xxx"\}, \ldots], \ldots\}, \ldots]}
\end{tcolorbox}

\end{document}